\pdfoutput=1
\documentclass[10pt,twocolumn,letterpaper]{article}

\PassOptionsToPackage{table}{xcolor}
\usepackage[pagenumbers]{cvpr}

\usepackage{graphicx}
\usepackage[table]{xcolor}
\usepackage{amsmath}
\usepackage{booktabs}
\usepackage{multirow}
\usepackage{array}
\usepackage{tabularx}
\usepackage{threeparttable}
\usepackage{pifont}

\newcommand{\methodName}{VeriTouch}
\newcommand{\IDMName}{VeriIDM}
\newcommand{\WMName}{VeriGen}
\newcommand{\VerifierName}{VeriEval}
\DeclareRobustCommand{\Nzero}{\ensuremath{\mathcal{N}_0}}

\definecolor{cvprblue}{rgb}{0.21,0.49,0.74}
\usepackage[pagebackref,breaklinks,colorlinks,allcolors=cvprblue]{hyperref}

\def\confName{CVPR}
\def\confYear{2027}

\title{\methodName: Verified Visuo-Tactile World Models that\\Improve Robot Policy Learning}

\author{
Shuxun Wang$^{1,*}$\enspace
Xiaonan Jiang$^{2,*}$\enspace
Zhaoyu Wang$^{1,*}$\enspace
Ziyuan Chu$^{1,*}$\enspace
Xihang Yue$^{1}$\enspace
Pengfei Liu$^{3}$\\[0.2em]
Haodong Zhu$^{4}$\enspace
Shiqi Liu$^{5}$\enspace
Mingzhi Zheng$^{6}$\enspace
Liang Su$^{7}$\enspace
Linchao Zhu$^{1,\dagger}$\\[0.65em]
\textsuperscript{1}\textbf{Zhejiang University}\quad
\textsuperscript{2}City University of Hong Kong\quad
\textsuperscript{3}Nanyang Technological University\\[0.10em]
\textsuperscript{4}Technical University of Munich\quad
\textsuperscript{5}Zhejiang University City College\\[0.10em]
\textsuperscript{6}Nanjing University\quad
\textsuperscript{7}Hangzhou Institute for Advanced Study, UCAS\\[0.55em]
$^{*}$Equal contribution.\quad $^{\dagger}$Corresponding author.
}

\makeatletter
\patchcmd{\@maketitle}{\vskip .375in}{\vspace*{-1.5cm}}{}{\PackageError{veritouch}{CVPR title spacing patch failed}{}}
\g@addto@macro\@maketitle{%
  \noindent\begin{minipage}{\textwidth}
  \centering
  \includegraphics[width=\linewidth]{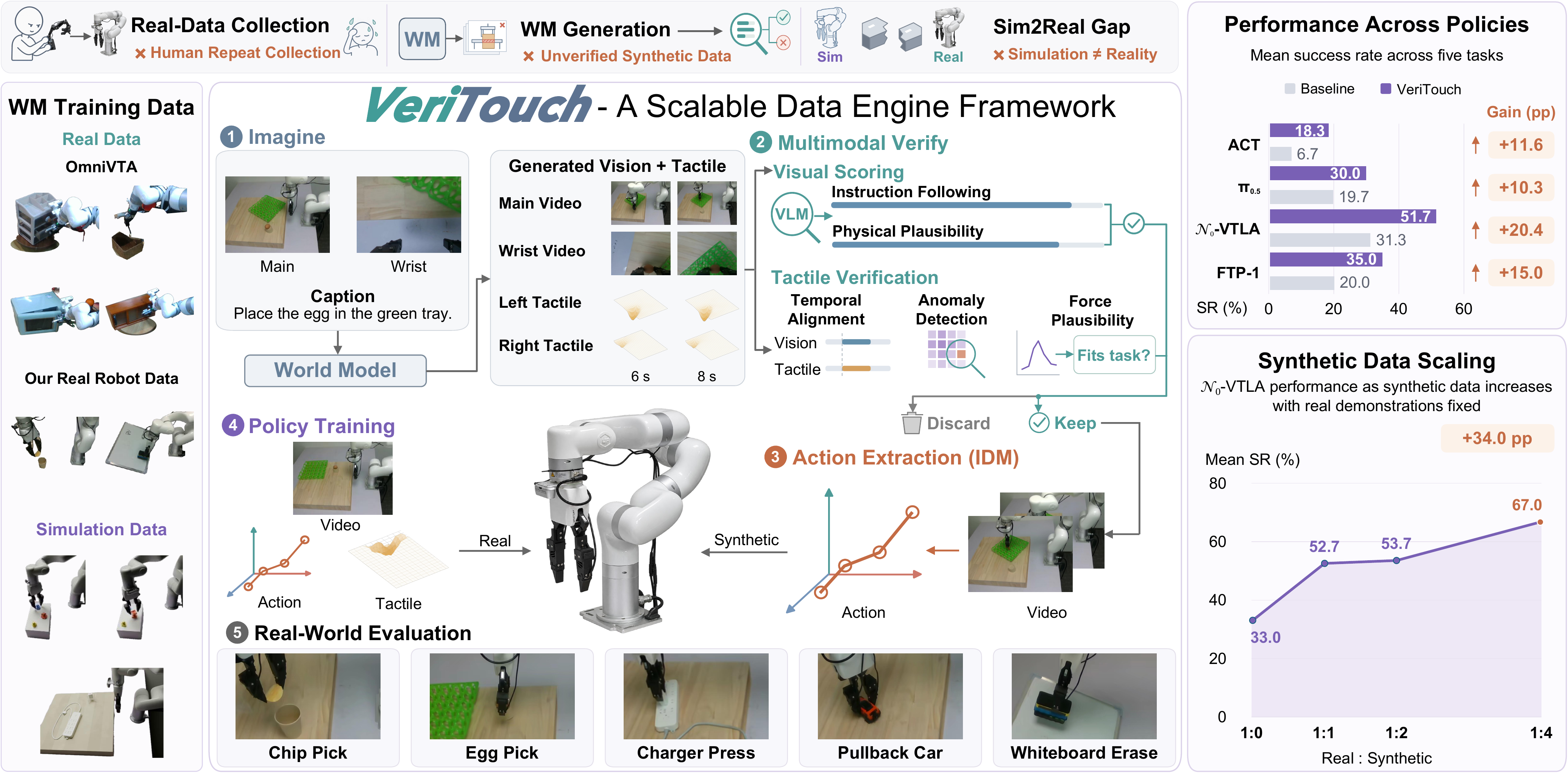}
  \captionsetup{hypcap=false}
  \captionof{figure}{\textbf{Overview of \methodName{}.} The data engine generates visuo-tactile trajectories, filters them with multimodal verification, and extracts actions for policy training. Adding verified synthetic data improves success rates across four policies on five real-world tasks.}
  \label{fig:teaser}
  \end{minipage}\par
  \vspace{0.6em}
}
\makeatother

\begin{document}

\maketitle
\begingroup
\renewcommand{\thefootnote}{}
\makeatletter
\long\def\@makefntext#1{\noindent#1}
\makeatother
\footnotetext{\raggedright Project page: \href{https://zjuscl.github.io/veritouch/\#overview}{https://zjuscl.github.io/veritouch/}\\
Emails: \href{mailto:shuxun@zju.edu.cn}{shuxun@zju.edu.cn}, \href{mailto:zhulinchao@zju.edu.cn}{zhulinchao@zju.edu.cn}.}
\endgroup
\begin{abstract}
Scaling visuo-tactile policy learning is constrained by the high cost of real-world data collection and the difficulty of building realistic simulations. Generative models offer a scalable source of training data, yet useful demonstrations require plausible tactile feedback, temporal alignment with visual contact, and accurate actions.
To this end, we present \textbf{\methodName{}}, a framework for generating and verifying synthetic visuo-tactile trajectories for policy learning. Conditioned on initial images and task instructions, our world model, \WMName{}, jointly generates high-fidelity dual-view videos and realistic tactile sequences over complete manipulation trajectories. By integrating a dedicated tactile expert with a pretrained video model, it captures realistic visual details and tactile deformation dynamics through joint visuo-tactile modeling. Our multimodal verifier \VerifierName{} then checks visual quality while focusing on tactile reliability, assessing tactile realism, contact-force plausibility, and temporal alignment between visible contact and tactile signals, thereby directly filtering out unreliable sequences. Finally, our inverse dynamics model, \IDMName{}, combines global scene information from both camera views with fine-grained gripper details and motion across neighboring frames to extract actions from verified videos.
Experiments on five real-world tasks show that synthetic data improves all four evaluated policies, with further scaling synthetic trajectories substantially boosting policy performance without additional real-world data collection.
\end{abstract}

\raggedbottom

\section{Introduction}
\label{sec:intro}
Contact-rich manipulation requires robots to reason about physical contact, which is often difficult to infer from visual observations alone. By providing direct feedback on contact states, tactile sensing enables visuo-tactile policies to better handle contact-rich tasks~\cite{neoteai2026n0vtla,yuan2026ftp,bi2026vla}. However, collecting visuo-tactile data in the real world requires expensive tactile sensors~\cite{do2022densetact} and labor-intensive teleoperation~\cite{dass2023pato,wu2023gello}. Simulation reduces the need for real-world data collection but requires realistic environments and accurate modeling of objects' physical properties~\cite{aljalbout2026reality,zhang2020contact,wang2022tacto}. To reduce these costs, recent work synthesizes manipulation videos and
extracts the corresponding actions for policy training~\cite{jang2025dreamgen,he2026cosmossurgical}.

However, extending video-based synthesis to visuo-tactile demonstrations requires generating plausible tactile responses that are synchronized with visible contact. Existing approaches
augment training with synthetic rollouts~\cite{huang2026vitacworld} or
corrective segments for policy failures~\cite{Liu_2026_TACO}, with repeated
rounds of data generation and policy retraining. Other approaches rely on simulation to generate or validate physically feasible trajectories~\cite{jiang2026grounded,he2026vdreamer}. Meanwhile, existing verification methods assess kinematic and tactile plausibility~\cite{Liu_2026_TACO} or score predicted tactile signals against task instructions~\cite{Wu_2026_ViTaL}. While these methods improve different aspects of data generation and quality control, several challenges remain. Iterative augmentation requires repeated data generation and policy retraining, while simulation-based approaches depend on additional environments, increasing the cost and complexity of generating training data. Existing verification methods can identify specific errors, but lack a joint assessment of tactile plausibility and temporal alignment with visual contact throughout complete demonstrations. These quality requirements, together with the costs of iterative augmentation and simulation, motivate a more efficient and comprehensive pipeline for generating reliable visuo-tactile training data.

As shown in Fig.~\ref{fig:teaser}, we introduce \methodName{}, a framework that jointly generates visual and tactile observations, verifies the resulting sequences without iterative retraining and simulation replay, and extracts actions from the sequences that pass verification. Our world model, \WMName{}, extends a pretrained video backbone~\cite{agarwal2026cosmos} with a dedicated tactile expert. It uses cross-view attention to combine scene context with close-up contact details and jointly models visual and tactile dynamics. Alongside checks for physical plausibility and instruction following, our multimodal verifier, \VerifierName{}, assesses tactile reconstruction error and task-dependent contact forces, and checks the temporal alignment of visual and tactile contact intervals. By filtering out unreliable sequences before action extraction, \methodName{} produces higher-quality synthetic visuo-tactile training data.
Finally, \IDMName{} uses local gripper cues and temporal context to extract actions from the retained videos.

We evaluate \methodName{} on five real-world contact-rich tasks using two vision-based and two visuo-tactile policies. Augmenting real demonstrations with our synthetic data improves all four policies, increasing the overall average success rate from 19.4\% with real data alone to 33.8\%. Moreover, policy performance continues to improve as the synthetic dataset scales without collecting additional real-world demonstrations, more than doubling the average success rate from 33.0\% to 67.0\% when the amount of synthetic data reaches four times that of real data. Ablations show that generated tactile signals improve policy performance and that training on data selected by our multimodal verifier yields higher success rates than training on synthetic data with visual-only filtering or no filtering. These results demonstrate that verified synthetic visuo-tactile trajectories improve policy performance on contact-rich manipulation tasks.
Our main contributions are as follows:
\begin{itemize}
    \item We introduce \methodName{}, a framework for synthesizing high-quality visuo-tactile training trajectories. Its world model jointly captures visual and tactile contact dynamics, while an IDM preserves fine motion details to recover precise actions.
    \item We develop a multimodal verifier that checks tactile plausibility and visual--tactile contact consistency alongside visual quality, filtering unreliable observations before extracting actions without physical or simulated replay.
    \item We improve all four policies and show that scaling synthetic data more than doubles average success rate over the real-data baseline without collecting additional real-world demonstrations.
\end{itemize}

\section{Related Work}
\label{sec:related}

\paragraph{World Models for Robot Data Generation.}
Generative world models can synthesize videos of robot manipulation from initial images and task instructions. DreamGen~\cite{jang2025dreamgen} converts generated videos into robot demonstrations by recovering pseudo-actions, while RoboDream~\cite{robodream2026} expands scene diversity through motion-conditioned generation. For contact-rich manipulation, visuo-tactile world models predict future visual and tactile observations to support planning and control, sometimes jointly predicting robot actions~\cite{higuera2026visuo,zheng2026omnivta,team2026n_0,tian2026vt,lou2026dream}, whereas Physical Canvas~\cite{xing2026physicalcanvas} uses a learned simulator for policy evaluation and optimization. Closer to our focus on visuo-tactile data synthesis, ViTacWorld~\cite{huang2026vitacworld} generates action-conditioned visuo-tactile rollouts, and TACO~\cite{Liu_2026_TACO} produces corrective segments near policy failures for iterative post-training. Sharing the goal of expanding visuo-tactile training data, \methodName{} jointly models visual scene evolution and tactile deformation dynamics to synthesize observation sequences covering the full manipulation process from initial images and task instructions, without conditioning on future robot actions.

\paragraph{Verification of Synthetic Demonstrations.}
Visual verifiers use video-language models to judge whether generated videos follow the task instruction and depict physically plausible motion~\cite{jang2025dreamgen,bansal2025videophy,bansal2026videophy}. By contrast, tactile verifiers assess the physical contact directly. TACO filters corrective segments by checking predicted force and torque signals against task-specific bounds~\cite{Liu_2026_TACO}. ViTaL uses both modalities in policy steering: visual predictions guide long-horizon action selection, while predicted tactile features are scored against phase-specific contact descriptions to refine short-horizon actions~\cite{Wu_2026_ViTaL}. For data filtering, a plausible video and a plausible tactile sequence can still disagree on when contact occurs. \methodName{} therefore verifies visual quality, tactile plausibility, and cross-modal contact timing throughout each demonstration.

\section{Method}
\label{sec:method}

\subsection{Problem Formulation}
\label{sec:problem_formulation}

\noindent\textbf{Visuo-tactile demonstrations.}
In this work, we consider contact-rich manipulation with a robot equipped with a main camera, a wrist camera, and two tactile sensors mounted on the gripper. At time $t$, the robot receives an observation $o_t=(I^{\mathrm{m}}_t, I^{\mathrm{w}}_t, \mathbf{X}^{\mathrm{L}}_t, \mathbf{X}^{\mathrm{R}}_t)$, where $I^{\mathrm{m}}_t$ and $I^{\mathrm{w}}_t$ are the main- and wrist-camera images and $\mathbf{X}^{\mathrm{L}}_t,\mathbf{X}^{\mathrm{R}}_t\in\mathbb{R}^{N\times 3}$ are the 3D displacement fields sampled at $N$ points on the left and right tactile sensors, and executes an action $a_t$. A demonstration of a task with instruction $\ell$ is a trajectory
\begin{equation}
    \tau=\bigl(\ell,\ \{(o_t,\, a_t)\}_{t=0}^{T}\bigr).
    \label{eq:trajectory}
\end{equation}

\noindent\textbf{Core problem.}
Given a small set of real-world demonstrations $\mathcal{D}_{\mathrm{real}}$, our goal is to synthesize a dataset $\mathcal{D}_{\mathrm{syn}}$ of new trajectories in the same format as Eq.~\eqref{eq:trajectory}. The synthetic data can be used to train VLA models together with the limited real-world data (Sec.~\ref{sec:policy_learning}). Since these trajectories are obtained without physical or simulated replay, each synthetic trajectory must satisfy three requirements. (i) It should span the full manipulation process and differ from those in $\mathcal{D}_{\mathrm{real}}$. (ii) Its visual and tactile signals should be physically plausible and consistent with each other. (iii) It should be annotated with actions consistent with its observations.

\begin{figure*}[t]
    \centering
    \includegraphics[width=0.90\textwidth]{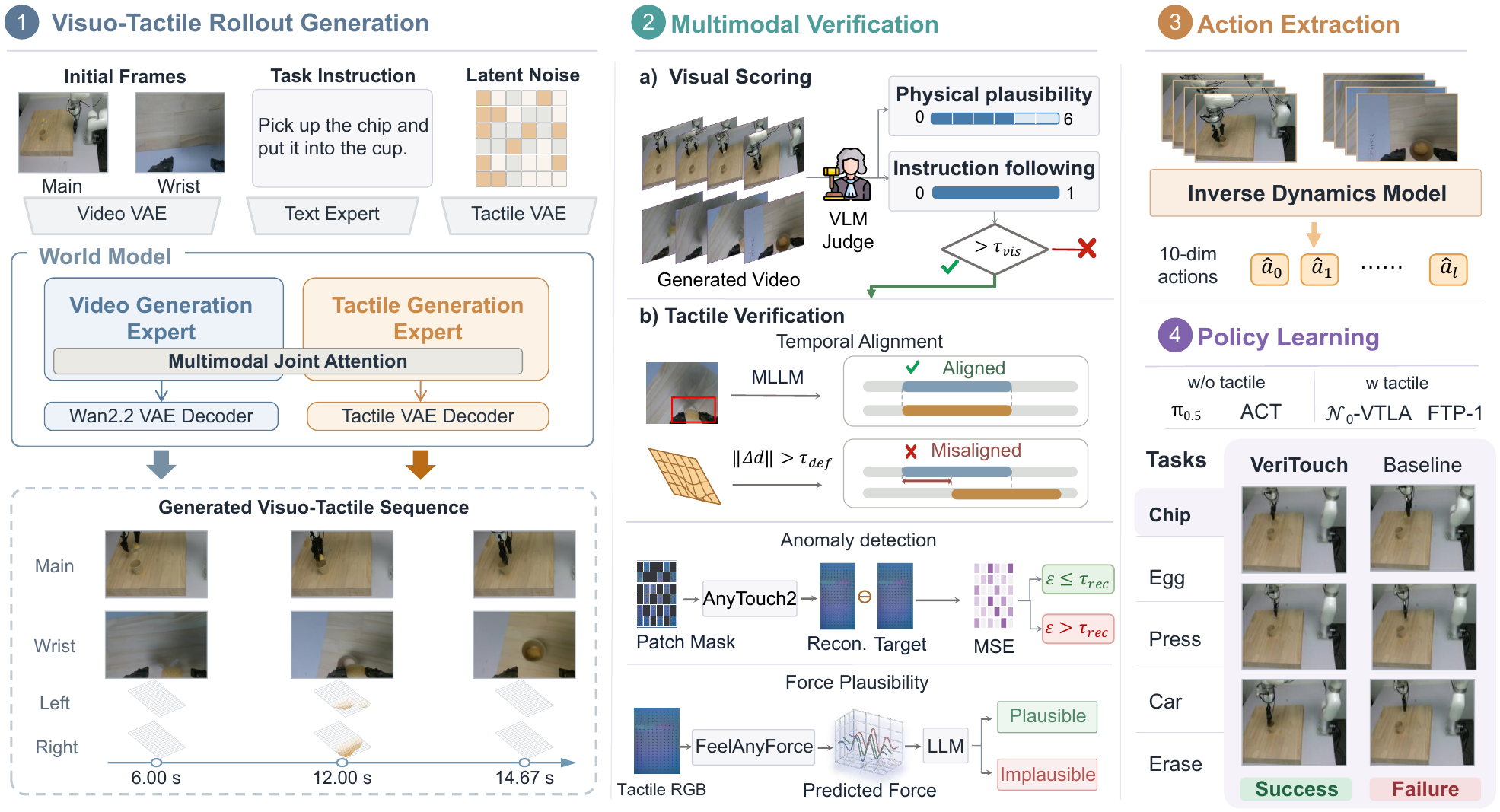}
    \caption{\textbf{Overview of the \methodName{} framework.}
Given initial multi-view observations and a task instruction, VeriTouch jointly generates dual-view video and tactile sequences through \WMName{}.
\VerifierName{} then filters unreliable trajectories by assessing visual plausibility, instruction following, tactile anomalies, force plausibility, and visuo-tactile contact consistency.
\IDMName{} extracts robot actions from the verified videos, producing synthetic demonstrations for training both vision-based and visuo-tactile policies. }
    \label{fig:method_overview}
\end{figure*}

\subsection{\methodName{} Framework}
\label{sec:instantiation}

\subsubsection{Overview}
\label{sec:framework_overview}

To address this problem, we propose \methodName{}. It consists of three components, a generator $G$, a verifier $V$, and an extractor $E$, which respectively address requirements (i), (ii), and (iii). The three components are formulated as
\begin{equation}
    \begin{alignedat}{3}
    &\text{Generator}\quad && G:\ \bigl(I_0,\, \ell\bigr) && \;\rightarrow\; \hat o_{1:T},\\
    &\text{Verifier}       && V:\ \bigl(\hat o_{1:T},\, \ell\bigr) && \;\rightarrow\; \{0,1\},\\
    &\text{Extractor}        && E:\ \bigl(\hat I^{\mathrm{m}}_{1:T},\,\hat I^{\mathrm{w}}_{1:T}\bigr) && \;\rightarrow\; \hat a_{1:T}.
    \end{alignedat}
    \label{eq:components}
\end{equation}
where $I_0=(I^{\mathrm{m}}_0, I^{\mathrm{w}}_0)$ denotes the initial images and $\hat o_{1:T}=(\hat I^{\mathrm{m}}_{1:T}, \hat I^{\mathrm{w}}_{1:T}, \widehat{\mathbf{X}}^{\mathrm{L}}_{1:T}, \widehat{\mathbf{X}}^{\mathrm{R}}_{1:T})$ denotes a generated observation sequence. The generator $G$ samples an observation sequence $\hat o_{1:T}$ conditioned only on the initial images $I_0$ and the instruction $\ell$, without any action input. The verifier $V$ accepts $\hat o_{1:T}$ if $V(\hat o_{1:T},\ell)=1$ and rejects it otherwise. The extractor $E$ takes the two synchronized RGB streams from each accepted observation sequence as input and predicts the action sequence $\hat a_{1:T}$. The synthetic dataset contains all accepted samples together with their extracted actions:
\begin{equation}
    \mathcal{D}_{\mathrm{syn}}=\bigl\{\bigl(\ell,\, \hat o_{1:T},\, \hat a_{1:T}\bigr) \bigm| V(\hat o_{1:T},\, \ell)=1\bigr\}.
    \label{eq:synthetic_dataset}
\end{equation}
In this paper, we instantiate $G$ with the visuo-tactile world model \WMName{} (Sec.~\ref{sec:world_model}), $V$ with \VerifierName{}, a multimodal verifier that combines visual and tactile tests (Sec.~\ref{sec:multimodal_verifier}), and $E$ with \IDMName{}, an inverse dynamics model that extracts actions from the generated videos (Sec.~\ref{sec:idm}). Fig.~\ref{fig:method_overview} illustrates the resulting pipeline.

\noindent\textbf{Discussion.}
 By separating generation, verification, and action extraction, \methodName{} provides a pipeline for synthesizing complete visuo-tactile demonstrations without physical or simulated replay. Unlike iterative augmentation approaches~\cite{huang2026vitacworld,Liu_2026_TACO}, our framework does not require repeated rounds of policy retraining. Instead, generated trajectories are verified for visual and tactile plausibility, including temporal contact consistency, before actions are extracted. This allows synthetic training data to be expanded without collecting additional real-world demonstrations.

\subsubsection{Generator: Visuo-Tactile World Model}
\label{sec:world_model}

For the generator $G$, we propose \WMName{}, a visuo-tactile world model that generates the full observation sequence $\hat o_{1:T}$ from the initial images $I_0$ and the instruction $\ell$. Its core design is to incorporate a tactile modality into a pretrained video generation model. It builds on the Mixture-of-Transformers architecture of Cosmos3-Nano~\cite{agarwal2026cosmos} and reuses its pretrained video and text experts. We extend the video expert, which originally generates a single view, to generate the main and wrist views jointly (Appendix~\ref{sec:appendix_wm_architecture}). To generate bilateral tactile signals together with the videos, we further introduce a tactile expert, which is described below.

\noindent\textbf{Tactile generation.}
We add a tactile expert that generates the tactile deformation of both sensors. Following OmniVTA~\cite{zheng2026omnivta}, we represent the deformation of each sensor by 3D displacement fields on its sensing surface, \ie, $\mathbf{X}^{\mathrm{L}}_t$ and $\mathbf{X}^{\mathrm{R}}_t$ in Sec.~\ref{sec:problem_formulation}, rather than by raw tactile images. Compared with tactile RGB images, these displacement fields capture contact-induced surface deformation at a much lower spatial resolution, enabling efficient tactile modeling.
To generate these displacements, we train a TactileVAE on OmniViTac that encodes them into latent tokens, and design the tactile expert as a Diffusion Transformer ~\cite{peebles2023scalable} that generates tactile latent sequences.
Specifically, the tactile expert alternates spatial and temporal attention blocks. Spatial attention jointly processes tokens from both sensors at each latent timestep, while temporal attention attends to tokens at the same spatial location across timesteps. Together, these blocks capture how the deformation signals from both tactile sensors evolve during contact.

The tactile expert is connected to the video expert through joint attention on a shared time axis, so that the generated tactile signals follow the contact visible in the videos. Details are provided in Appendix~\ref{sec:appendix_wm_architecture}.

\noindent\textbf{Training.}
\WMName{} is trained jointly on video and tactile latents with a rectified flow-matching objective~\cite{liu2022flow}. Built on OmniVTA~\cite{zheng2026omnivta}, the tactile loss gives more weight to regions with large or rapid deformation, which are otherwise outweighed by the many nearly static regions. We first pretrain the model on 16,593 trajectories from OmniViTac~\cite{zheng2026omnivta}, then adapt it on 3,056 trajectories from the Gello robot subset of OmniViTac and 3,275 task-aligned  visuo-tactile simulation trajectories, and finally fine-tune it on 200 real-robot trajectories from the five target tasks. Training data, objectives, and optimization details are provided in Appendix~\ref{sec:app_wm_training}.

\subsubsection{Verifier: Multimodal Verification}
\label{sec:multimodal_verifier}

For the verifier $V$, we introduce \VerifierName{}, which combines visual and tactile checks and accepts a generated sample only if it passes both.
Verification only operates on the generated observations without physical or simulated replay.

\noindent\textbf{Visual verification.}
Errors in fine physical details, such as implausible gripper-object contact or unrealistic object deformation, compromise the accuracy of action extraction from the generated video.
To detect such errors, we fine-tune a vision-language model (VLM) on about 900 human-annotated generated videos. The model evaluates each video along evaluation dimensions covering instruction following and physical plausibility, producing both numerical scores and explanations of its judgments. Any video whose overall score falls below a predefined threshold $\tau_{\text{vis}}$ is discarded. Evaluation dimensions and other details are provided in Appendix~\ref{sec:verifier_implementation}.

\noindent\textbf{Tactile verification.}
The tactile validation assesses three properties.
First, the contact intervals in the tactile signals should align with those in the video, since temporal misalignment is the most common error in generated tactile data.
We train a small VLM to determine whether contact occurs in each wrist-camera frame. Tactile contact intervals are obtained from the pointwise $L_2$ norms of the 3D displacement fields, and samples with misaligned intervals are rejected.
Second, the tactile signals should conform to the distribution of real sensor readings. We convert them to tactile RGB images (see Appendix~\ref{sec:appendix_tactile_rgb_reconstruction}) using a renderer built on the Xense SDK~\cite{xense_sdk}, pass them through the pretrained AnyTouch2~\cite{Feng_2026_AnyTouch2} encoder and decoder, and reject samples whose masked reconstruction mean squared error exceeds a predefined threshold $\tau_{\mathrm{rec}}$.
Third, the contact force should be commensurate with the physical demands of the task.
We estimate it with FeelAnyForce~\cite{Shahidzadeh_2025_FeelAnyForce} and use a VLM to judge whether its magnitude is physically plausible for the instruction. Details and the evaluation of all three tests are provided in Appendix~\ref{sec:verifier_quality}.

\subsubsection{Extractor: Inverse Dynamics Model}
\label{sec:idm}

For the extractor $E$, we propose \IDMName{}, an inverse dynamics model that predicts robot actions from synchronized dual-view RGB videos. The model combines global and local visual features and uses neighboring frames to refine its action predictions.

\noindent\textbf{Model Architecture.}
To capture the workspace layout, \IDMName{} fuses multiscale features extracted from both camera views by a shared ResNet-18~\cite{He_2016_CVPR}. \IDMName{} augments the shared representation with two parallel detail branches. The dense spatial branch retains a finer feature grid from an intermediate ResNet-18 layer, and the lightweight CNN branch extracts local features directly from both RGB views. Temporal convolutions then refine the shared representation using neighboring frames. The pose head uses the refined representation. The gripper head estimates closure from local cues such as the visible gripper opening and adds a residual correction from the shared temporal features to account for motion across frames. Architecture details are provided in Appendix~\ref{sec:app_idm_architecture}.

\noindent\textbf{Training.}
\IDMName{} is trained on 355 real-robot and 540 simulated trajectories from eight tasks. The loss supervises per-frame action predictions and penalizes errors in position increments and relative rotations between adjacent frames. Training data, objectives, and configurations are provided in Appendix~\ref{sec:idm_setup}.

\noindent\textbf{Action Extraction.}
At inference, \IDMName{} processes the video in overlapping 17-frame clips. When a frame appears in multiple clips, \IDMName{} fuses its action predictions, giving greater weight to estimates near each clip's center, where the preceding and following context is more balanced. This weighting lets the contribution of each clip change gradually across the overlap, reducing discontinuities at clip boundaries.

\section{Experiments}
\label{sec:experiments}

\begin{table*}[!htbp]
\centering
\caption{\textbf{Real-world policy performance.}
Baseline uses real demonstrations only. \methodName{} w/o \VerifierName{} adds unfiltered synthetic demonstrations, while \methodName{} adds verified synthetic demonstrations; both use equal numbers of real and synthetic demonstrations.
SR denotes the average success score over 30 trials, while Bo3 reports the best-of-three score averaged across 10 test layouts.
Overall averages results across five tasks.
Bold indicates the highest SR within each policy. All scores are percentages.}
\label{tab:real_policy}
\footnotesize
\setlength{\tabcolsep}{2pt}
\begin{tabularx}{\textwidth}{@{}lcl*{12}{>{\centering\arraybackslash}X}@{}}
\toprule
 & & & \multicolumn{2}{c}{\shortstack{\textit{Chip}\\\textit{Pick}}} & \multicolumn{2}{c}{\shortstack{\textit{Egg}\\\textit{Pick}}} & \multicolumn{2}{c}{\shortstack{\textit{Charger}\\\textit{Press}}} & \multicolumn{2}{c}{\shortstack{\textit{Pullback}\\\textit{Car}}} & \multicolumn{2}{c}{\shortstack{\textit{Whiteboard}\\\textit{Erase}}} & \multicolumn{2}{c}{Overall} \\
\cmidrule(lr){4-5}\cmidrule(lr){6-7}\cmidrule(lr){8-9}\cmidrule(lr){10-11}\cmidrule(lr){12-13}\cmidrule(lr){14-15}
Policy & Tactile & Method & Bo3 & SR & Bo3 & SR & Bo3 & SR & Bo3 & SR & Bo3 & SR & Bo3 & SR \\
\midrule
& & Baseline & 10.0 & 3.3 & 20.0 & 20.0 & 0.0 & 0.0 & 10.0 & 3.3 & 10.0 & 6.7 & 10.0 & 6.7 \\
& & \methodName{} w/o \VerifierName{} & 10.0 & 6.7 & 20.0 & 20.0 & 10.0 & 10.0 & 20.0 & 6.7 & 10.0 & 3.3 & 14.0 & 9.3 \\
\rowcolor{blue!5}
\cellcolor{white}\multirow{-3}{*}{ACT~\cite{zhao2023learning}} & \cellcolor{white}\multirow{-3}{*}{\ding{55}} & \methodName{} & 30.0 & \textbf{10.0} & 60.0 & \textbf{50.0} & 30.0 & \textbf{13.3} & 10.0 & \textbf{10.0} & 15.0 & \textbf{8.3} & 29.0 & \textbf{18.3} \\
\addlinespace[4pt]
& & Baseline & 0.0 & 0.0 & 30.0 & 23.3 & 40.0 & 23.3 & 40.0 & 30.0 & 35.0 & 21.7 & 29.0 & 19.7 \\
& & \methodName{} w/o \VerifierName{} & 20.0 & \textbf{10.0} & 50.0 & 30.0 & 40.0 & 30.0 & 40.0 & 36.7 & 5.0 & 1.7 & 31.0 & 21.7 \\
\rowcolor{blue!5}
\cellcolor{white}\multirow{-3}{*}{$\pi_{0.5}$~\cite{physicalintelligence2025pi05}} & \cellcolor{white}\multirow{-3}{*}{\ding{55}} & \methodName{} & 20.0 & 6.7 & 60.0 & \textbf{40.0} & 50.0 & \textbf{33.3} & 70.0 & \textbf{40.0} & 35.0 & \textbf{30.0} & 47.0 & \textbf{30.0} \\
\addlinespace[4pt]
& & Baseline & 40.0 & 20.0 & 60.0 & 50.0 & 30.0 & 16.7 & 90.0 & 50.0 & 35.0 & 20.0 & 51.0 & 31.3 \\
& & \methodName{} w/o \VerifierName{} & 50.0 & 40.0 & 60.0 & 50.0 & 30.0 & 23.3 & 50.0 & 33.3 & 30.0 & 20.0 & 44.0 & 33.3 \\
\rowcolor{blue!5}
\cellcolor{white}\multirow{-3}{*}{\Nzero{}-VTLA~\cite{neoteai2026n0vtla}} & \cellcolor{white}\multirow{-3}{*}{\ding{51}} & \methodName{} & 60.0 & \textbf{43.3} & 80.0 & \textbf{63.3} & 50.0 & \textbf{33.3} & 100.0 & \textbf{73.3} & 60.0 & \textbf{45.0} & 70.0 & \textbf{51.7} \\
\addlinespace[4pt]
& & Baseline & 30.0 & 23.3 & 30.0 & 23.3 & 40.0 & 20.0 & 30.0 & 16.7 & 35.0 & 16.7 & 33.0 & 20.0 \\
& & \methodName{} w/o \VerifierName{} & 50.0 & 33.3 & 40.0 & 36.7 & 30.0 & 23.3 & 30.0 & 30.0 & 20.0 & 13.3 & 34.0 & 27.3 \\
\rowcolor{blue!5}
\cellcolor{white}\multirow{-3}{*}{FTP-1~\cite{yuan2026ftp}} & \cellcolor{white}\multirow{-3}{*}{\ding{51}} & \methodName{} & 60.0 & \textbf{43.3} & 50.0 & \textbf{40.0} & 50.0 & \textbf{36.7} & 50.0 & \textbf{36.7} & 35.0 & \textbf{18.3} & 49.0 & \textbf{35.0} \\
\bottomrule
\end{tabularx}
\end{table*}

We evaluate \methodName{} on five real-world contact-rich tasks: \textit{chip pick}, \textit{egg pick}, \textit{charger press}, \textit{pullback car}, and \textit{whiteboard erase}. We collect 50 real-world demonstrations for each task. For policy training, we use all 50 demonstrations for the first two tasks and 25 for each of the remaining three tasks. Our experiments first examine policy performance and the effect of increasing synthetic data, then evaluate generated observations and extracted actions, followed by component ablations. Hardware and data collection details are provided in Appendix~\ref{sec:app_policy_setup}.

\subsection{Synthetic Data for Policy Learning}
\label{sec:policy_learning}
\paragraph{Experimental Settings.}
To evaluate the effectiveness of synthetic demonstrations across different observation modalities, we consider four policies: two vision-based policies, ACT~\cite{zhao2023learning} and $\pi_{0.5}$~\cite{physicalintelligence2025pi05}, and two visuo-tactile policies, \Nzero{}-VTLA~\cite{neoteai2026n0vtla} and FTP-1~\cite{yuan2026ftp}. For each task, we capture main- and wrist-camera images of multiple real-world initial configurations and use VeriGen to generate synthetic trajectories conditioned on these images and task instructions. We compare three training settings: real demonstrations only (\textit{Baseline}), real demonstrations augmented with unfiltered synthetic trajectories (\methodName{} w/o \VerifierName{}), and real demonstrations augmented with verified synthetic trajectories (\methodName{}). Both synthetic settings use the same generator and action extractor, but only the latter applies multimodal verification before action extraction. Since the IDM output more closely reflects the robot’s current state, we apply a temporal shift to obtain the target actions for policy training. Unless otherwise specified, we use equal numbers of real and synthetic demonstrations, sampling them at a 1:1 ratio within each training batch.  For each policy--task pair, we use the same training steps and other hyperparameters across all three settings. We evaluate each policy on 10 unseen initial layouts per task, with three trials per layout. We report SR as the average score over all 30 trials and Best-of-3 (Bo3) as the average of the highest score among the three trials for each of the 10 layouts, both expressed as percentages. Full success scores 1 and failure scores 0; partial success on \textit{whiteboard erase} scores 0.5. Further training and evaluation details are provided in Appendix~\ref{sec:appendix_n0vtla}.

\paragraph{Performance Across Policies.}
Table~\ref{tab:real_policy} shows that verified synthetic demonstrations improve average SR across all four policies. Averaged across policies, SR increases from 19.4\% with real data alone to 33.8\% with \methodName{}, while Bo3 increases from 30.8\% to 48.8\%. Without verification, average SR reaches 22.9\%, while average Bo3 remains at 30.8\%, matching the real-data baseline. Thus, adding generated trajectories alone does not consistently improve performance across unseen layouts. Verification improves both aggregate metrics for every policy, supporting its role in selecting useful synthetic demonstrations for contact-rich policy learning.

\begin{figure}[!t]
\centering
\includegraphics[width=\columnwidth]{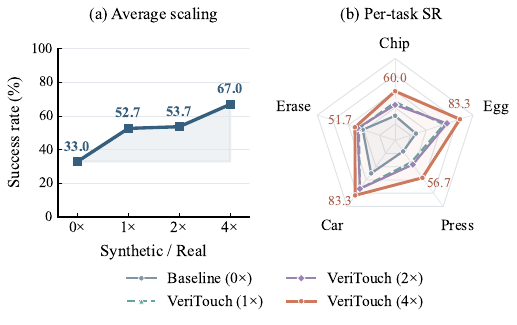}
\caption{\textbf{Effect of synthetic data scaling.}
We vary the amount of synthetic data used to train \Nzero{}-VTLA while keeping real demonstrations and the \methodName{} pipeline fixed.
(a) Average SR across five tasks at different synthetic-to-real data ratios.
(b) Per-task SR across four data ratios. Labels show the SR at $4\times$.}
\label{fig:data_scaling}
\end{figure}

\paragraph{Synthetic Data Scaling.}
We further examine whether increasing verified synthetic data can improve policy performance without collecting additional real-world demonstrations. We conduct this experiment using \Nzero{}-VTLA while keeping the real demonstrations and the \methodName{} pipeline fixed. We use the same training hyperparameters across data ratios for each task, with a separate configuration from Table~\ref{tab:real_policy} to allow sufficient policy training (Appendix~\ref{sec:app_scaling_configuration}). As shown in Fig.~\ref{fig:data_scaling}, the average SR generally improves with increasing synthetic data, reaching 67.0\% with four times as many synthetic demonstrations as real ones, more than doubling the real-only baseline of 33.0\%. These results highlight the scalability of \methodName{}, enabling substantial policy improvements through additional verified demonstrations without further real-world data collection.

\begin{figure*}[t]
\centering
\begin{minipage}{\textwidth}
\centering
\includegraphics[width=\textwidth]{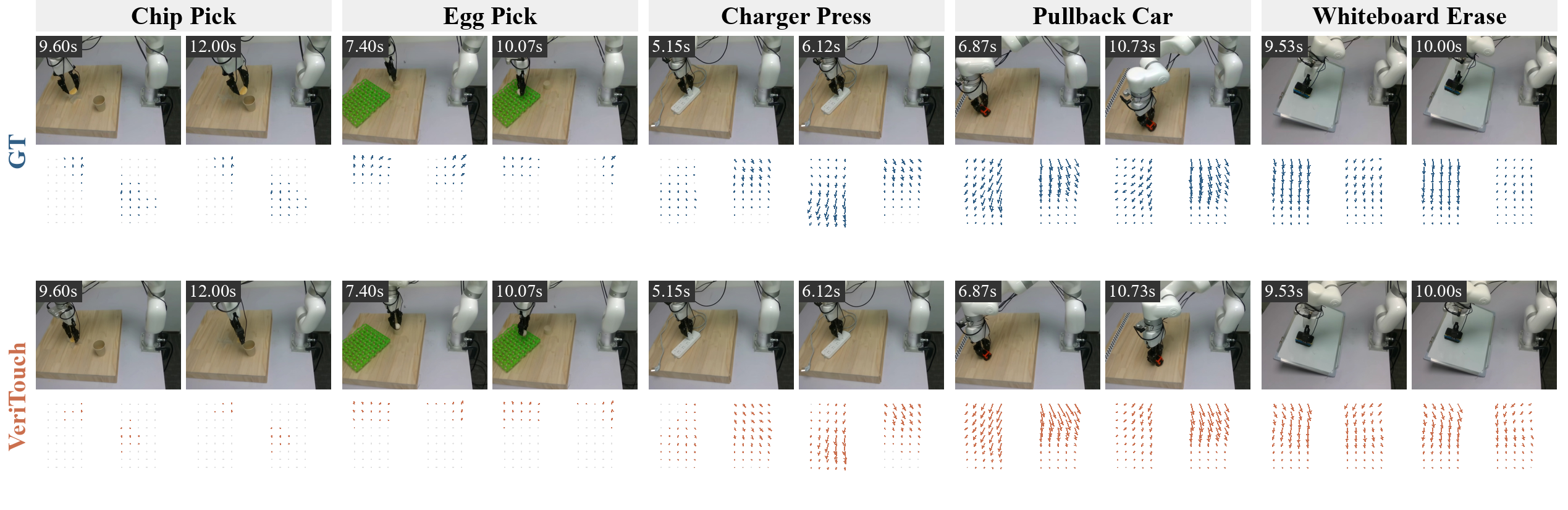}
\captionsetup{skip=4pt}\captionof{figure}{\textbf{Comparison of generated and real visuo-tactile trajectories.}
Ground truth (top) and VeriTouch (bottom) are compared at two matched timestamps across five held-out task episodes.
The tactile vector fields below each frame visualize the XY displacements of the left and right sensors, using a shared $3\times$ display scale.}
\label{fig:verified_trajectories}
\end{minipage}
\end{figure*}

\subsection{Quality of Synthetic Demonstrations}

\paragraph{Tactile and Visual Generation Quality.}
\label{sec:wm_generation_quality}

Figure~\ref{fig:verified_trajectories} compares generated visual and bilateral tactile observations with ground truth from held-out test episodes at matched timestamps. The XY displacement arrows show how the direction and magnitude of tactile deformation change over time for each finger.

Quantitatively, we compare \WMName{} with two visuo-tactile world models, $\mathcal{N}_0$-TWAM~\cite{team2026n_0} and Dream-Tac~\cite{lou2026dream}, and five video generation models. All methods are evaluated on 25 test episodes with three rollouts per episode.

We first evaluate tactile deformation accuracy and contact dynamics in Table~\ref{tab:wm_tactile_quality}. MSE and 3DEPE measure errors in 3D displacement fields, while Shape Cosine measures the spatial similarity of contact shapes. Temporal IoU measures the overlap between predicted and reference contact intervals, while Temporal Corr. measures the temporal correlation of tactile displacement magnitudes. As shown in Table~\ref{tab:wm_tactile_quality}, \WMName{} achieves the lowest 3DEPE of 0.12\,mm and ties for the lowest MSE at 0.035\,mm$^2$. Compared with $\mathcal{N}_0$-TWAM, it improves Shape Cosine from 0.74 to 0.80, indicating more accurate contact shapes. Temporal IoU increases from 0.59 to 0.76, while Temporal Corr. rises from 0.48 to 0.72,
demonstrating better temporal alignment with real contact in tactile signals. The lower VAE-output token count also highlights the compactness of our 3D tactile deformation fields compared with tactile RGB images. Detailed metric definitions and evaluation protocols are provided in Appendix~\ref{sec:appendix_wm_evaluation}.

We further assess visual fidelity using PSNR, SSIM, and LPIPS, and follow DreamGen Bench~\cite{jang2025dreamgen} to evaluate instruction following (IF) and physics alignment (PA). For IF, we separately report scores from Qwen3-VL-8B-Instruct~\cite{bai2025qwen3} and GPT-6.1 Terra-High. For PA, we average the physical plausibility scores from Qwen3-VL-8B-Instruct and VideoCon-Physics~\cite{bansal2025videophy}. As shown in Table~\ref{tab:wm_performance}, \WMName{} achieves the best PSNR, SSIM, and LPIPS among all evaluated models, exceeding Cosmos-Super by 3.14\,dB in PSNR. It also achieves the highest GPT-rated IF score of 100\%, while maintaining a PA score comparable to other leading models. These results show strong visual fidelity, instruction following, and physical plausibility.

\begin{table}[!t]
\centering
\small
\setlength{\tabcolsep}{2.5pt}
\renewcommand{\arraystretch}{1.1}
\caption{
Comparison of visuo-tactile world models on the spatial and temporal fidelity of generated 3D displacement fields. We also compare the average number of VAE-output tokens per case, where k denotes 1,000 tokens.
}
\label{tab:wm_tactile_quality}
\resizebox{\columnwidth}{!}{%
\begin{tabular}{@{}l*{6}{c}@{}}
\toprule
Method
& \shortstack{VAE tokens\\(k/case)}
& \shortstack{MSE\,$\scriptstyle\downarrow$\\(mm$^2$)}
& \shortstack{3DEPE\,$\scriptstyle\downarrow$\\(mm)}
& \shortstack{Shape\\Cosine\,$\scriptstyle\uparrow$}
& \shortstack{Temp.\\IoU\,$\scriptstyle\uparrow$}
& \shortstack{Temp.\\Corr.\,$\scriptstyle\uparrow$} \\
\midrule
\Nzero{}-TWAM & 25.96 & \textbf{0.035} & 0.13 & 0.74 & 0.59 & 0.48 \\
Dream-Tac & 317.99 & 0.083 & 0.32 & 0.58 & 0.41 & 0.37 \\
\midrule
\rowcolor{blue!5}
\WMName{} & \textbf{11.31} & \textbf{0.035} & \textbf{0.12} & \textbf{0.80} & \textbf{0.76} & \textbf{0.72} \\
\bottomrule
\end{tabular}%
}
\end{table}

\begin{table}[!t]
\centering
\small
\setlength{\tabcolsep}{2pt}
\renewcommand{\arraystretch}{1.1}
\caption{Visual generation quality and DreamGen Bench* evaluation. IF-Q and IF-G denote instruction following rated by Qwen3-VL-8B-Instruct~\cite{bai2025qwen3} and GPT-6.1 Terra-High, respectively. IF and PA are percentages.}
\label{tab:wm_performance}
\resizebox{\columnwidth}{!}{%
\begin{tabular}{@{}l*{7}{c}@{}}
\toprule
\multirow{2}{*}{Method}
& \multirow{2}{*}{Tactile}
& \multicolumn{3}{c}{Visual Quality}
& \multicolumn{3}{c}{DreamGen Bench*} \\
\cmidrule(lr){3-5}\cmidrule(lr){6-8}
& & PSNR\,$\scriptstyle\uparrow$
& SSIM\,$\scriptstyle\uparrow$
& LPIPS\,$\scriptstyle\downarrow$
& IF-Q\,$\scriptstyle\uparrow$
& IF-G\,$\scriptstyle\uparrow$
& PA\,$\scriptstyle\uparrow$ \\
\midrule
Cosmos-Nano & \ding{55} & 18.06 & 0.82 & 0.38 & 57.33 & 33.33 & 52.63 \\
Wan2.2-A14B & \ding{55} & 14.04 & 0.73 & 0.58 & 50.67 & 20.00 & \textbf{53.66} \\
WoW-1 & \ding{55} & 18.14 & 0.82 & 0.39 & 58.67 & 20.00 & 47.69 \\
Cosmos-Super & \ding{55} & 18.44 & 0.83 & 0.38 & 69.33 & 36.00 & 52.34 \\
MiniMax-H3 & \ding{55} & 14.94 & 0.69 & 0.52 & \textbf{81.33} & 86.67 & 53.48 \\
\midrule
\Nzero{}-TWAM & \ding{51} & 21.34 & 0.86 & 0.29 & 66.67 & 90.67 & 52.43 \\
Dream-Tac & \ding{51} & 19.23 & 0.79 & 0.42 & 37.33 & 6.67 & 16.24 \\
\midrule
\rowcolor{blue!5}
\WMName{} & \ding{51} & \textbf{21.58} & \textbf{0.87} & \textbf{0.24}
& 69.33 & \textbf{100.00} & 52.56 \\
\bottomrule
\end{tabular}%
}
\parbox{\columnwidth}{\footnotesize\textsuperscript{*} We follow the DreamGen Bench's procedure~\cite{jang2025dreamgen} with updated judges.}
\end{table}

\begin{figure*}[!t]
\centering
\includegraphics[width=.95\textwidth]{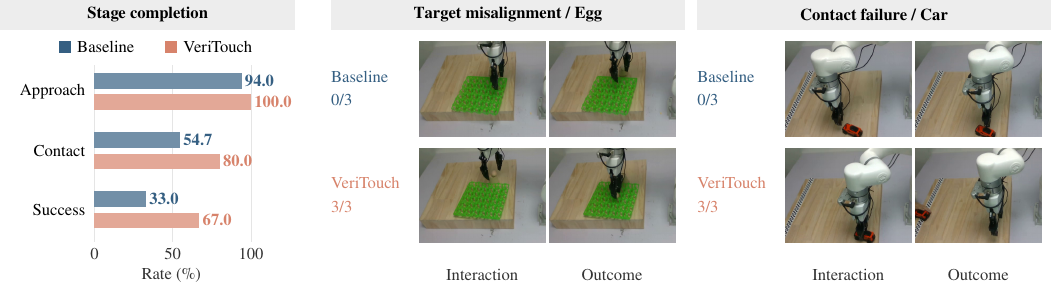}
\caption{\textbf{Stage-wise analysis of generalization gains.}
Left: Mean Approach, Contact, and Success scores across five tasks on unseen layouts.
Each score uses all 30 trials per task and condition. See Appendix~\ref{sec:app_policy_setup} (Stage-wise Scoring Protocol) for the scoring criteria.
Right: Representative examples of target misalignment in Egg Pick and failed grasping in Pullback Car.
The counts show successful trials out of three at each displayed layout.}
\label{fig:action_coverage}
\end{figure*}

\paragraph{Action Extraction Quality.}
\label{sec:idm_experiments}

To evaluate whether the extracted actions can be executed successfully, we replay them on a physical robot from aligned initial scenes and measure task success. For each task, we replay the actions extracted by each method once on each of the same 10 held-out real-world test trajectories. We compare \IDMName{} with EVA~\cite{Wang_2026_EVA}, DreamGen~\cite{jang2025dreamgen}, and a reimplementation of TACO's action prediction branch~\cite{Liu_2026_TACO}. All methods use the same camera views and trajectory-level split, with method-specific temporal contexts. Appendix~\ref{sec:idm_setup} provides the training configurations, and Appendix~\ref{sec:idm_realworld} describes the replay protocol.

\begin{table}[!t]
\centering
\small
\setlength{\tabcolsep}{2pt}
\caption{Real-world IDM replay success rates (\%).}
\label{tab:idm_realworld_success}
\begin{tabularx}{\columnwidth}{@{}>{\raggedright\arraybackslash}X*{6}{c}@{}}
\toprule
Method & \textit{Chip} & \textit{Egg} & \textit{Press} & \textit{Car} & \textit{Erase} & Avg. \\
\midrule
TACO\textsuperscript{*}~\cite{Liu_2026_TACO}
& 50\% & 90\% & 50\% & 70\% & 90\% & 70.0\% \\
DreamGen~\cite{jang2025dreamgen}
& 60\% & 90\% & 60\% & 70\% & 90\% & 74.0\% \\
EVA~\cite{Wang_2026_EVA}
& 80\% & 90\% & 60\% & \textbf{90\%} & \textbf{100\%} & 84.0\% \\
\rowcolor{blue!5}
\IDMName{}
& \textbf{90\%} & \textbf{100\%} & \textbf{100\%} & 80\% & \textbf{100\%} & \textbf{94.0\%} \\
 \bottomrule
\end{tabularx}
\par\smallskip
\parbox{\columnwidth}{\footnotesize\textsuperscript{*} We reimplemented TACO’s action prediction branch.}
\end{table}

As shown in Table~\ref{tab:idm_realworld_success}, \IDMName{} achieves the highest mean replay success rate across the five tasks (94\% vs. 84\% for EVA). Its largest gain over EVA is on \textit{charger press} (100\% vs. 60\%). The two methods tie on \textit{whiteboard erase}, while EVA leads on \textit{pullback car} (90\% vs. 80\%). Detailed offline reconstruction errors are reported in Appendix Table~\ref{tab:idm_comparison}.

\subsection{Understanding the Gains from Synthetic Data}
\paragraph{Effect of Multimodal Verification.}
Having compared verified and unfiltered synthetic data across four policies, we next examine whether tactile verification provides additional benefits beyond visual filtering. Using \Nzero{}-VTLA, we compare visual-only and full multimodal verification, with the unfiltered setting as a reference. We use the same candidate pool, training data size, and hyperparameters across all settings. As shown in Table~\ref{tab:verifier_policy_ablation}, adding tactile plausibility and visuo-tactile contact consistency checks increases average SR from 43.3\% to 51.7\%, compared with 33.3\% without verification. These results show that tactile-aware verification selects more effective training demonstrations than visual filtering alone.
\begin{table}[!htbp]
\centering
\caption{\textbf{Effect of multimodal verification on policy performance.}
We evaluate visual and tactile verification using \Nzero{}-VTLA (SR, \%).}
\label{tab:verifier_policy_ablation}
\footnotesize
\setlength{\tabcolsep}{2pt}
\begin{tabularx}{\columnwidth}{@{}>{\raggedright\arraybackslash}X*{8}{c}@{}}
\toprule
Method & Visual & Tactile & \textit{Chip} & \textit{Egg} & \textit{Press} & \textit{Car} & \textit{Erase} & Avg. \\
\midrule
& \ding{55} & \ding{55} & 40.0 & 50.0 & 23.3 & 33.3 & 20.0 & 33.3 \\
& \ding{51} & \ding{55} & 43.3 & 60.0 & 30.0 & 60.0 & 23.3 & 43.3 \\
\rowcolor{blue!5}
\cellcolor{white}\multirow{-3}{*}{\methodName{}} & \ding{51} & \ding{51} & \textbf{43.3} & \textbf{63.3} & \textbf{33.3} & \textbf{73.3} & \textbf{45.0} & \textbf{51.7} \\
\bottomrule
\end{tabularx}
\end{table}

\paragraph{Effect of Synthetic Tactile Data.}
We next investigate the contribution of synthetic tactile observations to downstream policy learning. Using \Nzero{}-VTLA, we replace the generated tactile observations in the verified demonstrations with no-contact inputs, while keeping the videos, actions, and sample counts unchanged. As shown in Table~\ref{tab:synthetic_tactile_ablation}, removing synthetic tactile observations reduces the average SR from 51.7\% to 26.0\%. This substantial drop highlights the importance of synthetic tactile observations in learning contact-rich manipulation, demonstrating that their contribution extends beyond the visual and action information provided by synthetic trajectories.

\begin{table}[!htbp]
\centering
\caption{\textbf{Effect of synthetic tactile observations on policy performance.}
Both settings use \Nzero{}-VTLA and the same verified demonstrations (SR, \%).}
\label{tab:synthetic_tactile_ablation}
\footnotesize
\setlength{\tabcolsep}{2pt}
\begin{tabularx}{\columnwidth}{@{}>{\raggedright\arraybackslash}X*{7}{c}@{}}
\toprule
Method & Synthetic Tactile & \textit{Chip} & \textit{Egg} & \textit{Press} & \textit{Car} & \textit{Erase} & Avg. \\
\midrule
& No-contact & 0.0 & 46.7 & 26.7 & 56.7 & 0.0 & 26.0 \\
\rowcolor{blue!5}
\cellcolor{white}\multirow{-2}{*}{\methodName{}} & Generated & \textbf{43.3} & \textbf{63.3} & \textbf{33.3} & \textbf{73.3} & \textbf{45.0} & \textbf{51.7} \\
\bottomrule
\end{tabularx}
\end{table}

\paragraph{Stage-wise Analysis of Generalization Gains.}
Figure~\ref{fig:action_coverage} compares the real-only baseline (1:0) with VeriTouch trained using four synthetic demonstrations per real demonstration (1:4). For each condition, we average the scores over all 30 trials per task and give the five tasks equal weight. Approach increases from 94.0\% to 100.0\%, Contact from 54.7\% to 80.0\%, and the task success score from 33.0\% to 67.0\%. In the Egg Pick example, the baseline reaches the vicinity of the egg but fails to align for grasping. In the Pullback Car example, it fails to secure the car. By contrast, VeriTouch successfully completes both tasks in all three trials at the corresponding layouts.

\section{Conclusion}
\label{sec:conclusion}

We presented \methodName{}, a framework that combines visuo-tactile generation, multimodal verification, and action extraction to produce synthetic demonstrations for robot policy learning. \methodName{} jointly generates visual and tactile sequences, filters unreliable trajectories through multimodal verification, and extracts corresponding actions without iterative policy refinement or simulation replay. Across five real-world tasks, verified synthetic data improves all four evaluated policies, whereas unfiltered data provides smaller and less consistent gains. Furthermore, synthetic data scaling leads to an overall increase in policy success rates as more demonstrations are generated, more than doubling the average success rate without collecting additional real-world data.
These findings establish verified visuo-tactile synthesis as an effective approach to scaling training data for contact-rich manipulation.
They further highlight that effective data scaling depends not only on generating more trajectories, but also on selecting reliable trajectories.

{\small
\bibliographystyle{ieeenat_fullname}
\bibliography{main}
}

\clearpage
\appendix
\onecolumn
\section*{Appendix Contents}
\phantomsection
\label{sec:appendix}
\begingroup
\small
\setlength{\parindent}{0pt}
\newcommand{\appcontentsline}[3]{\noindent\hspace*{#1}\hyperref[#2]{\ref*{#2}\quad #3}\nobreak\dotfill\pageref{#2}\par}
\vspace{0.55em}
\appcontentsline{0pt}{sec:appendix_wm_details}{VeriGen: Architecture, Training, and Evaluation}
\appcontentsline{1.5em}{sec:appendix_wm_architecture}{Architecture}
\appcontentsline{1.5em}{sec:app_wm_training}{Training}
\appcontentsline{1.5em}{sec:appendix_tactile_rgb_reconstruction}{Tactile RGB Reconstruction from Deformation}
\appcontentsline{1.5em}{sec:appendix_wm_evaluation}{Evaluation}
\vspace{0.8em}
\appcontentsline{0pt}{sec:verifier_quality}{VeriEval: Multimodal Verification}
\appcontentsline{1.5em}{sec:verifier_implementation}{Verification Procedure}
\appcontentsline{1.5em}{sec:app_verifier_evaluation}{Evaluation}
\appcontentsline{1.5em}{sec:verieval_training}{Training Hyperparameters}
\vspace{0.8em}
\appcontentsline{0pt}{sec:idm_training_details}{VeriIDM: Architecture, Training, and Evaluation}
\appcontentsline{1.5em}{sec:app_idm_architecture}{Architecture and Inference}
\appcontentsline{1.5em}{sec:idm_setup}{Training}
\appcontentsline{1.5em}{sec:app_idm_evaluation}{Evaluation}
\vspace{0.8em}
\appcontentsline{0pt}{sec:appendix_n0vtla}{Policy Learning and Additional Results}
\appcontentsline{1.5em}{sec:app_policy_setup}{Experimental Setup}
\appcontentsline{1.5em}{sec:app_policy_training}{Training Configurations}
\appcontentsline{1.5em}{sec:app_scaling_configuration}{Data Scaling}
\vspace{0.8em}
\appcontentsline{0pt}{sec:additional_related_work}{Additional Related Work}
\appcontentsline{1.5em}{sec:rw_vtla}{Visuo-Tactile-Language-Action Models}
\appcontentsline{1.5em}{sec:rw_idm}{Inverse Dynamics Models for Action Extraction}
\vspace{0.8em}
\appcontentsline{0pt}{sec:sim_env}{Simulation Environment and Data Collection}
\appcontentsline{1.5em}{sec:sim_env_1}{Real-to-Sim Scene Reconstruction}
\appcontentsline{1.5em}{sec:sim_env_2}{Task Definition}
\appcontentsline{1.5em}{sec:sim_env_3}{Visuo-Tactile Demonstration Collection}
\appcontentsline{1.5em}{sec:task_coverage}{Task Coverage Across Components}
\vspace{0.8em}
\appcontentsline{0pt}{sec:openvitac_eval}{End-to-End Pipeline Evaluation on OpenViTac}
\appcontentsline{1.5em}{sec:openvitac_eval_1}{Experimental Setup}
\appcontentsline{1.5em}{sec:openvitac_eval_2}{Pipeline Adaptation and Policy Training}
\appcontentsline{1.5em}{sec:openvitac_wm_qual}{Visuo-Tactile Generation Quality}
\appcontentsline{1.5em}{sec:openvitac_eval_3}{Task Scoring and Evaluation Protocol}
\appcontentsline{1.5em}{sec:openvitac_eval_4}{Downstream Policy Results}
\appcontentsline{1.5em}{sec:openvitac_configurations}{Training Configurations}
\vspace{0.8em}
\appcontentsline{0pt}{sec:visualization_comparison}{Visualization Comparison}
\appcontentsline{1.5em}{sec:real_eval_layouts}{Real-World Evaluation Layouts}
\appcontentsline{1.5em}{sec:realworld_qualitative_comparisons}{Real-World Qualitative Comparisons}
\appcontentsline{1.5em}{sec:openvitac_qualitative_comparisons}{OpenViTac Qualitative Comparisons}
\vspace{0.8em}
\endgroup
\twocolumn

\renewcommand{\topfraction}{0.95}
\renewcommand{\dbltopfraction}{0.95}
\renewcommand{\textfraction}{0.05}
\renewcommand{\floatpagefraction}{0.75}
\renewcommand{\dblfloatpagefraction}{0.75}
\setcounter{topnumber}{3}
\setcounter{bottomnumber}{2}
\setcounter{totalnumber}{5}
\setcounter{dbltopnumber}{3}
\setlength{\textfloatsep}{12pt plus 2pt minus 2pt}
\setlength{\dbltextfloatsep}{12pt plus 2pt minus 2pt}
\setlength{\floatsep}{8pt plus 2pt minus 2pt}
\setlength{\dblfloatsep}{8pt plus 2pt minus 2pt}
\setlength{\intextsep}{10pt plus 2pt minus 2pt}

\newsavebox{\appendixtablebox}
\newcommand{\appendixfittable}[2]{%
    \sbox{\appendixtablebox}{#2}%
    \ifdim\wd\appendixtablebox>#1\relax
        \resizebox{#1}{!}{\usebox{\appendixtablebox}}%
    \else
        \usebox{\appendixtablebox}%
    \fi
}

\makeatletter
\setlength{\@fptop}{0pt}
\setlength{\@fpsep}{12pt plus 2pt minus 2pt}
\setlength{\@fpbot}{0pt plus 1fil}
\setlength{\@dblfptop}{0pt}
\setlength{\@dblfpsep}{12pt plus 2pt minus 2pt}
\setlength{\@dblfpbot}{0pt plus 1fil}
\makeatother

\section{VeriGen: Architecture, Training, and Evaluation}
\label{sec:appendix_wm_details}
\newcommand{\wmexpertheading}[1]{\paragraph{#1.}}

\subsection{Architecture}
\label{sec:appendix_wm_architecture}

\WMName{} jointly generates main- and wrist-camera videos and bilateral tactile deformation sequences from initial camera images and a task caption. The task caption is the textual task instruction used to condition generation. It builds on the Mixture-of-Transformers (MoT) architecture of Cosmos3-Nano~\cite{agarwal2026cosmos}, retaining its pretrained video and text experts and introducing a tactile generation expert. Video and tactile features interact through multimodal joint attention during denoising. Initial tactile observations and future actions are not required.

\subsubsection{Dual-View Video Branch}
\label{sec:appendix_video_expert}

\paragraph{Cosmos3 Initialization and Inherited Architecture.}
\WMName{} uses the pretrained video and text experts of Cosmos3-Nano~\cite{agarwal2026cosmos} as its video--language backbone. The video expert retains the 36-layer Transformer, 4096-dimensional hidden representations, attention projections, and feed-forward networks (FFNs). We extend its input organization and attention computation to jointly process the main- and wrist-camera views. Both views share the video expert and retain distinct view identities and spatiotemporal coordinates. We refer to the two views as Main and Wrist below.

\paragraph{Independent Main/Wrist Encoding and Sequence Organization.}
The frozen Wan2.2 VAE~\cite{team2025wan} encodes Main and Wrist videos separately. At each latent timestep, we divide the 48-channel spatial grid into $2\times2$ patches and flatten each patch into a 192-dimensional vector. A shared linear projection maps these vectors to the video expert's 4096-dimensional hidden space.

For a dual-view sample, the input consists of a task-caption prefix followed by two video subsequences. Each video subsequence contains clean first-frame tokens followed by noisy future tokens:

\begin{equation}
\label{eq:app_video_token_sequence}
\begin{aligned}
\mathcal S=[&\mathcal S_{\mathrm{text}};
\mathcal C_{\mathrm{Main}};\mathcal F_{\mathrm{Main}}^\rho;\mathcal C_{\mathrm{Wrist}};\mathcal F_{\mathrm{Wrist}}^\rho].
\end{aligned}
\end{equation}

Here, $\mathcal S_{\mathrm{text}}$ denotes the task-caption tokens, $\mathcal C_c$ denotes the clean first-frame tokens of view $c$, and $\mathcal F_c^\rho$ denotes its future tokens at denoising noise level $\rho$. Semicolons indicate concatenation along the token dimension. The frozen text expert processes the caption prefix, and the shared video expert processes both video subsequences.

\paragraph{View Identity and Dual-View Attention.}
We assign Main and Wrist view IDs 0 and 1, respectively. A learned view embedding is added to every video token after the input projection:

\begin{equation}
\label{eq:app_video_view_identity}
h_i^{(c)}=W_{\mathrm{in}}p_i^{(c)}+\mathbf e_c,
\qquad c\in\{0,1\}.
\end{equation}

Here, $p_i^{(c)}$ is patch $i$ from view $c$, $W_{\mathrm{in}}$ is the shared input projection, and $\mathbf e_c$ is the view embedding. First-frame and future tokens from the same view receive the same embedding.

Video tokens attend to the task-caption prefix and to tokens within and across views. Within-view attention applies rotary positional encoding (RoPE) along the temporal, height, and width dimensions. Cross-view attention applies temporal RoPE only, since the two cameras have different spatial coordinate systems. Tokens at the same sequence time receive the same temporal coordinate. Attention scores over caption tokens, within-view tokens, and cross-view tokens are normalized jointly.

For wrist-only samples from OmniViTac, the video expert processes the caption prefix and the Wrist subsequence, following the single-view handling in Sec.~\ref{sec:appendix_data_preprocessing}.

\paragraph{First-Frame, Text, and Noise Conditioning.}
The initial images provide the visual conditions. Future latents from each view receive independent Gaussian noise at a shared noise level $\rho$, following the formulation in Sec.~\ref{sec:appendix_flow_matching_objectives}.

Following Cosmos3, the frozen text expert processes the caption prefix using causal attention. Video tokens from both views access its keys and values, while text tokens do not attend to video tokens. The caption features are therefore independent of the noisy video inputs. Visual--tactile joint attention subsequently passes task information from the video expert to the tactile expert.


\paragraph{Handling Different Resolutions.}
We construct the token sequence separately for each view's resolution. The following examples use the full 16-second generation configuration. Image sizes are reported as width $\times$ height, and latent tensors follow the $[C,T,H,W]$ convention for channels, time, height, and width.

A $640\times480$ video produces latents of shape $[48,61,30,40]$. Patchification yields $15\times20=300$ tokens per latent timestep and 18,300 tokens in total.

For Main videos recorded at $1280\times720$, we resize the images to $768\times432$ while preserving the aspect ratio. The VAE produces latents of shape $[48,61,27,48]$. We append one zero row along the latent height dimension before $2\times2$ patchification, producing a $28\times48$ grid. This yields $14\times24=336$ tokens per timestep and 20,496 tokens in total. The added row is removed when restoring the output.

The branch supports a $768\times432$ Main view paired with a $640\times480$ Wrist view, two $640\times480$ views, and wrist-only samples. These configurations share model parameters and temporal coordinates. Each view retains its own spatial grid, token count, and spatial positional encoding.

\paragraph{Dual-View Prediction Outputs.}
A shared output projection predicts a 192-dimensional flow-velocity vector for each video patch. We rearrange these predictions into a latent tensor for each view and remove any rows added for spatial patchification. Each predicted velocity tensor therefore has the same shape as its target video latent.

During generation, the two views are denoised jointly with their initial visual conditions fixed, then decoded separately by the frozen video VAE. The loss and joint inference settings are detailed in Secs.~\ref{sec:appendix_flow_matching_objectives} and~\ref{sec:appendix_optimization_inference}, respectively.

\subsubsection{Tactile Representation and Generation}
\label{sec:app_contents_5}

\paragraph{TactileVAE: 3D Displacement Representation and Design Motivation.}
\label{sec:appendix_tactile_vae}
Following OmniVTA~\cite{zheng2026omnivta}, we represent tactile surface deformation using 3D marker displacements. This displacement field has a much lower spatial resolution than tactile RGB images, providing a compact input for tactile modeling. Our TactileVAE follows OmniVTA's design of a spatio-temporal encoder and an implicit deformation decoder.

\paragraph{Spatio-Temporal Encoder.}
Each sensor provides 3D marker displacements at 700 uniformly distributed locations. We arrange each observation as a $35\times20\times3$ grid, with three channels for marker displacement along the $x$, $y$, and $z$ axes.

The encoder consists of a projection-in layer implemented with causal 3D convolution, residual blocks, spatial and temporal downsampling, and a projection-out layer that predicts the posterior parameters. Spatial downsampling reduces the grid from $35\times20$ to $7\times4$, while two temporal downsampling modules with stride 2 reduce the sequence length by a factor of four. For a single-sensor sequence of $T$ frames after length alignment, the tensor shapes are

\begin{equation}
\label{eq:app_tvae_encoder_shape}
[T,700,3]\rightarrow[3,T,35,20]\rightarrow[64,T/4,7,4].
\end{equation}

Latent tensors use the $[C,T,H,W]$ convention for channels, time, height, and width. Each tactile latent timestep contains 28 spatial locations with 64 features per location. The encoder predicts a posterior mean and log variance. Training uses reparameterized samples, and reconstruction evaluation uses the posterior mean.

The left and right sensors share the encoder and decoder and are encoded separately. A 16-second sequence sampled at 60 Hz contains 960 frames and produces a $[64,240,7,4]$ latent for each sensor. The world model orders the resulting representations by sensor side, with the left side before the right.

\paragraph{Implicit Deformation Decoder.}
Following OmniVTA, we use an implicit neural representation (INR) decoder to model the tactile deformation field as a continuous function of spatial coordinates. Given a spatial query point $\mathbf x\in[-1,1]^2$ and latent feature map $z$, it reconstructs the displacement sequence as

\begin{equation}
\label{eq:app_tvae_decoder}
\widehat{\mathbf d}_{1:T}(\mathbf x)
=D_\theta\!\left(\gamma(\mathbf x),\Phi(z,\mathbf x)\right),
\end{equation}

where $D_\theta$ is the implicit decoder, $\Phi(z,\mathbf x)$ retrieves local features from the latent feature map by bilinear spatial interpolation, and $\gamma(\mathbf x)$ is the positional encoding. The decoder upsamples the local feature sequence from $T/4$ to $T$ timesteps, adds the positional encoding, and applies a fully connected residual block and an output projection to predict the 3D displacement sequence at the query location.

During training, we uniformly sample 700 continuous spatial query points on the tactile surface and obtain their ground-truth displacements by interpolating the recorded marker displacements. At reconstruction time, querying the original marker coordinates produces an output of shape $[T,700,3]$.

\paragraph{Training Loss.}
We retain the query-point reconstruction loss and KL regularization used in OmniVTA and introduce spatial gradient matching and temporal difference matching supervision. The reconstruction loss uses continuous spatial query points; spatial and temporal supervision use the original marker grid. Let $\mathbf d(\mathbf x_q)$ and $\widehat{\mathbf d}(\mathbf x_q)$ denote the ground-truth and reconstructed displacements at a sampled query point $\mathbf x_q$. The reconstruction loss is

\begin{equation}
\label{eq:app_tvae_reconstruction}
\mathcal L_{\mathrm{rec}}
=\operatorname{MSE}_{\Omega_q}
\left(\widehat{\mathbf d}(\mathbf x_q),\mathbf d(\mathbf x_q)\right).
\end{equation}

Here, $\operatorname{MSE}_{\Omega_q}$ averages squared error over valid query-point displacement elements, including sequence positions and displacement components. For posterior mean $\mu$ and log variance $\eta=\log\sigma^2$, the KL regularizer is

\begin{equation}
\label{eq:app_tvae_kl}
\mathcal L_{\mathrm{KL}}
=\frac{1}{2|\Omega_z|}
\sum_{i\in\Omega_z}
\left(\mu_i^2+\exp(\eta_i)-1-\eta_i\right),
\end{equation}

where $\Omega_z$ contains the valid latent elements in the target interval.

To account for scale differences across displacement components, we compute training-set means $\mu_{s,k}$ and standard deviations $\sigma_{s,k}$ separately for each sensor and displacement component, and standardize as
\[
\begin{gathered}
\widetilde X^{s}_{t,p,k}
=\frac{X^{s}_{t,p,k}-\mu_{s,k}}
{\max(\sigma_{s,k},10^{-8})},\\
s\in\{\mathrm{L},\mathrm{R}\},\qquad k\in\{x,y,z\}.
\end{gathered}
\]
Here, $t$ and $p$ index frames and markers, respectively.

Spatial supervision matches reconstructed and ground-truth displacement differences between neighboring markers on the original $35\times20$ grid. Let $\widehat{\widetilde{\mathbf X}}$ denote the reconstruction of the standardized field $\widetilde{\mathbf X}$. The operators $\Delta_x$ and $\Delta_y$ denote neighboring differences along the horizontal and vertical grid axes. The spatial loss is

\begin{equation}
\label{eq:app_tvae_spatial}
\begin{aligned}
\mathcal L_{\mathrm{spatial}}
={}&\frac12\operatorname{MAE}_{\Omega_x}
\left(\Delta_x\widehat{\widetilde{\mathbf X}},\Delta_x\widetilde{\mathbf X}\right)\\
&+\frac12\operatorname{MAE}_{\Omega_y}
\left(\Delta_y\widehat{\widetilde{\mathbf X}},\Delta_y\widetilde{\mathbf X}\right).
\end{aligned}
\end{equation}

Here, $\operatorname{MAE}_{\Omega}$ averages absolute error over valid elements. The sets $\Omega_x$ and $\Omega_y$ contain valid horizontal and vertical marker pairs, respectively, across all target frames and all three displacement components.

Temporal supervision matches local temporal differences in the standardized displacement fields at the native sampling rate. Using $t$ to index tactile observations, we define

\begin{equation}
\label{eq:app_tvae_temporal}
\mathcal L_{\mathrm{temporal}}
=\operatorname{MSE}_{\Omega_\Delta}
\left(
\widehat{\widetilde{\mathbf X}}_{t+1}-\widehat{\widetilde{\mathbf X}}_t,
\widetilde{\mathbf X}_{t+1}-\widetilde{\mathbf X}_t
\right).
\end{equation}

The set $\Omega_\Delta$ contains valid displacement elements from pairs of adjacent observations.

The total training objective is

\begin{equation}
\label{eq:app_tvae_total}
\mathcal L_{\mathrm{TactileVAE}}
=\mathcal L_{\mathrm{rec}}
+\beta\mathcal L_{\mathrm{KL}}
+\lambda_s\mathcal L_{\mathrm{spatial}}
+\lambda_\Delta\mathcal L_{\mathrm{temporal}},
\end{equation}

where $\beta=10^{-6}$ and $\lambda_s=\lambda_\Delta=1$.

The reconstruction and auxiliary losses share the same latent sample. Reconstruction and spatial supervision use valid target frames, while temporal supervision uses real, valid transitions. For an interior training window, we also supervise the difference from its preceding frame when that frame is available. At the trajectory origin, no preceding transition is constructed. Each loss is normalized by its valid elements, and padded positions are excluded. Training data and optimization settings are given below, and loss ablations are reported in Sec.~\ref{sec:appendix_tvae_loss_ablation}.

\paragraph{Training Details.}

We train TactileVAE for 70 epochs on the left and right tactile sequences from 12,512 OmniViTac trajectories~\cite{zheng2026omnivta}. These tactile data form the VAE training set; the world model uses 16,593 OmniViTac trajectories for public-data pretraining (Sec.~\ref{sec:appendix_data_preprocessing}). In each epoch, we sample 16 windows per sensor from every trajectory, giving both sides equal training exposure.

Each window contains 32 target frames. For interior windows, the encoder reads up to 128 preceding observation frames and the decoder reads up to 20 preceding latent positions. Only the target interval contributes to the loss. At the trajectory origin, both modules use zero initial states.

We use AdamW with a learning rate and weight decay of $10^{-4}$, an effective batch size of 512 windows, and gradient clipping at 1.0. Reconstruction, spatial gradient, and temporal difference errors are evaluated on an independent validation set every 2,000 updates. The encoder and decoder from epoch 70 are frozen during world model training.

\paragraph{Spatiotemporal Tactile Expert.}
\label{sec:appendix_tactile_expert}
\WMName{} uses a dedicated tactile expert with 2048-dimensional hidden representations to predict flow velocities in the tactile latent space. Its architecture follows the spatiotemporal Diffusion Transformer (DiT) design used in OmniVTA's world model~\cite{zheng2026omnivta,peebles2023scalable}. The expert contains 18 spatial and 18 temporal attention layers arranged in spatial--temporal pairs, matching the video expert's total depth of 36 layers. The left and right sensors share the Transformer backbone and use separate input projections, output projections, and sensor-side identities.

\paragraph{Bilateral Tokenization and Input Conditioning.}
For each sensor, TactileVAE produces a latent of shape $[64,T_z,7,4]$, where $T_z$ is the number of latent timesteps. We flatten the spatial grid into 28 tokens per timestep, each with 64 features. After adding noise, separate left- and right-side projections map these tokens to the tactile expert's hidden space. Each timestep contains 56 tokens, ordered by left and then right sensor.

We add learned sensor-side embeddings and side-specific row and column embeddings to identify each token's source and grid location. Both sides also receive a shared tactile modality embedding.

The tactile tokens receive an embedding of the same denoising noise level $\rho$ used by the video branch. Initial camera images and the task caption provide conditions through joint attention with the video expert. Generation requires no initial tactile observation.

\paragraph{Spatial Attention.}
At each latent timestep, spatial attention jointly processes all 56 tokens across the left and right $7\times4$ latent grids. We use standard multi-head attention with height- and width-axis RoPE applied to the queries and keys, followed by an output projection and a residual connection.

\paragraph{Temporal Attention.}
Temporal attention operates independently at each latent grid location of each sensor, producing 28 sequences per sensor and 56 in total. Each sequence uses bidirectional attention with temporal RoPE based on the preprocessed sequence time. Spatial and temporal attention use separate parameters. Sequence time is distinct from the denoising noise level $\rho$; its alignment with video time is defined in Sec.~\ref{sec:appendix_joint_attention}. Joint attention is inserted after temporal attention and before the following FFN.

\paragraph{Initialization and Flow-Velocity Prediction.}
The tactile expert is initialized independently, with zero-initialized residual output projections and final prediction heads. The left and right features pass through their respective heads to predict latent flow velocities. Each prediction retains the spatiotemporal layout of its input latent. The flow-matching objective is detailed in Sec.~\ref{sec:appendix_flow_matching_objectives}.

\subsubsection{Visual--Tactile Joint Attention}
\label{sec:appendix_joint_attention}

\paragraph{Joint Attention Computation.}
\WMName{} uses multimodal joint attention to exchange information between the video and tactile experts. The use of attention for cross-expert interaction is informed by T-Rex~\cite{niu2026t} and VT-WAM~\cite{tian2026vt}. Our joint generation model uses bidirectional attention between video and tactile tokens. We insert 18 joint attention modules, pairing video layers $2,4,\ldots,36$ with the 18 temporal attention layers of the tactile expert. Each connected layer completes its own attention computation before passing its features to joint attention.

Each modality uses its own normalization and projection layers to produce queries, keys, and values with a common projection width of 2048, divided into 32 attention heads with 64 dimensions per head. Main and Wrist share the video-side projections, and the two tactile sensors share the tactile-side projections. For each sample, we concatenate valid tokens in the order of Main, Wrist, left tactile, and right tactile, and compute full joint attention:

\begin{equation}
\label{eq:app_multimodal_joint_attention}
[O_v;O_t]
=\operatorname{softmax}\!\left(
\frac{[Q_v;Q_t][K_v;K_t]^\top}{\sqrt{d_h}}
\right)[V_v;V_t].
\end{equation}

Here, $v$ and $t$ denote the video and tactile modalities, respectively. The tensors $Q$, $K$, and $V$ denote queries, keys, and values; semicolons indicate concatenation along the token dimension; and $d_h$ is the attention-head dimension. Queries and keys include the temporal positional encoding defined below. Attention is normalized over all valid tokens, allowing the video expert to access tactile information and the tactile expert to access visual context.

We split the outputs into video and tactile features, project each back to its expert's hidden dimension, and apply residual updates:

\begin{equation}
\label{eq:app_joint_attention_residual}
H_v'=H_v+P_v(O_v),\qquad
H_t'=H_t+P_t(O_t),
\end{equation}

where $H_v$ and $H_t$ are the input features and $P_v$ and $P_t$ are the modality-specific output projections. The updated features are then passed to the FFNs of their respective experts.

\paragraph{TPS-Based Temporal Alignment.}
Video and tactile sequences are sampled at 15 Hz and 60 Hz, respectively. After temporal compression, their latent timesteps still represent different time intervals. Following the temporal steps per second (TPS) scaling in Cosmos3~\cite{agarwal2026cosmos}, we assign both modalities a common time coordinate system for RoPE in joint attention.

For modality $m\in\{v,t\}$, let $f_m$ denote its sampling rate and $c_m$ its VAE temporal compression ratio. The latent TPS and the coordinate increment between successive latent timesteps are

\begin{equation}
\label{eq:app_tps_alignment}
\mathrm{TPS}_m=\frac{f_m}{c_m},\qquad
\Delta p_m=\frac{\mathrm{TPS}_{\mathrm{base}}}{\mathrm{TPS}_m}.
\end{equation}

We retain the Cosmos3 base value $\mathrm{TPS}_{\mathrm{base}}=6$. Both VAEs have a temporal compression ratio of 4, giving $\mathrm{TPS}_v=3.75$ and $\mathrm{TPS}_t=15$, with coordinate increments of 1.6 and 0.4, respectively.

The initial visual observation defines time zero, and its video latent has coordinate 0. Future video latents have coordinates $1.6,3.2,\ldots$, and tactile latents have coordinates $0.4,0.8,\ldots$. The $i$-th future video latent therefore shares coordinate $1.6i$ with the $4i$-th tactile latent. All spatial tokens within a latent timestep receive the same time coordinate. This provides a common time reference for both camera views and both tactile sensors while retaining their respective temporal resolutions.

\paragraph{Valid Tokens and Missing Views.}
Joint attention uses the computation masks defined in Sec.~\ref{sec:appendix_data_preprocessing} and processes the available views together with both tactile sensors. Each view retains its own token count, while all streams use the common temporal coordinates defined above.

\subsection{Training}
\label{sec:app_wm_training}

\subsubsection{Data Construction and Preprocessing}
\label{sec:appendix_data_preprocessing}

\paragraph{Data Sources and Splits.}
\WMName{} uses public real-world trajectories and task-aligned simulated interactions for pretraining, followed by fine-tuning on real-robot demonstrations from the target tasks.

\paragraph{Pretraining.}
We first pretrain the model on 16,593 trajectories from OmniViTac~\cite{zheng2026omnivta}, spanning grasping, assembly, adjustment, cutting, peeling, and wiping. Since many OmniViTac videos contain human hands due to the data collection setup, we then adapt the model on a mixture of 3,056 trajectories from the Gello robot subset of OmniViTac and 3,275 task-aligned visuo-tactile simulation trajectories. Training batches sample the two sources at a $1{:}1$ ratio. The resulting pretrained weights initialize fine-tuning on the five target tasks.

\paragraph{Fine-tuning.}
We fine-tune the pretrained model on five real-robot tasks: \textit{chip pick}, \textit{egg pick}, \textit{whiteboard erase}, \textit{pullback car}, and \textit{charger press}. Each task contributes 50 trajectories, split into 40 training, 5 validation, and 5 test trajectories. Across the five tasks, this gives 200 training, 25 validation, and 25 test trajectories for the world model. All observations from a trajectory remain in the same split.

\paragraph{Task-Level Captions.}
We use GPT-5.6-Sol with medium reasoning effort to generate task-level captions from manipulation videos. The captions describe visible entities, manipulation steps, and final states. Public-data trajectories of the same task share a caption. Simulation captions are matched to the task version and manipulation process, then checked and revised against reference trajectories.

Each caption contains three fields: the entities field describes the manipulating agents and objects, the actions field lists manipulation steps in temporal order, and the temporal caption field describes the complete interaction. We preserve field and list order, serialize the full caption as JSON text, and pass it through the Cosmos3 tokenizer to the frozen text expert.

\paragraph{Temporal Preprocessing.}
The initial visual observation defines the start of each generation sequence. For world-model training, video observations and bilateral tactile deformation sequences are sampled at 15 Hz and 60 Hz, respectively, for a maximum duration of 16 seconds. Both video views retain their first frames as conditions. Tactile targets start at the first subsequent sampling time, and observations across streams are selected using their timestamps.

For target real-robot trajectories longer than 16 seconds, we uniformly sample video frames over the full interaction and apply the same time mapping to both tactile sequences. This compresses the interaction into a 16-second generation interval. Shorter trajectories retain their natural duration and use the sampling rates above. Positional encoding uses this preprocessed sequence time, which is a scaled version of recording time for compressed trajectories. The same mapping is applied to both camera views and both tactile sensors.

Both VAEs compress time by a factor of four. Before encoding, we repeat the last observation to extend each future-video or tactile sequence to a multiple of four, adding at most three observations. The terminal latent produced from this extension is included in the valid encoded length and participates in attention and supervision. This last-observation repetition is distinct from the zero-padding used to batch encoded sequences.

Let $N_v$ denote the number of valid future video frames, excluding the initial condition, and $N_t$ the number of valid tactile observations. Both are counted before last-observation repetition. The valid latent lengths are

\begin{equation}
\label{eq:app_valid_encoded_lengths}
L_v=1+\left\lceil\frac{N_v}{4}\right\rceil,
\qquad
L_t=\left\lceil\frac{N_t}{4}\right\rceil,
\end{equation}

where the additional video latent timestep corresponds to the first-frame condition.

\paragraph{Resolution and Batching.}
Wrist images use a resolution of $640\times480$. Main images either retain $640\times480$ or are resized from $1280\times720$ to $768\times432$ while preserving the aspect ratio. The corresponding latent grids and patchification are described in Sec.~\ref{sec:appendix_video_expert}. Table~\ref{tab:app_wm_view_configurations} summarizes view configurations in the public-data pretraining set; image resolutions are reported as width $\times$ height.

\begin{table}[!tb]
\centering
\caption{View configurations in the public-data pretraining set. Image resolutions are given as width $\times$ height.}
\label{tab:app_wm_view_configurations}
\small
\setlength{\tabcolsep}{3pt}
\begin{tabular*}{\columnwidth}{@{\extracolsep{\fill}}lccr@{}}
\toprule
Configuration & Main & Wrist & \shortstack{Training\\trajectories} \\
\midrule
Dual & $640\times480$ & $640\times480$ & 3,056 \\
Dual & $768\times432$ & $640\times480$ & 3,960 \\
Wrist only & --- & $640\times480$ & 9,577 \\
\bottomrule
\end{tabular*}
\end{table}

During public-data pretraining, batches group trajectories by view configuration and traverse samples in random order within each group, giving trajectories approximately equal training exposure. During Gello--simulation adaptation, batches use the $1{:}1$ source mixture while retaining each sample's available views and spatial grids.

OmniViTac includes wrist-only trajectories, so we also support samples without a Main view. These samples contain the caption prefix, the Wrist video subsequence, and bilateral tactile inputs. Wrist retains view ID 1 and its corresponding embedding; no tokens are added for the missing Main view.

\paragraph{Validity Masks.}
We distinguish positions used in attention from positions used in the loss. Both masks are determined by valid latent lengths and broadcast over spatial locations and channels. Indexing the first video latent frame by 0, the computation and supervision masks for a view with valid length $L_v$ are

\begin{equation}
\label{eq:app_video_validity_masks}
\begin{aligned}
M_{\mathrm{comp}}^v(j)&=\mathbf 1[0\le j<L_v],\\
M_{\mathrm{sup}}^v(j)&=\mathbf 1[1\le j<L_v].
\end{aligned}
\end{equation}

The first video latent remains clean, participates in attention, and is excluded from the loss. Valid future video latents participate in both attention and supervision. For tactile latents, the computation and supervision masks both cover all $L_t$ valid timesteps.

After encoding, shorter sequences are zero-padded to capacities of 61 video and 240 tactile latent timesteps. Both masks are zero at these padding positions, which are excluded from attention. Within the tactile expert, padded positions remain zero after residual updates. Observations with zero tactile deformation remain valid; validity is determined by sequence length, not displacement magnitude. Wrist-only samples use masks for Wrist and both tactile sensors.

\subsubsection{Joint Flow-Matching Objective}
\label{sec:appendix_flow_matching_objectives}

\paragraph{Flow-Matching Formulation.}
We adopt rectified flow matching~\cite{liu2022flow} to jointly train video and tactile generation in latent space. For modality $m\in\{v,t\}$, where $v$ and $t$ denote video and touch, respectively, we construct noisy target latents by interpolating between the standardized clean target $z_0^m$ and Gaussian noise $\epsilon^m\sim\mathcal N(0,I)$:

\begin{equation}
\label{eq:app_joint_flow_targets}
z_\rho^m=(1-\rho)z_0^m+\rho\epsilon^m.
\end{equation}

We use mode sampling as in Cosmos3~\cite{agarwal2026cosmos} and apply a rectified-flow shift reparameterization with shift 5 to obtain $\rho\in[0,1]$. Within each training sample, video and tactile latents share the same $\rho$, while each camera view and tactile sensor receives independently sampled noise. At $\rho=0$, the target latents are clean; at $\rho=1$, they contain only noise.

Each expert predicts the latent flow velocity $\hat u^m$ from the noisy latents at noise level $\rho$, conditioned on the initial visual observations and task caption. The target flow velocity is

\begin{equation}
\label{eq:app_target_flow_velocity}
u^{m,*}=\epsilon^m-z_0^m.
\end{equation}

Tactile targets are obtained by separately encoding the left and right displacement sequences with the frozen TactileVAE. Noising and supervision use the supervision masks defined in Sec.~\ref{sec:appendix_data_preprocessing}.

\paragraph{Video Generation Loss.}
Let $\mathcal V$ denote the views present in a sample, and let $M^v_{a,i}$ be the supervision mask for scalar latent element $i$ of view $a$. With predicted and target velocities denoted by $\hat u^v_{a,i}$ and $u^{v,*}_{a,i}$, respectively, the video loss is

\begin{equation}
\label{eq:app_video_flow_loss}
\mathcal L_v
=\frac{1}{|\mathcal V|}\sum_{a\in\mathcal V}
\frac{\sum_i M^v_{a,i}
\left(\hat u^v_{a,i}-u^{v,*}_{a,i}\right)^2}
{\sum_i M^v_{a,i}}.
\end{equation}

We average squared velocity error over valid future elements within each view, then average the view-level losses equally. Views with different resolutions therefore receive the same loss weight.

\paragraph{Tactile Generation Loss.}

Following OmniVTA~\cite{zheng2026omnivta}, we construct amplitude and dynamics weights from the $L_2$ norms of ground-truth displacement vectors and their differences between consecutive observations, respectively. Both quantities are clipped to $[0,1]$, with the first-frame temporal difference set to zero. The weights are computed before displacement standardization, while the flow-velocity prediction error is computed in standardized latent space.

We map the clipped values to the tactile latent grid by aggregating over four successive observations and the corresponding $5\times5$ marker region. Mean aggregation yields the amplitude weight $w_{\mathrm{amp}}$, and root-mean-square aggregation yields the dynamics weight $w_{\mathrm{dyn}}$.

For sensor $s$, latent timestep $j$, and grid location $p$, let $e_{j,p}^s=\|\hat u_{j,p}^{t,s}-u_{j,p}^{t,s,*}\|_2^2/64$ denote the velocity error averaged over the 64 latent channels. The tactile loss is

\begin{equation}
\label{eq:app_tactile_flow_loss}
\begin{aligned}
\mathcal L_t=\frac{1}{N_{\mathrm{valid}}^t}\sum_{s,j,p}M_j^t
\Bigl(&1+\lambda_{\mathrm{dyn}}w_{\mathrm{dyn},j,p}^s\\
&+\lambda_{\mathrm{amp}}w_{\mathrm{amp},j,p}^s\Bigr)e_{j,p}^s,
\end{aligned}
\end{equation}

where $M_j^t$ is the tactile supervision mask and $N_{\mathrm{valid}}^t=2\times7\times4\sum_j M_j^t$ counts valid positions across both sensors. The base term supervises every valid position. The amplitude and dynamics terms increase the contribution of larger displacements and more rapid changes. All terms use the same valid-position normalization.

\paragraph{Joint Objective and Supervision.}
The joint training objective is the weighted sum of the video and tactile flow-matching losses:

\begin{equation}
\label{eq:app_joint_flow_objective}
\mathcal L_{\mathrm{WM}}
=\lambda_v\mathcal L_v+\lambda_t\mathcal L_t.
\end{equation}

The coefficients $\lambda_v$ and $\lambda_t$ control the contributions of the two modalities. We set $\lambda_v=\lambda_t=1$ and $\lambda_{\mathrm{dyn}}=\lambda_{\mathrm{amp}}=1$. Each modality loss is normalized within a sample as defined above, then averaged over the batch.

\subsubsection{Training and Inference}
\label{sec:appendix_optimization_inference}

\paragraph{Initialization and Trainable Parameters.}
The video expert is initialized from Cosmos3-Nano, while the tactile expert and joint attention layers are initialized independently. We train these modules while keeping the video VAE, pretrained TactileVAE, and text expert frozen.

\paragraph{Optimization Configuration.}
We train on 24 NVIDIA A800 GPUs with a global batch size of 24. Pretraining comprises an OmniViTac stage and a subsequent Gello--simulation adaptation stage, both using peak learning rates of $2\times10^{-5}$ for the video expert and $4\times10^{-5}$ for the tactile expert and joint attention layers. Target-task fine-tuning uses $1\times10^{-5}$ and $2\times10^{-5}$ for these parameter groups, respectively.


The OmniViTac stage uses a learning-rate schedule defined over 34,570 steps with 3,457 warmup steps. We use its step-9,000 exponential moving average (EMA) checkpoint to initialize Gello--simulation adaptation. Adaptation runs for 4,000 updates, followed by 1,000 fine-tuning updates on the five real-robot tasks. Adaptation and fine-tuning run their full schedules. At each stage transition, we load the preceding stage's EMA weights. We use the final fine-tuning EMA weights for evaluation and data generation. Table~\ref{tab:wm_optimization_configuration} summarizes these optimization settings.

\begin{table}[!tb]
\centering
\caption{Optimization configuration of \WMName{}. Slash-separated values correspond to the OmniViTac and Gello--simulation stages of pretraining, followed by target-task fine-tuning. Checkpoint steps are counted separately within each stage.}
\label{tab:wm_optimization_configuration}
\footnotesize
\setlength{\tabcolsep}{3pt}
\renewcommand{\arraystretch}{1.12}
\begin{tabularx}{\columnwidth}{@{}>{\raggedright\arraybackslash}p{0.33\columnwidth}>{\raggedright\arraybackslash}X@{}}
\toprule
Setting & Configuration \\
\midrule
Optimizer & AdamW ($\beta_1=0.9$, $\beta_2=0.95$, $\epsilon=10^{-6}$, weight decay $=0$) \\
Checkpoint step & 9,000 / 4,000 / 1,000 steps \\
Peak LR, video & $2\times10^{-5}$ / $2\times10^{-5}$ / $1\times10^{-5}$ \\
Peak LR, tactile/joint & $4\times10^{-5}$ / $4\times10^{-5}$ / $2\times10^{-5}$ \\
LR schedule & Linear warmup + cosine decay \\
Warmup & 3,457 / 100 / 50 steps \\
LR schedule horizon & 34,570 / 4,000 / 1,000 steps \\
Global batch size & 24 \\
EMA decay cap & 0.999 \\
\bottomrule
\end{tabularx}
\end{table}

\paragraph{Joint Inference.}
Given the initial Main and Wrist images and a task caption, we jointly generate videos and bilateral tactile deformation using UniPC~\cite{zhao2023unipc} with 20 sampling steps and a noise-level shift of 5. \WMName{} outputs Main and Wrist videos at 15 Hz and bilateral tactile deformation sequences at 60 Hz. The frozen VAEs decode the generated latents, and the deformation fields are rendered as tactile RGB images for verification (Sec.~\ref{sec:appendix_tactile_rgb_reconstruction}).

\subsection{Tactile RGB Reconstruction from Deformation}
\label{sec:appendix_tactile_rgb_reconstruction}

\WMName{} predicts a sequence of 3D displacement fields for each fingertip on a $35\times20$ sensor grid. To render a tactile image, we use the reference surface mesh for the corresponding sensor from the Xense SDK~\cite{xense_sdk}. For each frame, we express the predicted field relative to its reference frame and add the resulting 700 displacement vectors to the corresponding mesh vertices. We align the deformed mesh with the Photon renderer's coordinate frame and convert its coordinates from millimeters to centimeters before rendering. Repeating this process independently for each fingertip yields two tactile RGB sequences with the same timestamps as the predicted fields.

\VerifierName{} uses these images for the masked reconstruction check with AnyTouch2~\cite{Feng_2026_AnyTouch2} and for force estimation with FeelAnyForce~\cite{Shahidzadeh_2025_FeelAnyForce}. The contact timing check operates directly on the 3D displacement fields (Appendix~\ref{sec:verifier_implementation}).

\subsection{Evaluation}
\label{sec:appendix_wm_evaluation}

\subsubsection{Evaluation Setup and Baseline Implementations}
\label{sec:appendix_wm_baselines}

To compare generation quality as fairly as possible, we evaluate all methods on the same 25 test trajectories from the five real-world tasks. Each method generates three rollouts per trajectory using the same task caption and initial Main/Wrist images. For video baselines without native support for separate view conditions, we follow DreamGen~\cite{jang2025dreamgen} and concatenate the two initial images into a single conditioning image. The generated views are separated for evaluation.

Cosmos-Nano (16B), Cosmos-Super (64B)~\cite{agarwal2026cosmos}, Wan2.2-A14B~\cite{team2025wan}, WoW-1 (14B)~\cite{chi2025wow}, and MiniMax-H3 (33B)~\cite{minimax2026h3} are evaluated zero-shot using their released weights. We adapt $\mathcal{N}_0$-TWAM~\cite{team2026n_0} and Dream-Tac~\cite{lou2026dream} following their original papers and public implementations, retaining their core architectures, joint prediction objectives, and chunkwise generation mechanisms while using the same initial observations for conditioning. 
We initialize $\mathcal{N}_0$-TWAM from its publicly released base checkpoint \texttt{n0-twam-base}, which has undergone large-scale visuo-tactile pretraining on NeoData~\cite{neoteaiteam2026n0foundationagetactileintelligence}. The NeoData contains over 30,000 hours and approximately 1.4 million episodes across six embodiments and 450 tasks.
We initialize Dream-Tac from the official pretrained Cosmos-Predict2-2B Video2World~\cite{nvidia2025cosmospredict2} checkpoint.
As with \WMName{}, both models are trained for 4,000 updates on the Gello--simulation training data and then fine-tuned for 1,000 updates on the same 200 real-robot training trajectories. Both use a global batch size of 24 and retain their respective optimization settings.

We compute video and tactile metrics at each model's native output frequencies for these modalities. Following the frame-rate alignment approach in Appendix B of PISA Experiments~\cite{li2025pisa}, we pair each predicted observation with ground truth at the same timestamp on the shared preprocessed sequence time axis (Sec.~\ref{sec:appendix_data_preprocessing}). For trajectories longer than 16 seconds, this axis compresses the full interaction into 16 seconds using the same time mapping for both camera views and both tactile sensors; shorter trajectories retain their recording-time intervals. For the visuo-tactile models, both video and tactile metrics for $\mathcal{N}_0$-TWAM are computed at its native output rate of 10 Hz. Dream-Tac retains its chunkwise prediction scheme with an action horizon of $H=20$. The prediction endpoints of full chunks are spaced $4/3$ seconds apart on this shared sequence time axis, aligning exactly with the sampling times of the 15 Hz video and 60 Hz tactile ground truth. Video and tactile metrics are computed at these native prediction endpoints, using the corresponding ground-truth observations.

All rollouts are evaluated before verification filtering. We compare only valid future observations, excluding conditioning frames and padding. Each metric is first averaged over the three rollouts of a test trajectory and then equally across all test trajectories.

\subsubsection{Generation Quality Metrics}
\label{sec:app_contents_12}

\paragraph{Visual Metrics.}
We evaluate visual fidelity using PSNR, SSIM, and LPIPS against ground-truth videos. Scores are averaged over valid future frames within each view and then equally across Main and Wrist.

\paragraph{DreamGen Bench* Evaluation.}
Following DreamGen Bench~\cite{jang2025dreamgen}, we retain its frame sampling procedures and evaluation prompts, using updated VLM judges and adding a description of the horizontally concatenated Main/Wrist layout to the Qwen/GPT prompts. For instruction following (IF), Qwen3-VL-8B-Instruct~\cite{bai2025qwen3} and GPT-6.1 Terra-High receive 49 and 8 sampled frames, respectively, along with the task caption, and output a binary judgment of whether the full task was completed. For physics alignment (PA), Qwen3 processes the video at 2 fps, while VideoCon-Physics~\cite{bansal2025videophy} samples 32 frames using its original loader; neither component receives the task caption. PA equally averages Qwen3's binary score and VideoCon-Physics' continuous Yes probability (normalized over Yes and No). IF and PA are reported as percentages.

\paragraph{Tactile Metrics.}
\label{sec:appendix_wm_tactile_metrics}
We compare generated and ground-truth 3D displacement fields on the original sensor grids in millimeters. For methods that generate tactile RGB images, we recover the displacement fields using the sensor SDK. Deformation magnitude maps use the same initial reference for each sensor.

\begin{itemize}
\setlength{\itemsep}{2pt}
\item \textbf{MSE:} Mean squared displacement error over components, markers, and future observations, reported in mm$^2$.
\item \textbf{3DEPE:} Mean Euclidean distance between predicted and ground-truth displacement vectors over markers and future observations, reported in mm.
\item \textbf{Shape Cosine:} Cosine similarity between the full deformation magnitude maps after translating them to align their squared-magnitude-weighted centroids with bilinear interpolation. We evaluate times when both predicted and reference observations contain contact.
\item \textbf{Temporal IoU:} Intersection-over-union of predicted and reference contact times. Contact requires an eight-connected region of at least eight markers above 0.12 mm with a peak above 0.20 mm, and thresholds are fixed on training data. The same contact criterion is used for Shape Cosine. We exclude a sensor sequence from the Temporal IoU average if both its predicted and reference contact sets are empty.
\item \textbf{Temporal Correlation (Temporal Corr.):} Zero-lag Pearson correlation between the predicted and \mbox{ground-truth} spatially averaged displacement-magnitude curves, measuring how deformation intensity varies over time.
\end{itemize}

We compute tactile metrics separately for the two sensors and average their scores.During real-world data collection, different executions of the same manipulation can produce tactile responses at different locations on the sensor surface. We analyze 35 selected tasks, including all five tasks in our real data and five tasks from each of OmniViTac's six manipulation categories, covering 5,662 trajectories with valid peak centroids on both sensors.

\begin{figure}[!htbp]
\centering
\begin{minipage}{\columnwidth}
\centering
\captionsetup{hypcap=false}
\captionof{table}{Cross-episode peak-centroid differences in marker-grid rows and columns. Each row reports the median across five task medians; entries give left / right sensor values. Episode counts include trajectories with defined centroids on both sensors.}
\label{tab:peak_centroid_axis_offsets}
\begingroup
\footnotesize
\setlength{\tabcolsep}{2pt}
\renewcommand{\arraystretch}{1.12}
\begin{tabular*}{\columnwidth}{@{\extracolsep{\fill}}lccc@{}}
\toprule
Group & Episodes & \shortstack{Row\\difference\\Left / Right} & \shortstack{Column\\difference\\Left / Right} \\
\midrule
Our real data & 250 & 3.82 / 3.86 & 3.29 / 3.37 \\
OmniViTac's Grasping & 495 & 5.34 / 5.12 & 1.00 / 1.63 \\
OmniViTac's Assembly & 497 & 3.02 / 2.63 & 1.19 / 1.93 \\
OmniViTac's Adjustment & 1489 & 4.36 / 4.26 & 0.92 / 1.15 \\
OmniViTac's Cutting & 391 & 5.65 / 5.56 & 1.96 / 1.53 \\
OmniViTac's Peeling & 1599 & 0.84 / 0.85 & 0.87 / 0.82 \\
OmniViTac's Wiping & 941 & 1.97 / 2.36 & 1.90 / 1.80 \\
\bottomrule
\end{tabular*}
\endgroup
\par\vspace{7pt}
\includegraphics[width=\columnwidth]{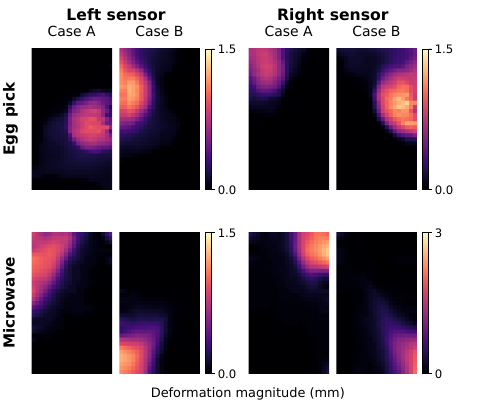}
\captionof{figure}{Bilateral peak deformation-magnitude maps for two successful demonstrations of each task. Each sensor is shown at its own energy peak, with a scale shared across its two cases.}
\label{fig:appendix_contact_shape_peaks}
\end{minipage}
\end{figure}

For each sensor in each trajectory, we select the frame with maximum full-field deformation energy and compute its centroid, weighting each marker by its squared deformation magnitude. We compare absolute row and column differences between peak centroids of trajectories from the same task and collection. Table~\ref{tab:peak_centroid_axis_offsets} shows larger row differences in the Grasping and Cutting groups, while the Peeling group summaries remain below one grid interval along both axes.

Figure~\ref{fig:appendix_contact_shape_peaks} further shows two real egg-picking demonstrations and two real microwave-opening demonstrations, with successful task completion confirmed from the recorded videos. The peak deformation responses occupy widely separated regions on both sensors, showing that successful executions of the same task can have noticeably different contact-response locations. Differences in contact-response location therefore do not necessarily indicate task failure. Based on this observation, we introduce Shape Cosine as a complement to displacement errors in the original sensor coordinates, measuring the similarity of deformation-magnitude distributions after centroid alignment.

\subsubsection{Initial-Frame Collection and Rollout Generation}
\label{sec:appendix_initial_frame_collection}

\paragraph{Initial-Frame Collection.}
To provide more synthetic demonstrations for downstream policy training, we collect paired Main/Wrist images of different initial object layouts for the five real-world tasks. Each image pair is combined with the corresponding task caption to condition \WMName{}. We record only the initial scene and avoid the test layouts used for downstream real-world policy evaluation. In total, we collect 1,287 image pairs: 286 for \textit{chip pick}, 299 for \textit{egg pick}, 310 for \textit{charger press}, 100 for \textit{pullback car}, and 292 for \textit{whiteboard erase}.

For each pair of initial images, we sample multiple candidate trajectories using different random seeds, jointly generating Main/Wrist videos and bilateral tactile deformation sequences. Generation duration is set for each task, with video at 15 Hz and tactile deformation at 60 Hz. \VerifierName{} screens the generated trajectories, and \IDMName{} assigns action labels to the retained trajectories to construct synthetic demonstrations for policy training.

\subsubsection{Tactile Representation and Encoding Efficiency}
\label{sec:appendix_encoding_efficiency}

\paragraph{Encoding Efficiency.}
To verify the compactness and encoding efficiency of the 3D displacement representation described above, we compare VAE output token counts on the same 25 real-world test trajectories. We use 60 Hz tactile RGB sequences temporally aligned with the displacement fields and count tokens according to the tactile VAE configurations of $\mathcal{N}_0$-TWAM and Dream-Tac. The original tactile RGB images have a shape of $700\times400\times3$ (height $\times$ width $\times$ channels). Following each method's preprocessing, we resize the images to $128\times128\times3$ and $224\times224\times3$, corresponding to Wan2.2 VAE and Wan2.1 VAE, respectively. Our TactileVAE encodes displacement fields of shape $35\times20\times3$. The three paths correspond to mean VAE output token counts of 25.96k, 317.99k, and 11.31k per trajectory, respectively, showing that displacement fields have a more compact encoded representation for the same temporal coverage.

We further compare the encoding overhead of tactile RGB with Wan2.2 VAE and 3D displacement with TactileVAE. Under the same NVIDIA A800, FP32 precision, and batch size of 1, the mean VAE encoding times per complete bilateral sequence are 62.28 seconds and 0.077 seconds, respectively. The lower tactile token count and encoding overhead support our choice of 3D displacement fields for tactile modeling.

\subsubsection{Performance Across Training Checkpoints}
\label{sec:appendix_wm_stage_ablation}

\paragraph{World Model Adaptation.}
Table~\ref{tab:wm_stage_ablation} compares the model after public-data pretraining, Gello--simulation adaptation, and target-task fine-tuning, following the training procedure in Sec.~\ref{sec:appendix_optimization_inference}. Temporal contact agreement improves across these steps, with the full model also achieving the best visual fidelity.

\begin{table*}[!tbp]
\centering
\caption{World model performance across training checkpoints. The first two rows correspond to the two pretraining stages. IF-Q and IF-G use Qwen3-VL-8B-Instruct and GPT-6.1 Terra-High, respectively; IF and PA are reported as percentages. PA follows the evaluation protocol in Sec.~\ref{sec:app_contents_12}.}
\label{tab:wm_stage_ablation}
\footnotesize
\setlength{\tabcolsep}{2pt}
\renewcommand{\arraystretch}{1.15}
\appendixfittable{\textwidth}{%
\begin{tabular}{@{}l*{11}{c}@{}}
\toprule
\multirow{2}{*}{Checkpoint} & \multicolumn{3}{c}{Visual Quality} & \multicolumn{5}{c}{Tactile Quality} & \multicolumn{3}{c}{IF and PA} \\
\cmidrule(lr){2-4}\cmidrule(lr){5-9}\cmidrule(lr){10-12}
 & PSNR $\uparrow$ & SSIM $\uparrow$ & LPIPS $\downarrow$ & MSE $\downarrow$ & 3DEPE $\downarrow$ & \shortstack{Shape Cosine $\uparrow$} & \shortstack{Temporal IoU $\uparrow$} & \shortstack{Temporal Corr. $\uparrow$} & IF-Q $\uparrow$ & IF-G $\uparrow$ & PA $\uparrow$ \\
\midrule
Pretraining & 17.72 & 0.79 & 0.39 & 0.04 & 0.18 & 0.72 & 0.37 & 0.30 & 56.00 & 34.67 & 53.24 \\
Adaptation & 17.60 & 0.82 & 0.36 & 0.05 & 0.17 & 0.63 & 0.64 & 0.63 & 61.33 & 81.33 & \textbf{53.87} \\
Fine-tuning (Full) & \textbf{21.58} & \textbf{0.87} & \textbf{0.24} & \textbf{0.035} & \textbf{0.12} & \textbf{0.80} & \textbf{0.76} & \textbf{0.72} & \textbf{69.33} & \textbf{100.00} & 52.56 \\
\bottomrule
\end{tabular}%
}
\end{table*}

Gello--simulation adaptation improves instruction following over public-data pretraining, while higher Temporal IoU and Temporal Corr. indicate closer agreement with reference contact timing and deformation dynamics. These temporal gains are accompanied by a decrease in Shape Cosine from 0.72 to 0.63. Target-task fine-tuning then raises Shape Cosine to 0.80 while further improving both temporal metrics. The full model also gives the best visual fidelity, lowest pointwise displacement errors, and highest IF scores under both judges. PA is highest after Gello--simulation adaptation.

\subsubsection{TactileVAE Loss Ablation}
\label{sec:appendix_tvae_loss_ablation}

Table~\ref{tab:tactile_vae_loss_ablation} compares the two supervision terms introduced in Sec.~\ref{sec:app_contents_5}. Baseline uses reconstruction loss and KL regularization. Spatial and Temporal add spatial gradient matching and temporal difference matching, respectively; Ours uses both. All four variants share the same architecture, training data, normalization, and training budget of 70 epochs. We evaluate all four variants using posterior means on a separate test set of 308 trajectories, comprising 616 sensor streams.

Reconstruction RMSE compares the reconstructed and ground-truth displacements, temporal difference RMSE compares their displacement changes between adjacent frames, and spatial gradient RMSE compares their displacement differences between adjacent markers. Contact interval IoU measures the agreement between the ground-truth and reconstructed signals in contact timing.
 We use the 95th percentile of marker displacement magnitudes as the per-frame contact signal. For each sensor sequence, we estimate a noise cutoff from the median and median absolute deviation of low-amplitude ground-truth values. We set the contact threshold to the larger of this cutoff and 5\% of the ground-truth signal's peak, and apply it to the ground truth and all reconstructions.
 
\begin{table}[!tb]
\centering
\caption{TactileVAE loss ablation on 3D displacement fields. \mbox{RMSEs} use the original displacement units. Spatial gradient RMSE is averaged over sensor streams. Best values for each metric before rounding are bold.}
\label{tab:tactile_vae_loss_ablation}
\footnotesize
\setlength{\tabcolsep}{2pt}
\renewcommand{\arraystretch}{1.15}
\begin{tabularx}{\columnwidth}{@{}>{\raggedright\arraybackslash}Xrrrr@{}}
\toprule
Metric & Baseline & Spatial & Temporal & Ours \\
\midrule
Reconstruction RMSE $\downarrow$ & 0.0105 & 0.0099 & 0.0098 & \textbf{0.0090} \\
Temporal difference RMSE $\downarrow$ & 0.0073 & 0.0079 & 0.0058 & \textbf{0.0058} \\
Spatial gradient RMSE $\downarrow$ & 0.0100 & 0.0089 & 0.0097 & \textbf{0.0085} \\
Contact interval IoU $\uparrow$ & 0.8997 & 0.9092 & 0.9012 & \textbf{0.9245} \\
\bottomrule
\end{tabularx}
\end{table}

\paragraph{Effect of Spatial and Temporal Supervision.}
Spatial supervision reduces spatial gradient error but increases temporal difference error relative to Baseline, whereas temporal supervision improves temporal accuracy. Combining both terms achieves the lowest reconstruction and spatial gradient RMSEs and the highest contact interval IoU, while retaining the temporal accuracy of Temporal.

\subsubsection{Qualitative Comparisons across Methods}
\label{sec:appendix_cross_method_qualitative}

\paragraph{Cross-Method Comparisons.}
Task-wise qualitative comparisons are provided in Appendix~\ref{sec:realworld_qualitative_comparisons} (\mbox{Figs.~\ref{fig:realworld_chip_pick_qual}--\ref{fig:realworld_whiteboard_erase_qual}}). We compare \WMName{} with Cosmos-Nano and Cosmos-Super~\cite{agarwal2026cosmos}, Wan2.2-A14B~\cite{team2025wan}, WoW-1~\cite{chi2025wow}, MiniMax-H3~\cite{minimax2026h3}, Dream-Tac~\cite{lou2026dream}, and $\mathcal{N}_0$-TWAM~\cite{team2026n_0}. Each figure combines all methods for one real-world task at four aligned target times, showing ground-truth (GT) and generated Main/Wrist views together with bilateral tactile fields where available.

\section{VeriEval: Multimodal Verification}
\label{sec:verifier_quality}
\subsection{Verification Procedure}
\label{sec:verifier_implementation}
\paragraph{Visual verifier.}
We collect robot manipulation videos from online sources, extract their first frames, and caption them with GPT-5.6. Conditioned on these image--caption pairs, video generation models including Cosmos3~\cite{agarwal2026cosmos} and MiniMax-H3~\cite{minimax2026h3} generate approximately 900 videos of varying quality. Human annotators label whether each video contains violations and give the reasons. We use these annotations for full-parameter supervised fine-tuning of Qwen3.5-9B. Given a video and its caption, the fine-tuned model produces pass or fail judgments and the corresponding reasons along seven dimensions that cover instruction following and physical plausibility. The seven criteria are as follows.

\noindent\textbf{R1. Scene Entity Continuity.}
Tracks whether the same robots, components, and objects persist over time, without unexplained additions, losses, or identity swaps.

\noindent\textbf{R2. Shape and Deformation Realism.}
Assesses whether the visible shape, scale, and deformation of each body agree with its structural and material properties.

\noindent\textbf{R3. Mechanical Connectivity and Articulation.}
Part motion must respect the axes and ranges allowed by its joints or attachments, and separation must have a physically credible cause.

\noindent\textbf{R4. Spatial Exclusivity of Solids.}
Detects impossible passage through solid matter, such as a gripper crossing an intact object or an object sinking through a rigid tabletop.

\noindent\textbf{R5. Manipulation--Response Coupling.}
Checks whether the robot's interaction plausibly explains when
the object starts or stops moving and how its motion changes.

\noindent\textbf{R6. Motion and Stability Realism.}
Assesses whether an object remains at rest or moves as expected from its support, applied forces, and prior motion, including during falls, collisions, and changes in velocity.

\noindent\textbf{R7. Instruction Following.}
Compares the depicted manipulation with the instruction's specified objects, action, and any explicitly required outcome.

\begin{table}[!t]
\centering
\caption{Visual verifier quality on 100 human-annotated videos, reporting micro- and macro-averaged precision (P), recall (R), F1, overall accuracy (Acc.), and total-score Pearson $r$. P, R, F1, and Acc. are percentages.}
\label{tab:verifier}
\small
\setlength{\tabcolsep}{1.2pt}
\renewcommand{\arraystretch}{1.10}
\begin{tabular*}{\linewidth}{@{\extracolsep{\fill}}llrrrrrr@{}}
\toprule
\multirow{2}{*}{Method} & \multirow{2}{*}{Average}
& \multicolumn{3}{c}{\shortstack{Classification\\(\%) $\uparrow$}}
& \multirow{2}{*}{\shortstack{Acc.\\(\%) $\uparrow$}}
& \multirow{2}{*}{\shortstack{Pearson\\$r$ $\uparrow$}}
& \multirow{2}{*}{\shortstack{Valid\\outputs}} \\
\cmidrule(lr){3-5}
& & P & R & F1 & & & \\
\midrule
\multirow{2}{*}{Qwen3.5-9B} & Micro & 0.00 & 0.00 & 0.00 & \multirow{2}{*}{0.00} & \multirow{2}{*}{N/A} & \multirow{2}{*}{2/100} \\
 & Macro & 0.00 & 0.00 & 0.00 &  &  &  \\
\addlinespace[2pt]
\multirow{2}{*}{Qwen3.5-27B} & Micro & 50.84 & 95.14 & 66.27 & \multirow{2}{*}{51.00} & \multirow{2}{*}{0.1894} & \multirow{2}{*}{99/100} \\
 & Macro & 50.90 & 95.14 & 66.26 &  &  &  \\
\addlinespace[2pt]
\multirow{2}{*}{Qwen3.6-27B} & Micro & 51.18 & \textbf{99.43} & 67.57 & \multirow{2}{*}{52.29} & \multirow{2}{*}{0.3638} & \multirow{2}{*}{100/100} \\
 & Macro & 51.32 & \textbf{99.43} & 67.64 &  &  &  \\
\addlinespace[2pt]
\multirow{2}{*}{Qwen3.8-27B} & Micro & 29.79 & 4.00 & 7.05 & \multirow{2}{*}{4.00} & \multirow{2}{*}{0.3229} & \multirow{2}{*}{9/100} \\
 & Macro & 31.34 & 4.00 & 6.97 &  &  &  \\
\addlinespace[2pt]
\multirow{2}{*}{GPT-6 Astra} & Micro & 59.79 & 96.00 & 73.68 & \multirow{2}{*}{65.71} & \multirow{2}{*}{0.6565} & \multirow{2}{*}{100/100} \\
 & Macro & 61.07 & 96.00 & 74.09 &  &  &  \\
\addlinespace[2pt]
\multirow{2}{*}{GPT-5.6 Sol} & Micro & 55.76 & 96.86 & 70.77 & \multirow{2}{*}{57.00} & \multirow{2}{*}{0.4749} & \multirow{2}{*}{97/100} \\
 & Macro & 56.35 & 96.86 & 70.92 &  &  &  \\
\addlinespace[2pt]
\multirow{2}{*}{GPT-5.5} & Micro & 54.90 & 96.00 & 69.85 & \multirow{2}{*}{58.57} & \multirow{2}{*}{0.4385} & \multirow{2}{*}{100/100} \\
 & Macro & 55.27 & 96.00 & 69.90 &  &  &  \\
\addlinespace[2pt]
\multirow{2}{*}{\textbf{Ours}} & Micro & \textbf{71.64} & 82.29 & \textbf{76.60} & \multirow{2}{*}{\textbf{74.86}} & \multirow{2}{*}{\textbf{0.7951}} & \multirow{2}{*}{100/100} \\
 & Macro & \textbf{76.22} & 82.29 & \textbf{77.07} &  &  &  \\
\bottomrule
\end{tabular*}
\end{table}

\paragraph{Tactile verifier.}
The tactile verifier applies three checks in sequence, and a sample is retained only if it passes all of them. The first check addresses contact consistency. A VLM that we train determines from wrist-camera images whether the robot is in contact with the object, which yields the visual contact intervals. The tactile contact intervals are derived from the $L_2$ norm of the tactile data, and samples whose two sets of intervals are misaligned are discarded. The second check addresses the validity of the tactile signals. We randomly mask regions of the tactile RGB images, reconstruct them with the frozen AnyTouch2~\cite{Feng_2026_AnyTouch2} encoder and decoder, and discard samples with high reconstruction error. The third check addresses force plausibility. FeelAnyForce~\cite{Shahidzadeh_2025_FeelAnyForce} predicts three-dimensional forces from the tactile RGB images. The mean and maximum normal-force magnitudes $|F_z|$, together with the task caption, are given to Qwen3.5-4B, which judges whether the force magnitude is plausible.

\begin{table}[!t]
\centering
\caption{Tactile verifier quality on the constructed evaluation set. All metrics are percentages.}
\label{tab:app_tactile_verifier}
\small
\setlength{\tabcolsep}{2pt}
\begin{tabular*}{\linewidth}{@{\extracolsep{\fill}}lcccc@{}}
\toprule
Method & Precision $\uparrow$ & Recall $\uparrow$ & F1 $\uparrow$ & Accuracy $\uparrow$ \\
\midrule
TACO-style~\cite{Liu_2026_TACO} & 50.83 & 100.00 & 67.40 & 51.63 \\
VITAL-style~\cite{Wu_2026_ViTaL} & 49.77 & 94.78 & 65.27 & 49.57 \\
\midrule
\textbf{Ours} & 72.33 & 92.61 & \textbf{81.22} & \textbf{78.59} \\
\bottomrule
\end{tabular*}
\end{table}

\subsection{Evaluation}
\label{sec:app_verifier_evaluation}
\begin{table*}[!t]
\centering
\begin{minipage}[t]{\columnwidth}
\centering
\caption{Full-parameter fine-tuning configuration of the visual verifier.}
\label{tab:visual_verifier_finetuning}
\footnotesize
\setlength{\tabcolsep}{3pt}
\renewcommand{\arraystretch}{1.12}
\begin{tabularx}{\columnwidth}{@{}>{\raggedright\arraybackslash}p{0.31\columnwidth}>{\raggedright\arraybackslash}X@{}}
\toprule
\textbf{Parameter} & \textbf{Setting} \\
\midrule
Base VLM & Qwen3.5-9B \\
Optimizer & AdamW ($\beta_1=0.9$, $\beta_2=0.999$, $\epsilon=10^{-8}$) \\
Weight decay & 0.01 \\
Gradient clipping & 1.0 \\
Peak LR, language backbone & $3\times10^{-6}$ \\
Peak LR, vision encoder and projector & $3\times10^{-8}$ \\
Learning-rate schedule & 15 warm-up steps, then cosine decay to zero \\
Effective batch size & 32 videos (1 per GPU, 8 accumulation steps) \\
Hardware & $4\times$ NVIDIA H100 (80\,GB) \\
Training budget & 6 epochs (180 steps) \\
\bottomrule
\end{tabularx}
\end{minipage}\hfill
\begin{minipage}[t]{\columnwidth}
\centering
\nextfloat 
\caption{Full-parameter fine-tuning configuration of the wrist-view contact detector.}
\label{tab:wrist_contact_finetuning}
\footnotesize
\setlength{\tabcolsep}{3pt}
\renewcommand{\arraystretch}{1.12}
\begin{tabularx}{\columnwidth}{@{}>{\raggedright\arraybackslash}p{0.31\columnwidth}>{\raggedright\arraybackslash}X@{}}
\toprule
\textbf{Parameter} & \textbf{Setting} \\
\midrule
Base VLM & Qwen3.5-4B \\
Optimizer & AdamW ($\beta_1=0.9$, $\beta_2=0.999$, $\epsilon=10^{-8}$) \\
Weight decay & 0.01 \\
Gradient clipping & 1.0 \\
Peak LR, language backbone & $1\times10^{-5}$ \\
Peak LR, vision encoder and projector & $1\times10^{-6}$ \\
Learning-rate schedule & 109 warm-up steps, then cosine decay to zero \\
Effective batch size & 32 frames (2 per GPU, 16 accumulation steps) \\
Hardware & $1\times$ NVIDIA RTX PRO 6000 Blackwell (96\,GB) \\
Training budget & 3 epochs (2,178 steps) \\
\bottomrule
\end{tabularx}
\end{minipage}
\end{table*}

\subsubsection{Visual Verifier Evaluation}
\label{sec:visual_verifier_evaluation}
We evaluate on 100 videos with human-annotated binary scores for each of
seven rubrics. We compare each rubric prediction with its human label and
report micro- and macro-averaged precision, recall, and F1, together with
accuracy over all 700 rubric judgments. Micro averaging pools the judgments,
while macro averaging weights the seven rubrics equally. We also compute
Pearson correlation between predicted and human total scores, obtained by
summing the seven rubric scores for each video.
Table~\ref{tab:verifier} shows that our verifier achieves the highest
micro-F1 (76.60\%), macro-F1 (77.07\%), accuracy (74.86\%), and Pearson
correlation ($r=0.7951$), indicating closer agreement with human annotations.

\subsubsection{Tactile Verifier Evaluation}
\label{sec:tactile_verifier_evaluation}
Because the three checks operate as a single cascade, we evaluate them jointly: a sample counts as accepted only if it passes all three.
We construct 920 samples from 460 valid robot trajectories across eight tasks,
each paired with a corrupted counterpart. We apply four types of tactile
perturbations---spurious contact, contact dropout, spatial noise, and amplitude
scaling---with 115 samples per type, keeping wrist-camera videos and captions
unchanged. Valid samples are the positive class.

We compare with two adapted baselines. TACO-style~\cite{Liu_2026_TACO}
rejects samples whose estimated three-axis forces or temporal force changes
exceed the global ranges observed in the 460 valid trajectories.
VITAL-style~\cite{Wu_2026_ViTaL} measures cosine similarity between AnyTouch2
tactile embeddings and caption-derived phase descriptions, retaining samples
with positive aggregated scores.
In Table~\ref{tab:app_tactile_verifier}, our verifier achieves 81.22\% F1 and
78.59\% accuracy, exceeding the strongest baseline by 13.82 and 26.96
percentage points, respectively.

\subsection{Training Hyperparameters}
\label{sec:verieval_training}
\paragraph{Visual verifier.}
The visual verifier is a video-conditioned rubric scorer. Given a single-view video,
a conditioning image, and a task caption, it generates a reason and a binary decision for each
of seven criteria. We use 951 training examples, with videos sampled at 5\,fps and capped at
128 frames per example. Images are resized to $640\times480$. Fine-tuning updates the entire
Qwen3.5-9B model, including its visual and language components.

Table~\ref{tab:visual_verifier_finetuning} summarizes the optimization settings. The vision
encoder and multimodal projector use one hundredth of the language backbone's learning rate.
Training runs for six epochs on four H100 GPUs with an effective batch size of 32 videos, using
linear warm-up followed by cosine decay.

\paragraph{Wrist-view contact detector.}
We formulate contact detection as single-frame binary token prediction. Given a wrist-camera
image and a caption specifying the manipulated object, Qwen3.5-4B uses its existing
language-model head to predict a binary answer token. A positive label indicates that at least
one fingertip touches or holds the target. We update all parameters, including the vision
encoder, multimodal projector, language backbone, and tied input and output embeddings, and
minimize cross-entropy over the valid answer tokens.

Training uses more than 23K frames from 368 episodes. As detailed in
Table~\ref{tab:wrist_contact_finetuning}, the visual components use one tenth of the language
backbone's learning rate. Training runs on one GPU with an effective batch size of 32 frames, and
the three-epoch schedule uses linear warm-up and cosine decay.


\section{VeriIDM: Architecture, Training, and Evaluation}
\label{sec:idm_training_details}
\subsection{Architecture and Inference}
\label{sec:app_idm_architecture}
\begin{figure*}[!t]
\centering
    \captionsetup{hypcap=false}
    \includegraphics[width=\textwidth]{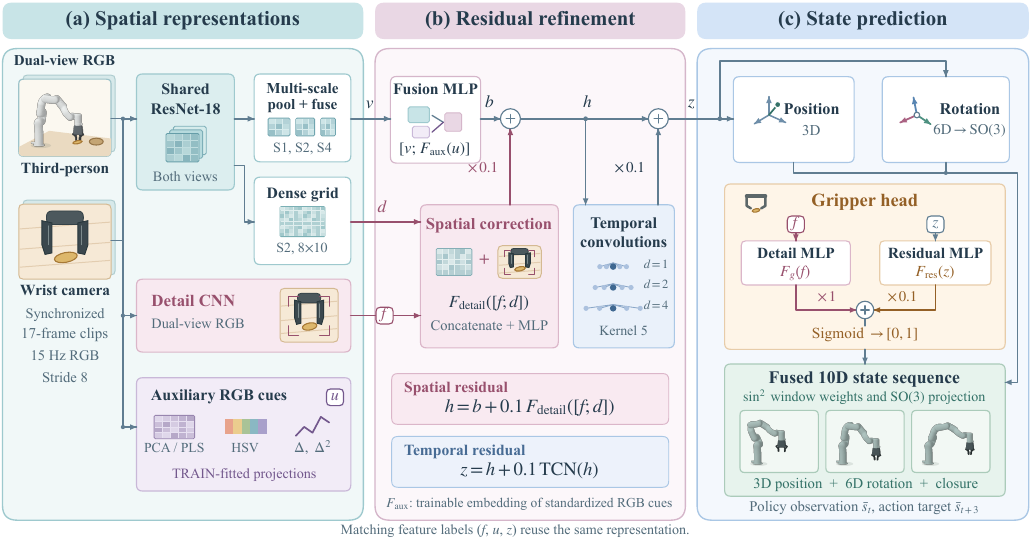}
    \caption{Overview of \IDMName{}. Dual-view spatial features
    undergo spatial and temporal residual refinement to predict pose and
    gripper closure. The gripper head combines a direct detail-based logit
    with a shared-feature correction.}
    \label{fig:idm_architecture}
\end{figure*}

\paragraph{Design motivation.}
Recovering robot trajectories from generated videos requires estimating the end-effector pose in the workspace and determining when and how far the gripper closes. One approach regresses a single action from a short RGB context and uses spatial softmax to map each feature channel to a two-dimensional coordinate~\cite{Wang_2026_EVA}. The coordinates preserve the locations of salient responses, while the underlying feature maps contain finer geometric cues such as visible gripper opening. A sequence predictor conditioned on the two endpoint images~\cite{jang2025dreamgen} can infer a plausible intervening motion. Similar endpoints can correspond to different closure times, and the intermediate images provide evidence about when the gripper actually closes. \IDMName{} addresses these two requirements with multiscale features for workspace layout, parallel branches for local geometry, and joint reconstruction of the states aligned with the observed frames throughout a clip. Figure~\ref{fig:idm_architecture} shows how these features reach the pose and gripper heads.

\paragraph{Dual-view spatial representations.}
The input consists of synchronized main and wrist RGB streams sampled at 15 Hz. Each view is resized to $256\times320$ and normalized with ImageNet statistics. A shared ImageNet-pretrained ResNet-18~\cite{He_2016_CVPR,Deng_2009_ImageNet} encodes both views and is fine-tuned with fixed batch-normalization running statistics. Average pooling maps residual stages 1 and 2 to $4\times5$ grids and stage 4 to a single spatial cell. Concatenation yields $4{,}352$ features per view. Separate view projections and a fusion layer produce the 256-dimensional multiscale descriptor $\mathbf v_t$. The grids retain spatial layout while the stage-4 features provide global context.

Two parallel branches provide local evidence that can be weakened by pooling. The dense spatial branch reduces each stage-2 feature map to 16 channels with a $1\times1$ convolution and pools it to an $8\times10$ grid. It concatenates the two grids and projects them to the 256-dimensional feature $\mathbf d_t$. The detail CNN operates on the two RGB views concatenated along channels and resized to $96\times128$. Three stride-2 convolutions with 12, 24, and 32 output channels, followed by pooling and projection, yield the 128-dimensional feature $\mathbf f_t$. Both features refine the shared representation, and $\mathbf f_t$ also enters the gripper head directly.

\paragraph{Auxiliary RGB descriptors.}
Fixed crops from the two views provide grayscale and gradient features, including a crop of the gripper region. A blockwise PCA projection~\cite{Jolliffe_2016_PCA} produces 256 components, and two 48-dimensional PLS projections~\cite{Wold_2001_PLS} use position and the full state as supervision, respectively. We append 96 statistics describing the location, shape, color, and validity of regions selected by fixed HSV thresholds. The resulting 448-dimensional vector $\mathbf a_t$ is concatenated with the centered differences $(\mathbf a_{t+1}-\mathbf a_{t-1})/2$ and $(\mathbf a_{t+2}-2\mathbf a_t+\mathbf a_{t-2})/4$, giving a 1,344-dimensional vector $\mathbf u_t$. These differences are computed within each recording with replicated endpoints. We fit PCA, PLS, and the auxiliary normalization on the training split and keep them fixed for validation and inference. Computing $\mathbf u_t$ at inference requires only RGB. The learned embedding $F_{\mathrm{aux}}$ maps the standardized vector to 256 dimensions.

\paragraph{Residual refinement and state heads.}
The model first combines the multiscale and auxiliary descriptors, then refines the shared representation with local detail and temporal context:
\begin{align}
\mathbf{b}_t &= F_{\mathrm{shared}}([\mathbf{v}_t;F_{\mathrm{aux}}(\mathbf{u}_t)]),\nonumber\\
\mathbf{h}_t &= \mathbf{b}_t+0.1F_{\mathrm{detail}}([\mathbf{f}_t;\mathbf{d}_t]),
\label{eq:ours_spatial_fusion}\\
\mathbf{z}_{1:T} &= \mathbf{h}_{1:T}+0.1\mathcal{T}_{\theta}(\mathbf{h}_{1:T}).
\label{eq:ours_temporal}
\end{align}
Here $[\cdot;\cdot]$ denotes concatenation and $T=17$. The temporal network~\cite{Bai_2018_TCN} uses three length-5 convolutions with dilations $1,2,4$, symmetric zero padding, and channel widths $256\to128\to128\to256$. GELU follows the first two convolutions. The residual connection retains the frame's spatial representation while the temporal convolution adds evidence from preceding and following frames. The auxiliary differences are computed before windowing, so they can also use neighboring frames just outside a window.

The pose heads read $\mathbf z_t$. The position head predicts $\widehat{\mathbf p}_t=\boldsymbol\mu_p+0.1F_p(\mathbf z_t)$ in meters, where $\boldsymbol\mu_p$ is the training-set mean. The orientation head predicts a 6D representation that is converted to a rotation matrix by orthonormalization~\cite{Zhou_2019_CVPR}. The gripper head combines a base logit from the detail CNN with a correction from the shared temporal representation:
\begin{equation}
\widehat g_t=\sigma\!\left(F_g(\mathbf{f}_t)+0.1F_{\mathrm{res}}(\mathbf{z}_t)\right).
\label{eq:ours_heads}
\end{equation}
The direct path lets the gripper head use local evidence such as visible gripper opening, while the residual adds context about how the gripper moves across the clip. Each window yields one state per input position. The 10D output stores Cartesian position, the first two columns of the rotation matrix, and gripper closure in $[0,1]$.

\paragraph{Inference and action labeling.}
With a 17-frame window and stride 8, neighboring full windows share nine frames. Each window can give a shared frame a different estimate because its temporal context differs. We combine these estimates using a normalized center-weighted average. A valid prediction at zero-based position $k$ receives weight $\omega_k=\sin^2(\pi(k+\tfrac12)/T)$. For each frame we divide each contributing weight by the sum of the weights for that frame. We average positions and closure values directly. For orientation we average rotation matrices and project the result onto $\mathrm{SO}(3)$ with determinant-corrected SVD~\cite{Kabsch_1976_Rotation}. This yields one state for each valid frame in the labeling sequence. In the downstream policy dataset, the fused state at frame $t$ serves as the robot-state observation and the fused state at frame $t+3$ serves as the action target.

\subsection{Training}
\label{sec:idm_setup}
\paragraph{Training supervision and window coverage.}
We supervise the absolute state at every valid frame and the motion between adjacent valid frames. The framewise loss is
\begin{align}
\ell_{\mathrm{state}}={}&
\frac{\|\widehat{\mathbf{p}}-\mathbf{p}\|_2}{0.02}
+0.25\frac{d_{\mathrm{ch}}(\widehat R,R)}{5\pi/180}\nonumber\\
&+3\frac{\operatorname{SmoothL1}_{0.05}(\widehat g,g)}{0.05},
\label{eq:ours_state_loss}
\end{align}
where $d_{\mathrm{ch}}(A,B)=\|A-B\|_{\mathrm F}/\sqrt{2}$. For adjacent frames, let $\Delta\mathbf p_t=\mathbf p_t-\mathbf p_{t-1}$ and $\Delta R_t=R_{t-1}^{\mathsf T}R_t$, with the predicted increments defined analogously. We add the translation and relative rotation penalties
\begin{align}
\ell_{\Delta p,t} &= 0.5\frac{\|\Delta\widehat{\mathbf p}_t-\Delta\mathbf p_t\|_2}{0.01},\\
\ell_{\Delta R,t} &= 0.125\frac{d_{\mathrm{ch}}(\Delta\widehat R_t,\Delta R_t)}{2.5\pi/180}.
\end{align}
The framewise loss constrains the robot's configuration, and the increment losses constrain how that configuration changes in the video.

We extract 17-frame windows at stride 8. Short terminal windows repeat the last valid input, with padded targets masked. For a frame $t$ covered by $c_t$ windows, each valid occurrence receives weight $1/c_t$. State and translation terms use this frame weight, with translation omitted at the first position of each window. The relative rotation term uses $1/c^\Delta_t$, where $c^\Delta_t$ counts windows containing both valid consecutive frames of the pair $(t-1,t)$. We sum the weighted terms and divide by the total valid frame weight in the batch.

\paragraph{Data and evaluation.}
\IDMName{} maps main- and wrist-camera RGB to end-effector states without task conditioning, so its training data need not be limited to the policy-evaluation tasks. We train it on eight contact-rich tasks: the five policy-evaluation tasks (\textit{chip pick}, \textit{egg pick}, \textit{charger press}, \textit{pullback car}, and \textit{whiteboard erase}) and three additional tasks (\textit{nut removal}, \textit{nut tightening}, and \textit{charger insertion}), which provide further examples of end-effector motion and gripper closure during contact. Table~\ref{tab:task_coverage} summarizes the tasks used by each component. The 895 training trajectories comprise 355 real-robot and 540 simulated trajectories, and we hold out 100 validation and 150 test trajectories using a fixed trajectory-level split. The offline comparison evaluates the 80
real-world test trajectories: ten per task across the eight tasks, comprising
23,729 frame-aligned labels. Robot replay (Sec.~\ref{sec:idm_realworld}) uses only the five policy-evaluation tasks. All methods use the same camera configuration
and are evaluated against the same ground-truth frames in the same 10D
end-effector state representation. We select checkpoints and inference settings on the
validation split under a common criterion (Sec.~\ref{sec:app_contents_23}) and report final
results on the held-out test set.

\paragraph{Baseline configuration sources.}
We compare with EVA~\cite{Wang_2026_EVA}, DreamGen~\cite{jang2025dreamgen},
and TACO~\cite{Liu_2026_TACO}.
We use the released EVA and DreamGen IDM implementations as starting points
and preserve their core prediction architectures.
We reimplement the action
prediction branch described in the TACO paper using a frozen DINOv2~\cite{oquab2023dinov2} encoder.

\paragraph{Task adaptation and inference.}
We adapt each baseline's data interface to our robot and two camera views.
EVA, DreamGen, TACO, and \IDMName{} use a 150-epoch cap. \IDMName{} uses AdamW with learning rate $3\times10^{-4}$,
weight decay $10^{-4}$, and effective batch size eight.

\subsubsection{Metrics and Checkpoint Selection}
\label{sec:app_contents_23}
Targets are absolute end-effector states: XYZ in meters, the first
two rotation-matrix columns, and gripper closure. For action MAE and
MSE, each component is standardized using statistics from the training split.
We also report Euclidean position error in millimeters, geodesic
rotation error in degrees, and absolute gripper error. Frame errors
are averaged within each trajectory, then across trajectories in
each task and domain, and finally across task and domain groups.
Checkpoints minimize the validation score, macro-averaged over task and domain groups,
\[
\mathcal{J}=\tfrac{1}{3}\left(
\frac{E_{\mathrm{pos}}}{20\,\mathrm{mm}}+
\frac{E_{\mathrm{rot}}}{5^\circ}+
\frac{E_{\mathrm{grip}}}{0.05}\right).
\]
Strict improvement is required; ties retain the earlier checkpoint.
The selected checkpoints come from epoch 132 for \IDMName{}, 17 for EVA,
148 for DreamGen, and 137 for TACO.

\paragraph{Released defaults and adaptations.}
Tables~\ref{tab:idm_training_configuration} and~\ref{tab:idm_optimization_configuration} list our architecture and optimization settings, respectively.
EVA retains its released ResNet-50/SpatialSoftmax predictor,
learning rate $3\times10^{-4}$, and effective batch size 32.
Its released configuration specifies 200,000 updates; our adaptation
instead uses the shared epoch-based selection protocol.
DreamGen retains its SigLIP2 and flow-matching action head,
learning rate $10^{-4}$, weight decay $10^{-5}$, and 5\% warm-up.
Its released defaults specify batch size 16 on one GPU and 10,000 updates;
we use effective batch size 32 and the 150-epoch cap.
Both implementations are adapted to dual-view inputs and 10D targets,
omit the released image augmentations, and use task and trajectory loss weights.
For TACO, the frozen visual encoder and the learning rate of
$3\times10^{-4}$ are our implementation choices.

\paragraph{Scope of the comparison.}
Observation contexts, internal normalization, sample weighting, and
compute budgets differ across models. The baselines use task and trajectory weighting, while \IDMName{}
corrects for overlapping label coverage. 

\begin{table}[!tb]
\centering
\caption{Inverse dynamics reconstruction on 80 real-world test trajectories across eight tasks. Errors use shared 10D targets and are averaged over trajectories and tasks. Action MAE and MSE use standardization fitted on the training split. All models use FP32 inference.}
\label{tab:idm_comparison}
\small
\setlength{\tabcolsep}{2pt}
\begin{tabularx}{\columnwidth}{@{}>{\raggedright\arraybackslash}X*{5}{c}@{}}
\toprule
Method & \shortstack{Action\\MAE $\downarrow$} & \shortstack{Action\\MSE $\downarrow$}
& \shortstack{Pos.\\(mm) $\downarrow$} & \shortstack{Rot.\\($^\circ$) $\downarrow$}
& \shortstack{Gripper\\MAE $\downarrow$} \\
\midrule
TACO\textsuperscript{*}~\cite{Liu_2026_TACO} & 0.08944 & 0.03420 & 13.514 & 2.933 & 0.01506 \\
DreamGen~\cite{jang2025dreamgen} & 0.08271 & 0.02915 & 11.510 & 2.723 & 0.01829 \\
EVA~\cite{Wang_2026_EVA} & 0.06153 & 0.01890 & 9.388 & \textbf{1.972} & 0.00925 \\
\rowcolor{blue!5}
\IDMName{} & \textbf{0.05621} & \textbf{0.01758}
& \textbf{5.515} & 2.189 & \textbf{0.00881} \\
\bottomrule
\end{tabularx}
\par\smallskip
\parbox{\columnwidth}{\footnotesize\textsuperscript{*} We reimplement TACO’s action prediction branch as described in the paper.}
\end{table}

\subsubsection{Benchmark Configurations}
\label{sec:app_idm_configs}
Tables~\ref{tab:idm_training_configuration} and~\ref{tab:idm_optimization_configuration} summarize the architecture, optimization, and inference settings used in the benchmark.

\begin{table*}[!tbp]
\centering
\begin{threeparttable}
\caption{Architecture and input configurations for the inverse dynamics benchmark.}
\label{tab:idm_training_configuration}

\small
\setlength{\tabcolsep}{3pt}
\renewcommand{\arraystretch}{1.08}

\begin{tabularx}{\textwidth}{
@{}>{\raggedright\arraybackslash}p{2.8cm}
*{3}{>{\raggedright\arraybackslash}X}>{\columncolor{blue!5}\raggedright\arraybackslash}X@{}
}
\toprule
Setting
& TACO~\cite{Liu_2026_TACO}\tnote{a}
& DreamGen~\cite{jang2025dreamgen}\tnote{b}
& EVA~\cite{Wang_2026_EVA}
& \textbf{\IDMName{}} \\
\midrule

Visual backbone
& DINOv2-B/14 + registers
& SigLIP2-L/16
& ResNet-50
& ResNet-18 + fine/dense \\

Visual adaptation
& Frozen encoder
& Partial tuning
& Full; SyncBN
& Full; fixed BN statistics \\

RGB input ($H\!\times\!W$)\tnote{c}
& $518\!\times\!518$ tiled
& $2\!\times\!(256\!\times\!256)$
& $256\!\times\!640$ tiled
& $2\!\times\!(256\!\times\!320)$ \\

Observed / predicted frames\tnote{c}
& $1/1$
& $2/16$
& $3/1$
& $17/17$ \\

Window stride (frames)
& 1 & 8 & 1 & 8 \\

Target normalization
& z-score
& Min/max $[-1,1]$
& z-score
& XYZ centering \\

Training loss
& Smooth $L_1$
& Flow velocity MSE
& Smooth $L_1$
& Per-frame pose and gripper loss + inter-frame pose loss \\

\bottomrule
\end{tabularx}
\begin{tablenotes}[flushleft]
\footnotesize

\item[a]
We reimplement TACO's action branch from the details given in its paper,
using a frozen DINOv2 encoder,
trainable directional pooling, F/T, and action modules,
and 12D F/T inputs estimated by inverse FEM.

\item[b]
SigLIP2 encoder layer 11 (zero-based),
the vision pooling head, and logit scale/bias remain frozen.
The fusion transformer and flow-matching DiT
have 4 and 8 layers, respectively.

\item[c]
All methods use two RGB views.
Tiled inputs place the main and wrist views side by side;
TACO additionally resizes the composite to $518\times518$.
At 15\,Hz, observation offsets are
TACO $\{0\}$,
DreamGen $\{0,16\}$,
EVA $\{-2,0,2\}$,
and \IDMName{} $\{0,\ldots,16\}$.

\end{tablenotes}
\end{threeparttable}
\end{table*}

\begin{table*}[!tbp]
\centering
\begin{threeparttable}
\caption{Optimization and inference configurations for the inverse dynamics
benchmark. Effective batch sizes count training
samples/windows. All methods use the same recording-level
data splits and validation-based checkpoint selection.}
\label{tab:idm_optimization_configuration}

\small
\setlength{\tabcolsep}{3pt}
\renewcommand{\arraystretch}{1.08}

\begin{tabularx}{\textwidth}{
@{}>{\raggedright\arraybackslash}p{2.8cm}
*{3}{>{\raggedright\arraybackslash}X}>{\columncolor{blue!5}\raggedright\arraybackslash}X@{}
}
\toprule
Setting
& TACO~\cite{Liu_2026_TACO}
& DreamGen~\cite{jang2025dreamgen}
& EVA~\cite{Wang_2026_EVA}
& \textbf{\IDMName{}} \\
\midrule

Optimizer\tnote{d}
& AdamW
& AdamW
& AdamW
& AdamW \\

Peak learning rate
& $3\!\times\!10^{-4}$
& $1\!\times\!10^{-4}$
& $3\!\times\!10^{-4}$
& $3\!\times\!10^{-4}$ \\

Adam $(\beta_1,\beta_2)$
& $(0.9,0.999)$
& $(0.95,0.999)$
& $(0.9,0.999)$
& $(0.9,0.999)$ \\

Weight decay\tnote{d}
& $10^{-2}$
& $10^{-5}$
& $10^{-2}$
& $10^{-4}$ \\

Gradient norm limit
& 1 & 1 & 1 & 10 \\

Effective batch size
& 32 & 32 & 32 & 8 \\

GPUs $\times$ microbatch $\times$ accumulation
& $2\!\times\!4\!\times\!4$
& $4\!\times\!8\!\times\!1$
& $4\!\times\!8\!\times\!1$
& $4\!\times\!2\!\times\!1$ \\

Warm-up
& 5\% of updates
& 5\% of updates
& 5\% of updates
& None \\

LR schedule
& Cosine to 0
& Cosine to 0
& Cosine to 0
& Cosine: 100 epochs; 5\% floor thereafter \\

\midrule

Maximum budget
& 150 epochs
& 150 epochs
& 150 epochs
& 150 epochs \\

Validation interval
& 1 epoch
& 1 epoch
& 1 epoch
& 1 epoch \\

Early-stop minimum / patience\tnote{e}
& 10 epochs / 10 checks
& 10 epochs / 10 checks
& 10 epochs / 10 checks
& 10 epochs / 10 checks \\

Inference precision
& FP32 & FP32 & FP32 & FP32 \\

\bottomrule
\end{tabularx}
\begin{tablenotes}[flushleft]
\footnotesize
\item[d]
Adam $\epsilon=10^{-8}$ for all methods.
TACO and DreamGen exclude biases and LayerNorm
parameters from weight decay.
\IDMName{} retains trainable BN affine parameters
while fixing running statistics.

\item[e]
Strictly lower validation $\mathcal{J}$ resets patience.
Warm-up percentages and decay horizons refer
to the maximum training budget.

\end{tablenotes}
\end{threeparttable}
\end{table*}

\subsection{Evaluation}
\label{sec:app_idm_evaluation}
\subsubsection{Offline Action Extraction}
\label{sec:idm_offline}
We report action MAE and MSE using a common per-dimension standardization
fitted on the training split, Euclidean position error in millimeters,
geodesic rotation error in degrees, and gripper closure MAE on the
$[0,1]$ scale. Each metric is averaged over frames within a trajectory,
then equally over trajectories within each task and over tasks.
We score raw predictions after each method's fixed decoding and window
aggregation.

Under this shared protocol, \IDMName{} achieves the lowest error on four of the five metrics (Table~\ref{tab:idm_comparison}). The largest advantage over EVA is in end-effector position: error falls from $9.388$\,mm to $5.515$\,mm, a $41.3\%$ reduction. Lower aggregate action and gripper errors further support the accuracy of the extracted trajectories, although EVA retains the best rotation accuracy. Robot replay provides a complementary test of whether these offline improvements translate into successful physical interactions.

\subsubsection{Robot Replay Protocol}
\label{sec:idm_realworld}
For each of the five tasks, every method is evaluated on the same ten held-out real-world test trajectories, with one physical replay per trajectory. To control for differences in the initial scene, we manually align the physical setup with the reference image using an RGB overlay and a pixel-difference heatmap (Appendix Fig.~\ref{fig:idm_pixel_alignment}). We then execute each IDM's extracted action sequence directly on the robot and measure task success.
\par\medskip
\noindent\begin{minipage}{\columnwidth}
    \makeatletter\def\@captype{figure}\makeatother
    \captionsetup{hypcap=false}
    \centering
    \includegraphics[width=0.75\linewidth]{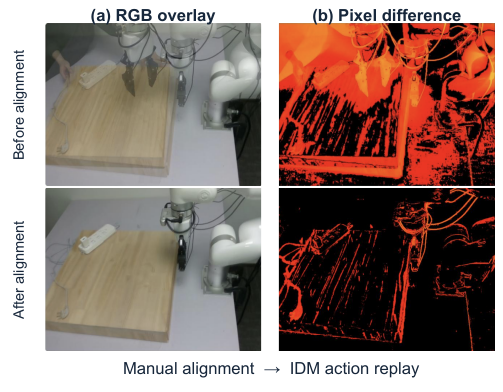}
    \caption{Manual scene alignment before IDM action replay. }
    \label{fig:idm_pixel_alignment}
\end{minipage}\par\medskip


\subsubsection{IDM Component Ablations}
\label{sec:idm_ablation_details}
\paragraph{Action Extraction Components.}
Table~\ref{tab:idm_ablation_replay} evaluates IDM components through robot replay. Removing the spatial or temporal residual reduces success on several tasks; removing gripper correction lowers Car success from 80\% to 40\%. These results highlight the importance of action precision for generating useful trajectories. Offline IDM errors are reported in Appendix Table~\ref{tab:idm_ablation_real80}.

\begin{table}[!htbp]
\centering
\caption{\textbf{Action extraction ablations.} Robot replay SR (\%) for IDM variants trained under the same protocol. Avg. is the unweighted mean across the five tasks.}
\label{tab:idm_ablation_replay}
\small
\setlength{\tabcolsep}{3pt}
\begin{tabularx}{\columnwidth}{@{}Xcccccc@{}}
\toprule
Variant & \textit{Chip} & \textit{Egg} & \textit{Press} & \textit{Car} & \textit{Erase} & Avg. \\
\midrule
\rowcolor{blue!5}
\IDMName{} & \textbf{90} & \textbf{100} & \textbf{100} & \textbf{80} & \textbf{100} & \textbf{94.0} \\
w/o auxiliary RGB & 80 & 90 & 90 & \textbf{80} & \textbf{100} & 88.0 \\
w/o spatial residual & \textbf{90} & \textbf{100} & 80 & 50 & 90 & 82.0 \\
w/o temporal residual & 70 & \textbf{100} & 70 & 60 & \textbf{100} & 80.0 \\
w/o gripper correction & 70 & \textbf{100} & 70 & 40 & 90 & 74.0 \\
\bottomrule
\end{tabularx}
\end{table}

\begin{table*}[!tbp]
\centering
\small
\caption{IDM component ablations on the same 80 real test trajectories.
Parentheses give signed percentage changes relative to the full \IDMName{} model,
computed from unrounded errors as $100(e_{\mathrm{abl}}/e_{\mathrm{full}}-1)$.
Lower is better for every metric.
Weighted Error is a dimensionless average of normalized test errors,
$\bigl(e_{\mathrm{pos}}/(20\,\mathrm{mm})+e_{\mathrm{rot}}/(5^\circ)+3e_{\mathrm{grip}}/0.05\bigr)/5$,
computed from unrounded values. The same scales apply to every variant. The weights of 1, 1, and 3 for normalized position, rotation, and gripper errors place greater emphasis on gripper accuracy.}
\label{tab:idm_ablation_real80}
\setlength{\tabcolsep}{3pt}
\begin{tabular}{@{}lrrrrrr@{}}
\toprule
Variant & Action MAE & Action MSE & Pos. (mm) & Rot. ($^\circ$) & Gripper MAE & Weighted Error $\downarrow$ \\
\midrule
\rowcolor{blue!5}
\IDMName{} & 0.05621 & 0.01758 & 5.515 & 2.189 & 0.00881 & \textbf{0.2485} \\
Without auxiliary RGB & \textbf{0.05024} & \textbf{0.01304} & \textbf{4.800} & \textbf{1.890} & 0.01224 & 0.2705 \\
& {\footnotesize $(-10.61\%)$} & {\footnotesize $(-25.84\%)$}
  & {\footnotesize $(-12.95\%)$} & {\footnotesize $(-13.63\%)$} & {\footnotesize $(+38.86\%)$} & \\
\addlinespace[2pt]
Without spatial residual & 0.06155 & 0.02214 & 6.050 & 2.282 & \textbf{0.00870} & 0.2561 \\
& {\footnotesize $(+9.50\%)$} & {\footnotesize $(+25.93\%)$}
  & {\footnotesize $(+9.71\%)$} & {\footnotesize $(+4.27\%)$} & {\footnotesize $(-1.34\%)$} & \\
\addlinespace[2pt]
Without temporal residual & 0.05991 & 0.01777 & 5.665 & 2.401 & 0.00887 & 0.2592 \\
& {\footnotesize $(+6.58\%)$} & {\footnotesize $(+1.05\%)$}
  & {\footnotesize $(+2.73\%)$} & {\footnotesize $(+9.72\%)$} & {\footnotesize $(+0.66\%)$} & \\
\addlinespace[2pt]
Without gripper correction & 0.05754 & 0.01849 & 5.982 & 1.957 & 0.01314 & 0.2958 \\
& {\footnotesize $(+2.36\%)$} & {\footnotesize $(+5.14\%)$}
  & {\footnotesize $(+8.48\%)$} & {\footnotesize $(-10.57\%)$} & {\footnotesize $(+49.12\%)$} & \\
\bottomrule
\end{tabular}
\end{table*}
Each of the four ablation variants makes a single change:
(i) zero and freeze the auxiliary embedding while preserving fusion dimensions;
(ii) bypass the spatial residual, $h=b$, while retaining the direct detail-CNN
gripper path; (iii) bypass the temporal residual, $z=h$, while retaining auxiliary
temporal features and temporal supervision; and (iv) remove the gripper
correction, $g=\operatorname{sigmoid}(F_g(f))$.
Variants start from the original initialization, not epoch 132.
The full model and all four ablations use an effective batch size of eight.

Table~\ref{tab:idm_ablation_real80} shows a trade-off between pose and
gripper accuracy. Removing auxiliary RGB improves mean position and
rotation errors while increasing gripper MAE from 0.00881 to 0.01224.
Removing the gripper correction raises gripper MAE to 0.01314 and
increases its task- and trajectory-balanced 95th percentile from
0.0353 to 0.0572. Removing either the spatial or temporal residual
increases mean position and rotation errors. 
These comparisons evaluate the same 80 held-out real-world test trajectories.

\section{Policy Learning and Additional Results}
\label{sec:appendix_n0vtla}
\subsection{Experimental Setup}
\label{sec:app_policy_setup}
We use an xArm7 equipped with two Intel RealSense D435i cameras for the main and wrist views and two Xense vision-based tactile sensors mounted on the gripper. We collect 50 demonstrations per task via GELLO teleoperation. The main policy experiments use all 50 demonstrations for \textit{chip pick} and \textit{egg pick}, and 25 for each remaining task.

Each policy is evaluated on 10 out-of-distribution initial layouts per task that do not appear in the training data, with three trials per layout. Appendix~\ref{sec:real_eval_layouts} shows the ten reference photographs used to set up each task. A trial receives 1 for success and 0 for failure. For \textit{whiteboard erase}, complete removal of the target markings receives 1; partial removal that leaves visible target markings receives 0.5; failure to erase the target markings receives 0. We use this partial-credit score when reporting SR for this task, rather than a strictly binary success rate. SR is the mean score across all 30 trials, and Best-of-3 (Bo3) is the average best score over the three trials for each layout, including half credit for partial success on \textit{whiteboard erase}. Both metrics are reported as percentages.

\subsubsection{Stage-wise Scoring Protocol}
\label{sec:app_stage_scoring}

\paragraph{Stage criteria.}
Approach receives 1 when the end effector reaches the vicinity of the task target during the trial, and 0 otherwise. This is a visual assessment without a calibrated distance threshold. In Egg Pick, reaching the vicinity of the egg can receive Approach credit even if the gripper is misaligned for grasping.

Contact receives 1 for the task-specific interaction and 0 otherwise. For Pullback Car, the gripper must secure the car in a controlled grasp. For Egg Pick and Chip Pick, the gripper must grasp and acquire the egg or chip. For Whiteboard Erase, the gripper must acquire the eraser. For Charger Press, the end effector must establish effective contact with the intended press target. Incidental brushes and empty gripper closures receive no Contact credit. A trial can retain Contact credit if it establishes the required interaction and subsequently fails to complete the task.

\paragraph{Aggregation.}
Each condition contains five tasks, with ten unseen layouts and three trials per layout for each task. For each metric, we sum the trial scores and divide by 30 within each task, then average the five task means and multiply by 100. Equivalently, the plotted percentage is the sum of the 150 trial scores divided by 150, times 100. All trials enter each denominator, including trials that did not reach an earlier stage. A single trial can receive credit for multiple stages. Success uses the recorded outcome score.

\subsection{Training Configurations}
\label{sec:app_policy_training}
For a given policy and task, the data settings within each comparison
use the same optimization budget and learning-rate schedule. Budgets
and schedules can vary across tasks, policies, and experiments.
We evaluate the final checkpoint for each policy and task.

\paragraph{Policy hyperparameters.}
\label{sec:app_policy_hyperparameters}
Tables~\ref{tab:n0vtla_settings}--\ref{tab:ftp1_settings} list the task-specific training schedules. All four policies use AdamW with a base learning rate of $5\times10^{-5}$, and $\epsilon=10^{-8}$. The action prediction horizon is 25 frames for chip pick, 15 for egg pick and pullback car, 10 for charger press, and 30 for whiteboard erase.

For \Nzero{}-VTLA, $\pi_{0.5}$, and FTP-1, we use AdamW coefficients $(0.9,0.95)$, weight decay $10^{-10}$, 1,000 warmup steps, a gradient clipping norm of 1, and 10 inference steps. In the 1:1 training configurations, the per-device batch size is 12, except for \Nzero{}-VTLA on egg pick, which uses 24. ACT uses a per-device batch size of 64, AdamW coefficients $(0.9,0.999)$, weight decay $10^{-4}$, a gradient clipping norm of 10, and a visual backbone learning rate of $10^{-5}$ with no warmup. In the schedule tables, \emph{No decay} indicates a constant learning rate after any warmup. Numeric endpoints indicate when cosine decay reaches $10\%$ of the peak rate.

\begin{table}[!htbp]
\centering
    \caption{\Nzero{}-VTLA training schedule for real-world policy evaluation.}
    \label{tab:n0vtla_settings}
    \small
    \begin{tabularx}{\columnwidth}{@{}Xcc@{}}
        \toprule
        Task & Updates & Decay endpoint \\
        \midrule
        \textit{chip pick} & $20{,}000$ & $20{,}000$ \\
        \textit{egg pick}, \textit{charger press}, \textit{pullback car}, \textit{whiteboard erase} & $20{,}000$ & No decay \\
        \bottomrule
    \end{tabularx}
    \par\smallskip
    \begin{minipage}{\columnwidth}\footnotesize
        We use a peak learning rate of $5\times10^{-6}$ for the tactile predictor, tactile encoder, and latent projection on all tasks except \textit{egg pick}. Task schedules also apply to the data-source, synthetic tactile, and verifier ablations.
    \end{minipage}
\end{table}

\begin{table}[!htbp]
\centering
    \caption{$\pi_{0.5}$ training schedule for real-world policy evaluation.}
    \label{tab:pi05_settings}
    \small
    \begin{tabularx}{\columnwidth}{@{}Xcc@{}}
        \toprule
        Task & Updates & Decay endpoint \\
        \midrule
        \textit{egg pick} & $20{,}000$ & $20{,}000$ \\
        \textit{chip pick}, \textit{charger press}, \textit{pullback car}, \textit{whiteboard erase} & $20{,}000$ & No decay \\
        \bottomrule
    \end{tabularx}
\end{table}

\begin{table}[!htbp]
\centering
    \caption{ACT training schedule for real-world policy evaluation.}
    \label{tab:act_settings}
    \small
    \begin{tabularx}{\columnwidth}{@{}Xcc@{}}
        \toprule
        Task & Updates & Decay endpoint \\
        \midrule
        \textit{chip pick}, \textit{egg pick}, \textit{pullback car} & $20{,}000$ & No decay \\
        \textit{charger press}, \textit{whiteboard erase} & $40{,}000$ & No decay \\
        \bottomrule
    \end{tabularx}
\end{table}

\begin{table}[!htbp]
\centering
    \caption{FTP-1 training schedule for real-world policy evaluation.}
    \label{tab:ftp1_settings}
    \small
    \begin{tabularx}{\columnwidth}{@{}Xcc@{}}
        \toprule
        Task & Updates & Decay endpoint \\
        \midrule
        \textit{chip pick}, \textit{egg pick}, \textit{charger press}, \textit{pullback car} & $20{,}000$ & $20{,}000$ \\
        \textit{whiteboard erase} & $30{,}000$ & $20{,}000$ \\
        \bottomrule
    \end{tabularx}
\end{table}

\subsection{Data Scaling}
\label{sec:app_scaling_configuration}
Table~\ref{tab:real_policy_schedules} reports the task-specific schedules for data scaling.
\par\smallskip\noindent
\begin{minipage}{\columnwidth}
\captionsetup{hypcap=false}
    \centering
    \captionof{table}{\Nzero{}-VTLA training schedules for the data scaling experiment.
    Within each task, the schedule is shared across the four
    real-to-synthetic data ratios.}
    \label{tab:real_policy_schedules}
    \small
    \begin{tabularx}{\columnwidth}{@{}Xcc@{}}
        \toprule
        Task & Updates & Decay endpoint \\
        \midrule
        \textit{chip pick}, \textit{egg pick} & $40{,}000$ & $20{,}000$ \\
        \textit{charger press}, \textit{pullback car} & $20{,}000$ & No decay \\
        \textit{whiteboard erase} & $40{,}000$ & $40{,}000$ \\
        \bottomrule
    \end{tabularx}
    \par\smallskip
    \begin{minipage}{\columnwidth}\footnotesize
        For \textit{egg pick}, the mixed settings vary the size of the synthetic
        pool and sample equally from the real and synthetic sources.
    \end{minipage}
\end{minipage}\par\smallskip

\section{Additional Related Work}
\label{sec:additional_related_work}
\subsection{Visuo-Tactile-Language-Action Models}
\label{sec:rw_vtla}

Vision-language-action (VLA) models have made rapid progress in recent years~\cite{zhao2023learning,kim2024openvla,nvidia2025gr00t,physicalintelligence2025pi05}. By taking a visual observation of the scene alongside a natural language instruction, a VLA policy directly predicts robot actions, enabling a single model to handle diverse manipulation tasks simply by changing the prompt.

However, during physical interaction, the gripper often occludes the very contact regions a robot needs to perceive. Visuo-tactile-language-action (VTLA) models incorporate touch to recover this missing feedback. For instance, VLA-Touch~\cite{bi2026vla} leverages tactile signals for both high-level planning and low-level action refinement, while \Nzero{}-VTLA~\cite{neoteai2026n0vtla} uses large-scale pretraining to anticipate contact dynamics over an upcoming action chunk and guide execution accordingly. FTP-1~\cite{yuan2026ftp} bridges hardware differences across tactile sensors by mapping their readings into a shared, morphology-aware representation. 
Building on these efforts, \methodName{} generates synthetic videos with synchronized tactile observations and extracts action labels from the videos retained after multimodal verification. The resulting synthetic demonstrations are combined with real-world demonstrations to train VLA and VTLA policies.

\subsection{Inverse Dynamics Models for Action Extraction}
\label{sec:rw_idm}
Inverse dynamics models recover action labels from visual observations, allowing generated robot videos to serve as policy-training data. DreamGen~\cite{jang2025dreamgen} adapts a video world model to the target robot embodiment, generates manipulation videos from initial images and task instructions, and recovers pseudo-actions using an inverse dynamics model or a latent action model. The resulting synthetic trajectories augment demonstrations for policy learning.

TACO~\cite{Liu_2026_TACO} uses an inverse dynamics and value model (IDVM) within a Recognize--Imagine--Label loop for policy post-training. Progress estimates identify failure-adjacent states in real rollouts, from which a visuo-tactile world model generates local corrective clips. The IDVM assigns actions and progress scores to these clips. Candidates are checked for kinematic feasibility and tactile plausibility, selected by predicted progress gain, and incorporated into iterative policy post-training.

EVA~\cite{Wang_2026_EVA} uses an IDM to decode action sequences from generated videos. It constructs rewards from action smoothness and compliance with velocity and acceleration limits, and uses these rewards for reinforcement-learning post-training of the video generator to improve action executability. In \methodName{}, \VerifierName{} screens generated visual and tactile observations before \IDMName{} extracts action labels. \IDMName{} combines dual-view spatial detail with neighboring-frame context to support accurate pose and gripper prediction.

\section{Simulation Environment and Data Collection}
\label{sec:sim_env}
We use our simulation environment to collect training trajectories for the world model and the inverse dynamics model.

\subsection{Real-to-Sim Scene Reconstruction}
\label{sec:sim_env_1}
\begin{figure*}[!tbp]
    \centering
    \begin{tabular}{@{}ccc@{}}
    \includegraphics[width=.30\textwidth]{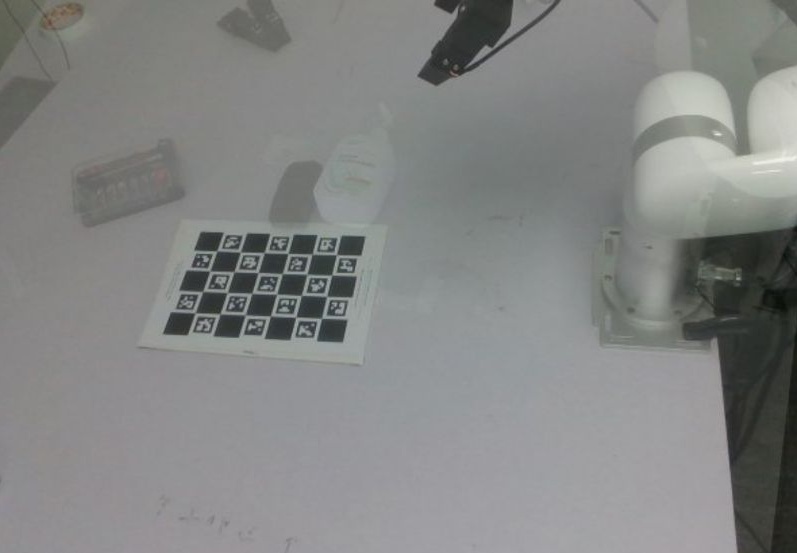} &
    \includegraphics[width=.30\textwidth]{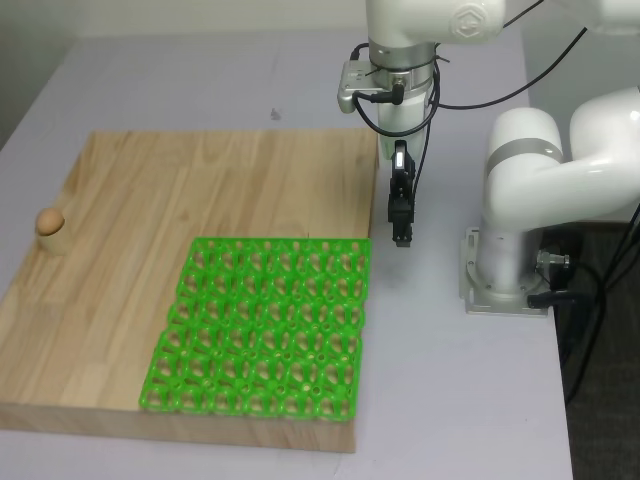} &
    \includegraphics[width=.30\textwidth]{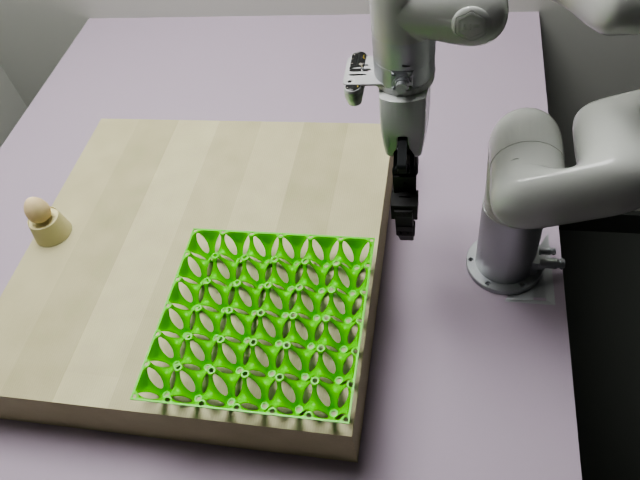} \\
    (a) Calibration target & (b) Real Egg Pick layout & (c) Simulated Egg Pick layout
    \end{tabular}
    \caption{Materials used to reconstruct the workspace and the Egg Pick task. The calibration target is shown in the physical workspace. The real and simulated initial frames show the egg on its support and the placement tray relative to the gripper.}
    \label{fig:r2s_egg_materials}
\end{figure*}

A calibration board on the tabletop defines a common frame for the xArm7, the workspace geometry, and the cameras. We reproduce the main camera pose and attach the wrist camera to the simulated flange so that its view follows the arm. Newton~\cite{newton2025physics} simulates contact and object motion in this frame, which the motion planner also uses, and Blender Cycles renders the two camera views. Figure~\ref{fig:r2s_egg_materials} shows the calibration board and compares the physical Egg Pick setup with its reconstruction.

\paragraph{Simulating fingertip touch.}
We model a Xense Photon optical tactile sensor on each gripper finger. When an object presses against a fingertip, the simulated gel pad deforms at the contact surface. We use the sensor reference mesh from the Xense SDK~\cite{xense_sdk} and render each deformed pad with the Photon renderer to produce one tactile RGB image per fingertip.

\subsection{Task Definition}
\label{sec:sim_env_2}
\begin{figure*}[!tbp]
    \centering
    \includegraphics[width=\textwidth]{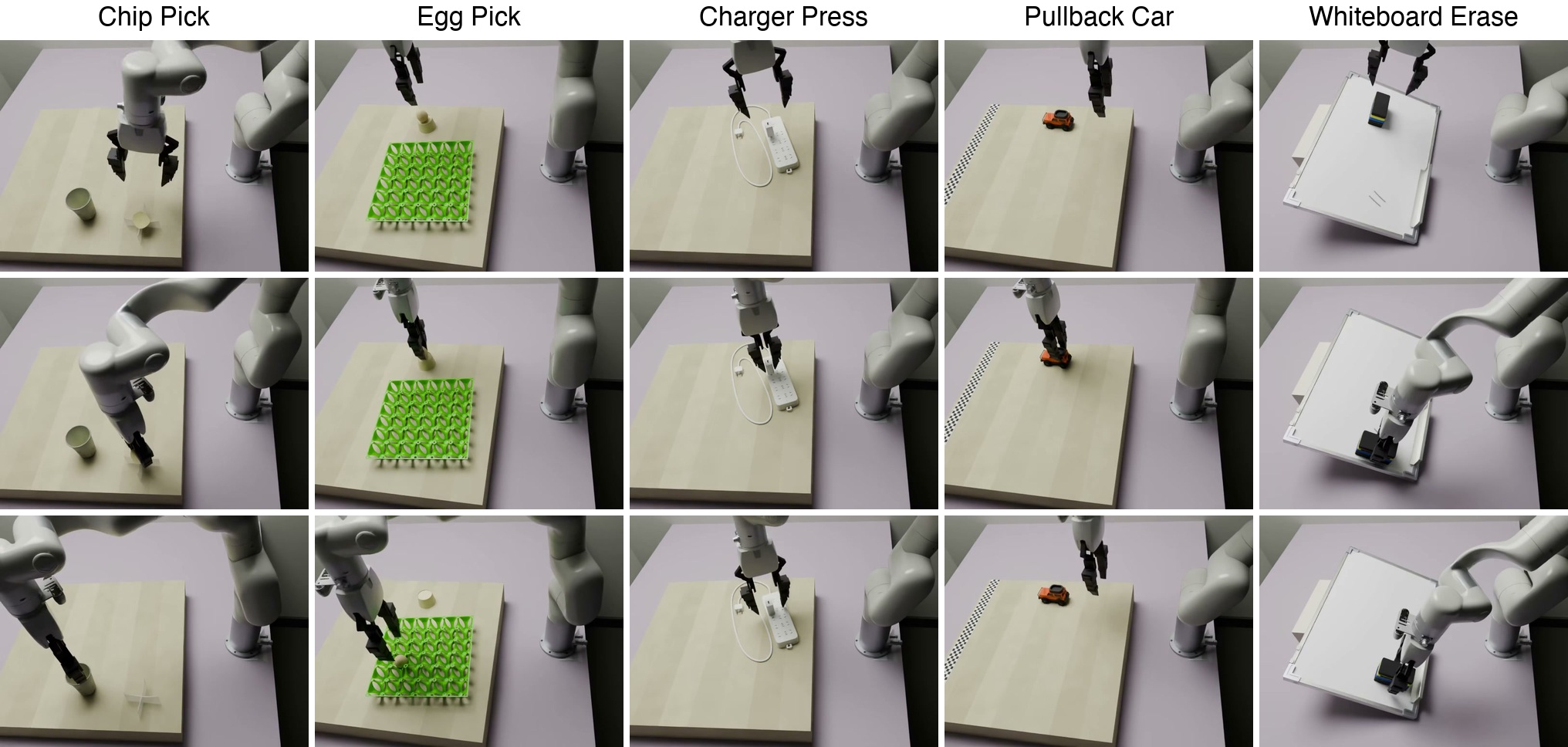}
    \caption{Main-camera views of the five simulated tasks. Each column follows a single demonstration from top to bottom, with frames sampled at 10\%, 50\%, and 90\% of its duration.}
    \label{fig:sim_task_sequences}
\end{figure*}

Figure~\ref{fig:sim_task_sequences} shows the robot performing the following five tasks in the reconstructed workspace.
\begin{itemize}
    \item \textit{Chip Pick.} Lift a chip from its holder and place it in the cup without breaking or dropping it.
    \item \textit{Egg Pick.} Lift the egg from its support and gently place it in the egg tray.
    \item \textit{Charger Press.} Press down on the partially inserted USB power adapter until it is fully seated in the power strip.
    \item \textit{Pullback Car.} Press the toy car down and pull it backward to wind its spring, then release it so it drives across the finish line.
    \item \textit{Whiteboard Erase.} Pick up the eraser and keep it in contact with the board while wiping all marks from the target area, then lift it away.
\end{itemize}

\subsection{Visuo-Tactile Demonstration Collection}
\label{sec:sim_env_3}
In the five-task policy dataset, we record the main and wrist camera views, the robot state, and the action at 15 Hz, and the left and right tactile images at 60 Hz. Timestamps and frame indices align these streams within each episode.

We collected simulated demonstrations using 10 NVIDIA GeForce RTX 4090 GPUs. The simulation dataset used for world model adaptation contains 3,639 entries. The five target tasks account for 3,006 entries (734 for Chip Pick, 673 for Egg Pick, 557 for Charger Press, 653 for Pullback Car, and 389 for Whiteboard Erase). The remaining 633 come from an additional screw-thread task. The training split contains 3,275 entries and the validation split contains 364. Appendices~\ref{sec:app_wm_training} and~\ref{sec:idm_setup} describe the datasets used to train \WMName{} and \IDMName{}, respectively.

\subsection{Task Coverage Across Components}
\label{sec:task_coverage}
Table~\ref{tab:task_coverage} lists the data and tasks used by each component. Policy learning and \WMName{} fine-tuning use the five target tasks. \IDMName{} is task-agnostic and is trained on three additional contact-rich tasks (Appendix~\ref{sec:idm_setup}).

\begin{table}[!tb]
\centering
\caption{Data and task coverage of each component. \emph{Five target} denotes \textit{chip pick}, \textit{egg pick}, \textit{charger press}, \textit{pullback car}, and \textit{whiteboard erase}. Trajectory counts are training trajectories unless noted.}
\label{tab:task_coverage}
\footnotesize
\setlength{\tabcolsep}{3pt}
\renewcommand{\arraystretch}{1.12}
\begin{tabularx}{\columnwidth}{@{}>{\raggedright\arraybackslash}p{0.27\columnwidth}>{\raggedright\arraybackslash}p{0.30\columnwidth}>{\raggedright\arraybackslash}X@{}}
\toprule
Component & Data & Tasks \\
\midrule
TactileVAE & 12,512 OmniViTac & OmniViTac \\
\WMName{} pretraining & 16,593 OmniViTac; 3,056 Gello + 3,275 sim. & OmniViTac; five target + screw thread (sim.) \\
\WMName{} fine-tuning & 200 real & Five target \\
\VerifierName{} (visual) & ${\sim}900$ generated videos & Online robot videos \\
\VerifierName{} (tactile) & 460 trajectories (eval.) & Eight tasks \\
\IDMName{} & 355 real + 540 sim. & Five target + nut removal, nut tightening, charger insertion \\
Policies & 25--50 real demos per task + synthetic & Five target \\
OpenViTac pipeline & 30--60 demos per task & Four OpenViTac tasks \\
\bottomrule
\end{tabularx}
\end{table}

\section{End-to-End Pipeline Evaluation on OpenViTac}
\label{sec:openvitac_eval}
We run the full VeriTouch pipeline on OpenViTac and evaluate the resulting policies on simulated manipulation tasks.

\subsection{Experimental Setup}
\label{sec:openvitac_eval_1}
We use four simulated tasks from OpenViTac~\cite{wu2026openvitac}: Pull Drawer, Insert USB, Insert Block v1, and Insert Block v2. Pull Drawer tests sustained contact during pulling, while the three insertion tasks require alignment with a receptacle. We use OpenViTac's Xense sensor configuration, in which each trajectory contains main and wrist camera images, left and right tactile images, robot states, and actions at 15 Hz. We use 30 original simulator demonstrations per task for training, except for Insert Block v2, which uses 60. Each task has five additional original demonstrations for validation.

\subsection{Pipeline Adaptation and Policy Training}
\label{sec:openvitac_eval_2}
Starting from task instructions and initial observations, VeriGen generates candidate sequences, VeriEval screens the observations, and VeriIDM assigns action targets to retained sequences. We combine these labeled sequences with the original simulator demonstrations for policy training. The training configurations cover $\pi_{0.5}$, $\mathcal{N}_0$-VTLA, and FTP-1. For $\pi_{0.5}$, we follow the OpenViTac visuo-tactile LoRA recipe~\cite{wu2026openvitac}, using four RGB inputs from the main camera, wrist camera, and two fingertip sensors.

\paragraph{VeriGen Adaptation.}
We adapt VeriGen to the OpenViTac simulation tasks using the same training settings as for the real-robot tasks. We initialize the model from the weights pretrained on the Gello--simulation mixture and fine-tune it for 1,000 steps on the 352 OpenViTac trajectories used for world-model training. The training objective and trainable modules follow Sec.~\ref{sec:appendix_optimization_inference}. As in real-robot task fine-tuning, the peak learning rates are $10^{-5}$ for the video expert and $2\times10^{-5}$ for the tactile expert and joint attention layers. We train on 12 NVIDIA A800 GPUs with two gradient accumulation steps to keep the same global batch size of 24 as in the real-robot experiments. The dataset for VeriGen training and evaluation contains 440 trajectories, with 110 per task. Each task is split into 88 training, 11 validation, and 11 test trajectories, yielding 352 training, 44 validation, and 44 test trajectories across the four tasks.

\paragraph{Tactile RGB Reconstruction.}
OpenViTac renders tactile RGB images from simulated depth maps. Their appearance differs from that of images produced by the physical sensors, so the hardware SDK renderer does not directly reproduce the simulator's tactile observations. We use a learned deformation-to-RGB model that takes the predicted 3D displacement field and a fixed reference tactile image for each fingertip. It warps the reference image and applies a learned correction for changes in appearance due to contact. The same reconstruction approach can also be used with reference images and displacement fields from physical sensors.

\paragraph{VeriIDM Adaptation.}
We train the VeriIDM architecture on 1,400 OpenViTac trajectories (350 per task) and use 60 additional trajectories (15 per task) for validation. We initialize the ResNet-18 backbone with ImageNet weights and initialize the prediction heads from scratch. The model takes synchronized main and wrist RGB clips in 17-frame windows with stride 8. At each timestep, the training target is the 10-dimensional end-effector command issued by the simulator, expressed as 3D position, two columns of the rotation matrix, and gripper closure.

\subsection{Visuo-Tactile Generation Quality}
\label{sec:openvitac_wm_qual}
\paragraph{Evaluation Protocol.}
To test how well the shared pretrained model adapts to new task settings, we fine-tune VeriGen on the OpenViTac simulation data using the same settings as for the real-robot tasks and evaluate generation quality with the same protocol. The evaluation set contains 44 held-out test trajectories across four tasks, with 11 trajectories per task. For each trajectory, we use the same task caption and initial Main/Wrist images to generate videos and bilateral tactile deformation sequences with three different random seeds, yielding 132 generated trajectories in total. The metrics and their computation follow Sec.~\ref{sec:app_contents_12}, covering visual quality, instruction following, physics alignment, and tactile deformation quality. We first average each metric over the three rollouts of the same test trajectory and then equally across test trajectories, reporting overall results across the four tasks.

\paragraph{Quantitative Results.}
Table~\ref{tab:openvitac_generation_quality} summarizes the overall results on OpenViTac. \WMName{} achieves 23.55\,dB PSNR, 0.70 SSIM, and 0.11 LPIPS, with instruction-following scores of 99.24\% and 100.00\%. For tactile generation, MSE and 3DEPE are 0.147\,mm$^2$ and 0.41\,mm, respectively, while a Shape Cosine of 0.92 and a Temporal IoU of 0.99 indicate agreement with ground truth in contact shape and timing.

\begin{table}[!htbp]
\centering
\footnotesize
\setlength{\tabcolsep}{1.8pt}
\renewcommand{\arraystretch}{1.1}
\captionsetup{justification=raggedright,singlelinecheck=false,skip=4pt}
\caption{\WMName{} visual and tactile generation quality on OpenViTac.}
\label{tab:openvitac_generation_quality}
\begin{tabular}{@{}c@{\hspace{3pt}}c@{}}
\toprule
\makebox[.16\columnwidth][c]{\textbf{\WMName{}}} &
\begin{minipage}[c]{.82\columnwidth}
\centering
\begin{tabular}{*{6}{>{\centering\arraybackslash}p{\dimexpr\linewidth/6-2\tabcolsep\relax}}}
\multicolumn{3}{c}{\textbf{Visual Quality}} & \multicolumn{3}{c}{\textbf{DreamGen Bench}} \\
\cmidrule(lr){1-3}\cmidrule(lr){4-6}
\mbox{PSNR\,$\scriptstyle\uparrow$} & \mbox{SSIM\,$\scriptstyle\uparrow$} & \mbox{LPIPS\,$\scriptstyle\downarrow$}
& \mbox{IF-Q\,$\scriptstyle\uparrow$} & \mbox{IF-G\,$\scriptstyle\uparrow$} & \mbox{PA\,$\scriptstyle\uparrow$} \\
23.55 & 0.70 & 0.11 & 99.24 & 100.00 & 54.69 \\
\end{tabular}
\par\vspace{5pt}
\setlength{\tabcolsep}{1pt}
\begin{tabular*}{\linewidth}{@{\extracolsep{\fill}}*{5}{c}@{}}
\multicolumn{5}{c}{\textbf{Tactile Quality}} \\
\cmidrule(lr){1-5}
\mbox{MSE\,$\scriptstyle\downarrow$} & \mbox{3DEPE\,$\scriptstyle\downarrow$}
& \mbox{Shape Cosine\,$\scriptstyle\uparrow$} & \mbox{Temp. IoU\,$\scriptstyle\uparrow$}
& \mbox{Temp. Corr.\,$\scriptstyle\uparrow$} \\
0.147 & 0.41 & 0.92 & 0.99 & 0.87 \\
\end{tabular*}
\end{minipage} \\
\bottomrule
\end{tabular}
\end{table}

\paragraph{Qualitative Examples.}
Selected paired GT and \WMName{} comparisons for the four OpenViTac tasks are provided in Appendix~\ref{sec:openvitac_qualitative_comparisons}.

\subsection{Task Scoring and Evaluation Protocol}
\label{sec:openvitac_eval_3}
OpenViTac evaluates task completion using simulator state~\cite{wu2026openvitac}. In Pull Drawer, the drawer must travel far enough along the pull direction while remaining within the allowed vertical displacement. In Insert USB, the evaluator checks lateral and depth alignment in the socket frame as well as end-effector height and angular alignment. For block insertion, the evaluator checks whether the target object's origin lies inside the designated success region in the receptacle frame after release and settling. Each evaluated rollout contributes one binary outcome. We report task success as the number of successful rollouts divided by the number evaluated, expressed as a percentage.

\subsection{Downstream Policy Results}
\label{sec:openvitac_eval_4}
For each policy and task, we compare closed-loop success after training on the original simulator demonstrations alone with success after adding VeriTouch trajectories. The original demonstration budget is fixed within each comparison. We report results separately for each task so that changes in drawer pulling and insertion performance can be assessed independently.

Table~\ref{tab:openvitac_downstream_results} reports the success rates.

\begin{table}[!htbp]
\centering
\caption{\textbf{Downstream policy performance on OpenViTac.}
Baseline uses only the original simulator demonstrations. \methodName{} adds an equal number of generated demonstrations.
Success rates (\%) are computed over 30 rollouts per condition. Bold indicates a gain over the corresponding baseline.}
\label{tab:openvitac_downstream_results}
\footnotesize
\setlength{\tabcolsep}{2pt}
\begin{tabularx}{\columnwidth}{@{}ll*{4}{>{\centering\arraybackslash}X}@{}}
\toprule
Policy & Method & \shortstack{Pull\\Drawer} & \shortstack{Insert\\USB} & \shortstack{Insert\\Block v1} & \shortstack{Insert\\Block v2} \\
\midrule
& Baseline & 23.3 & 0.0 & 0.0 & 6.7 \\
\rowcolor{blue!5}
\cellcolor{white}\multirow{-2}{*}{$\pi_{0.5}$ (w/ tactile)} & \methodName{} & \textbf{63.3} & \textbf{3.3} & \textbf{13.3} & \textbf{13.3} \\
\addlinespace[4pt]
& Baseline & 100.0 & 0.0 & 0.0 & 3.3 \\
\rowcolor{blue!5}
\cellcolor{white}\multirow{-2}{*}{\Nzero{}-VTLA} & \methodName{} & 100.0 & \textbf{6.7} & \textbf{10.0} & \textbf{10.0} \\
\addlinespace[4pt]
& Baseline & 86.7 & 0.0 & 10.0 & 3.3 \\
\rowcolor{blue!5}
\cellcolor{white}\multirow{-2}{*}{FTP-1} & \methodName{} & \textbf{100.0} & \textbf{10.0} & \textbf{13.3} & \textbf{6.7} \\
\bottomrule
\end{tabularx}
\end{table}

\subsection{Training Configurations}
\label{sec:openvitac_configurations}
Table~\ref{tab:openvitac_training_config} summarizes the policy training settings for the results in Table~\ref{tab:openvitac_downstream_results}. Baseline uses 30 original demonstrations for Pull Drawer, Insert USB, and Insert Block v1, and 60 for Insert Block v2. \methodName{} adds the same number of generated demonstrations in each case. The two conditions share the training hyperparameters for each policy and task.

\begin{table}[!htbp]
\centering
\caption{OpenViTac policy training configurations. The learning-rate columns give the peak and terminal rates in units of $10^{-5}$. Batch sizes are global.}
\label{tab:openvitac_training_config}
\footnotesize
\setlength{\tabcolsep}{2pt}
\begin{tabularx}{\columnwidth}{@{}Xlccccc@{}}
\toprule
Tasks & Policy & Batch & Updates & Warmup & Peak & End \\
\midrule
\multirow{3}{=}{Pull Drawer, Insert USB, Insert Block v1}
 & $\pi_{0.5}$ & 16 & 30,000 & 10,000 & 5 & 5 \\
 & \Nzero{}-VTLA & 16 & 20,000 & 1,000 & 5 & 5 \\
 & FTP-1 & 16 & 20,000 & 500 & 5 & 0.5 \\
\midrule
\multirow{3}{=}{Insert Block v2}
 & $\pi_{0.5}$ & 24 & 30,000 & 10,000 & 5 & 0.5 \\
 & \Nzero{}-VTLA & 24 & 20,000 & 500 & 2 & 0.2 \\
 & FTP-1 & 24 & 20,000 & 500 & 5 & 0.5 \\
\bottomrule
\end{tabularx}
\end{table}

We adapt $\pi_{0.5}$ using tactile inputs and LoRA. As in the real-world schedule tables, equal peak and terminal rates indicate no decay after warmup. Otherwise, the schedule uses cosine decay to $10\%$ of the peak rate.

\section{Visualization Comparison}
\label{sec:visualization_comparison}

\subsection{Real-World Evaluation Layouts}
\label{sec:real_eval_layouts}
Each photograph shows one of the ten initial layouts for the corresponding task in Appendix~\ref{sec:app_policy_setup}. Within each task, layouts 1--5 are shown in the top row and layouts 6--10 in the bottom row, from left to right.
\par\smallskip\noindent\begin{minipage}{\columnwidth}\textbf{Chip Pick}\par
\noindent \includegraphics[width=.192\linewidth]{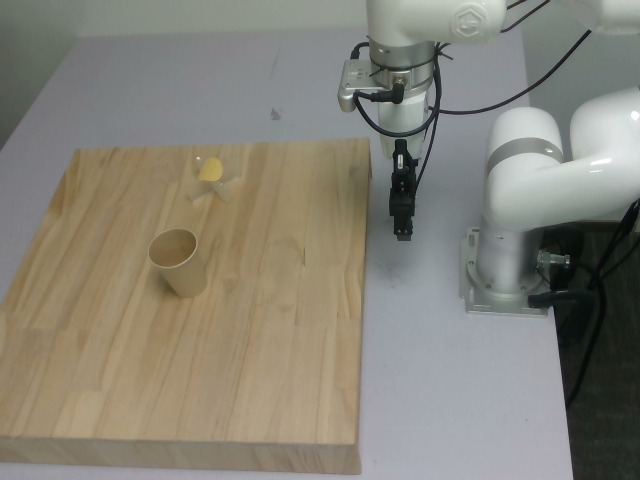}\hfill\includegraphics[width=.192\linewidth]{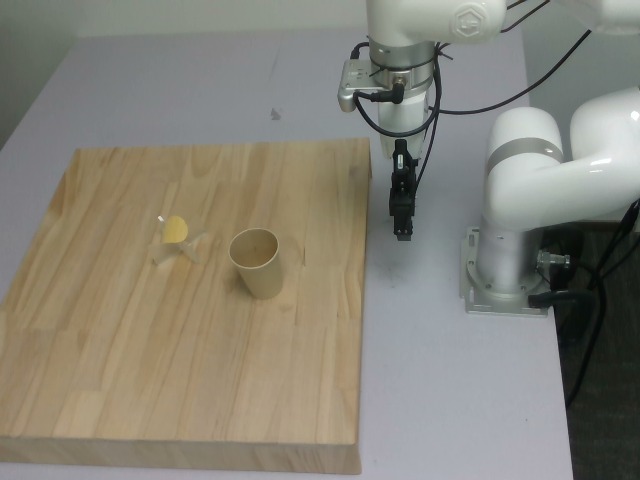}\hfill\includegraphics[width=.192\linewidth]{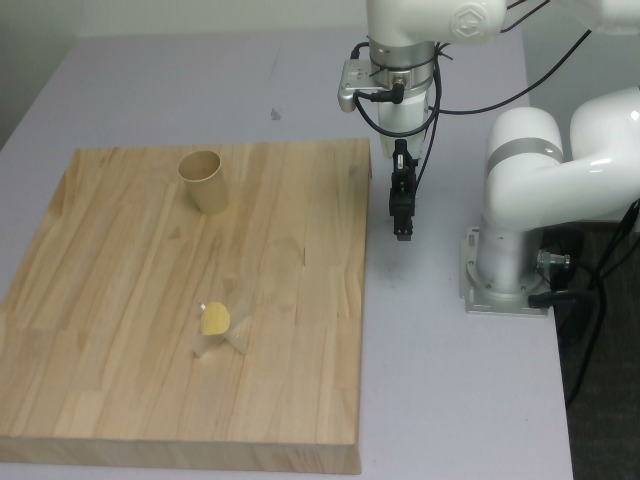}\hfill\includegraphics[width=.192\linewidth]{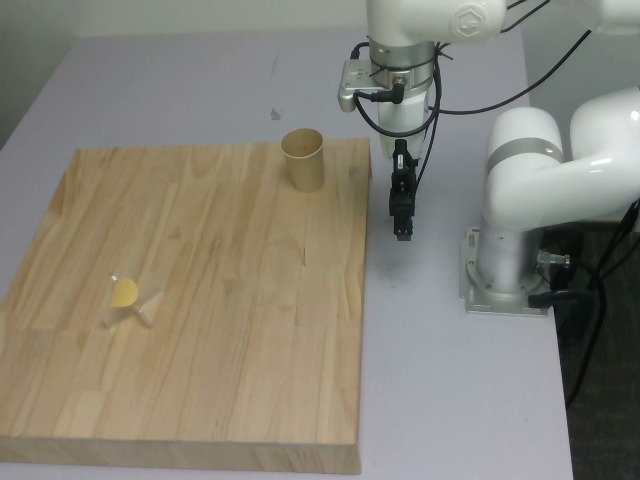}\hfill\includegraphics[width=.192\linewidth]{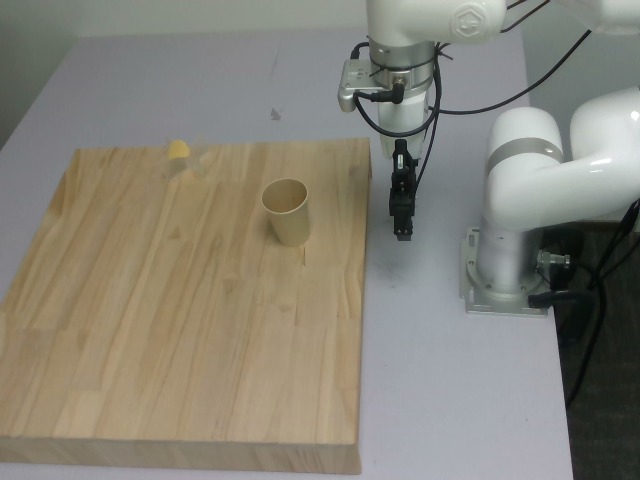}\par
\noindent \includegraphics[width=.192\linewidth]{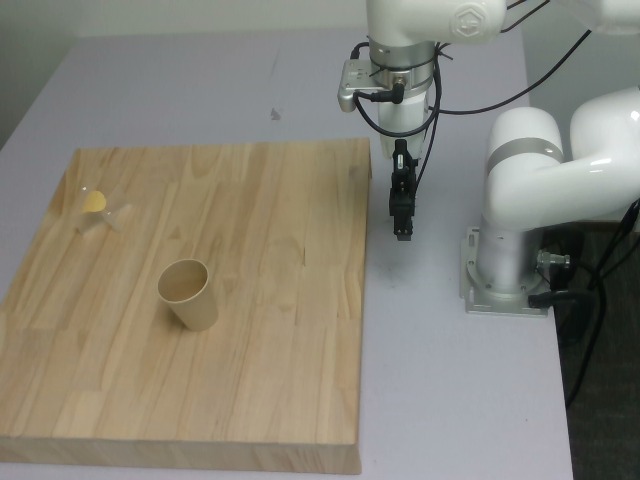}\hfill\includegraphics[width=.192\linewidth]{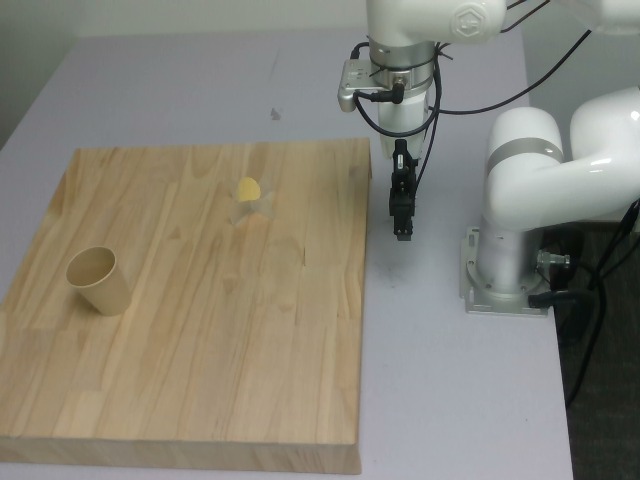}\hfill\includegraphics[width=.192\linewidth]{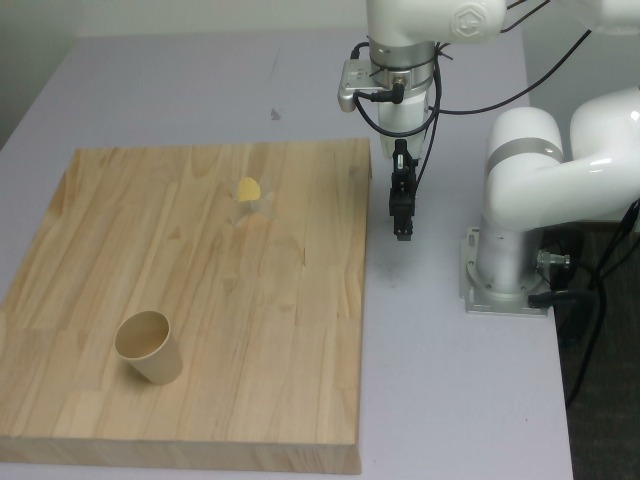}\hfill\includegraphics[width=.192\linewidth]{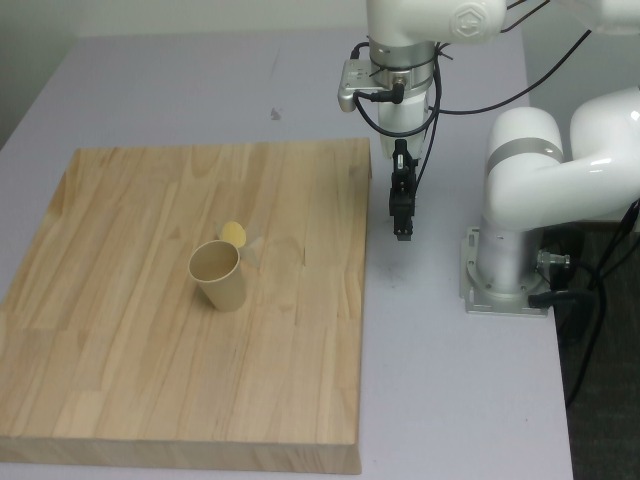}\hfill\includegraphics[width=.192\linewidth]{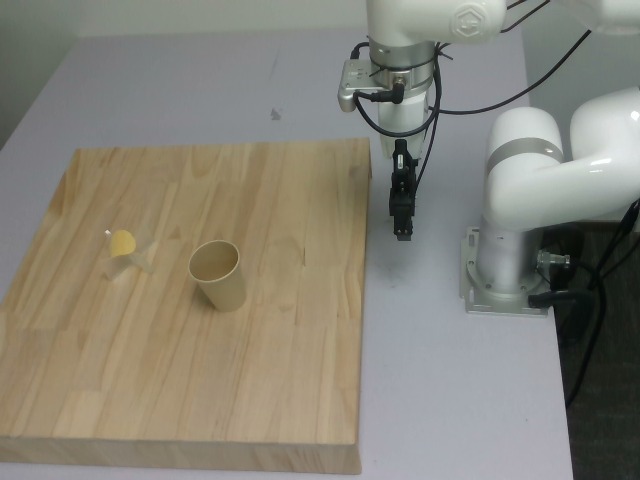}\par
\end{minipage}\par
\par\smallskip\noindent\begin{minipage}{\columnwidth}\textbf{Egg Pick}\par
\noindent \includegraphics[width=.192\linewidth]{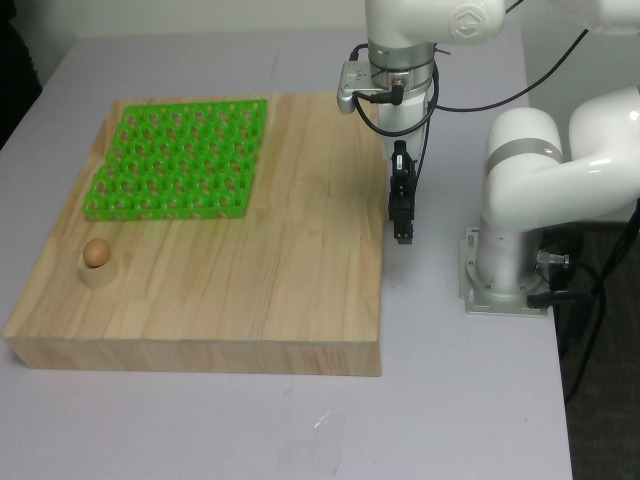}\hfill\includegraphics[width=.192\linewidth]{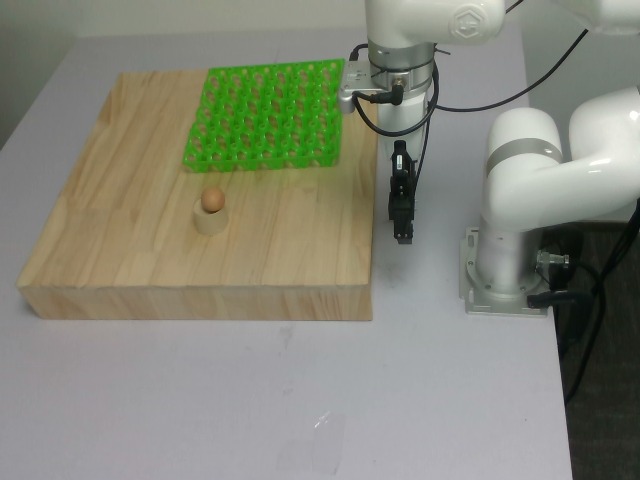}\hfill\includegraphics[width=.192\linewidth]{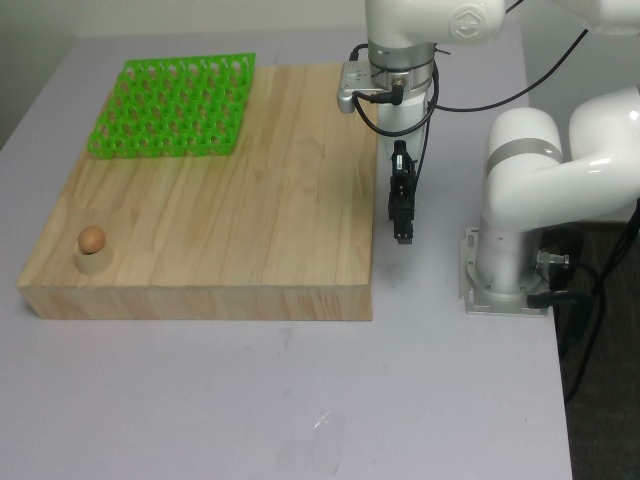}\hfill\includegraphics[width=.192\linewidth]{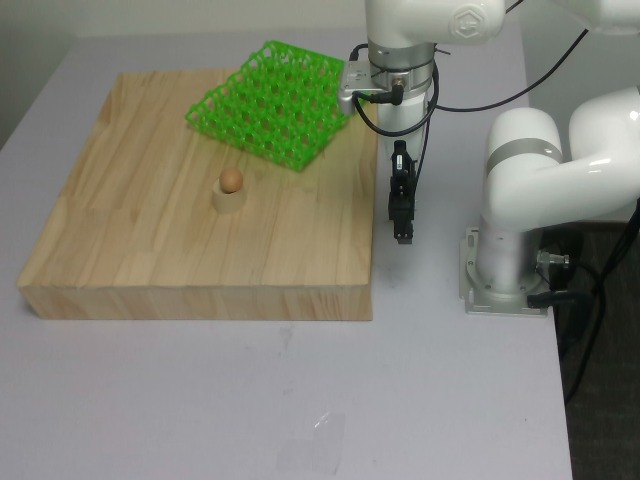}\hfill\includegraphics[width=.192\linewidth]{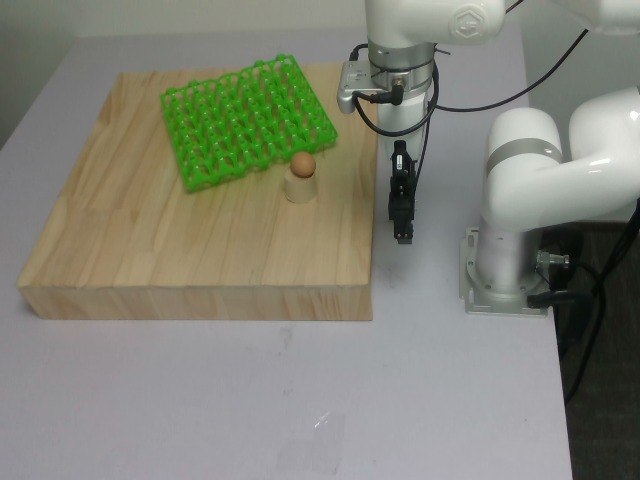}\par
\noindent \includegraphics[width=.192\linewidth]{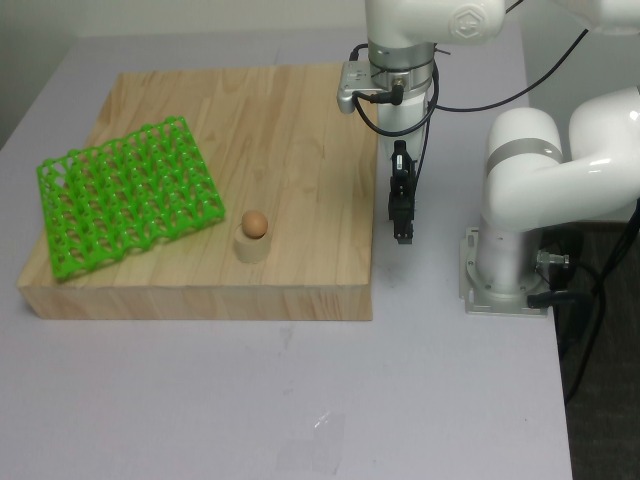}\hfill\includegraphics[width=.192\linewidth]{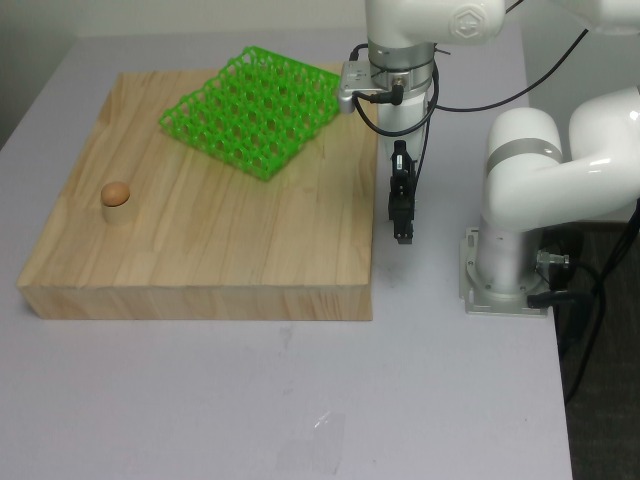}\hfill\includegraphics[width=.192\linewidth]{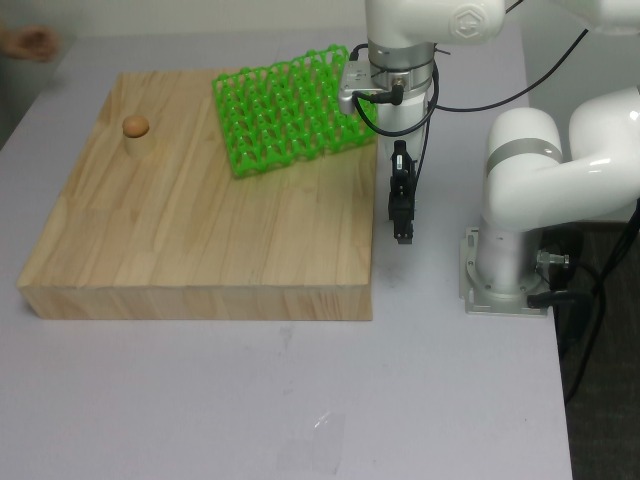}\hfill\includegraphics[width=.192\linewidth]{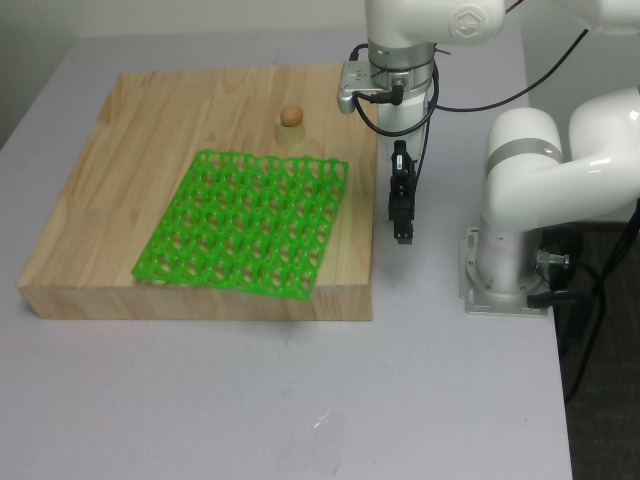}\hfill\includegraphics[width=.192\linewidth]{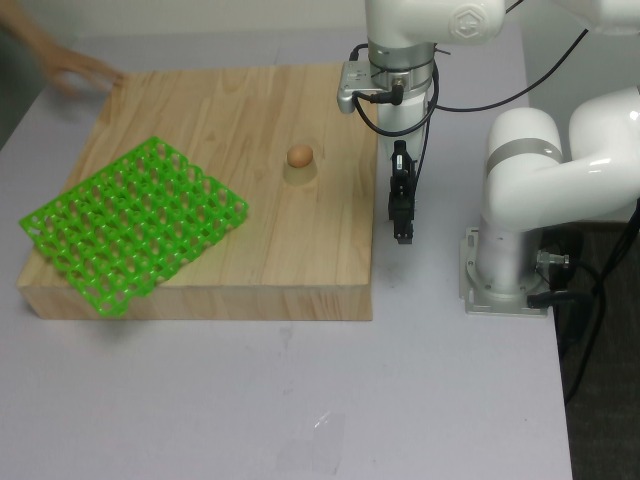}\par
\end{minipage}\par
\par\smallskip\noindent\begin{minipage}{\columnwidth}\textbf{Charger Press}\par
\noindent \includegraphics[width=.192\linewidth]{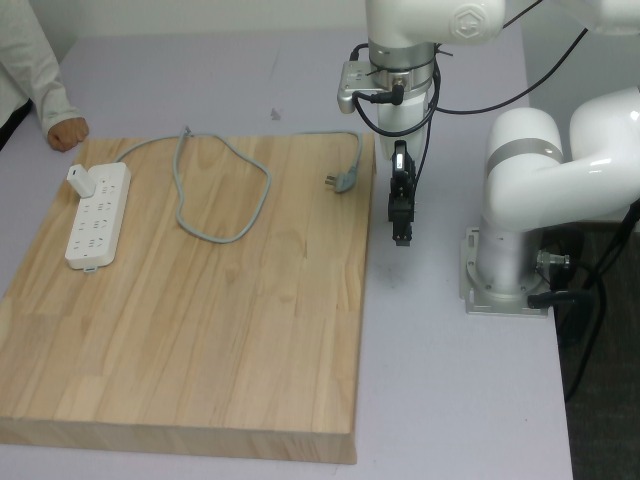}\hfill\includegraphics[width=.192\linewidth]{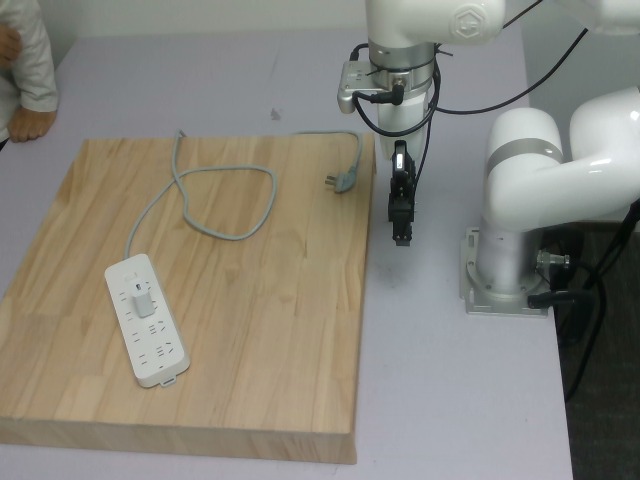}\hfill\includegraphics[width=.192\linewidth]{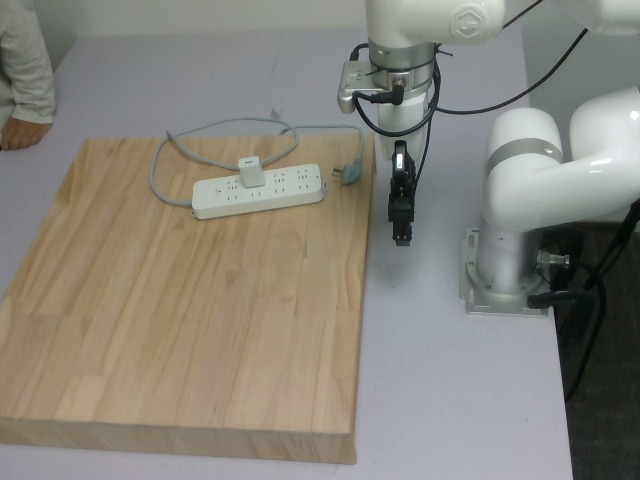}\hfill\includegraphics[width=.192\linewidth]{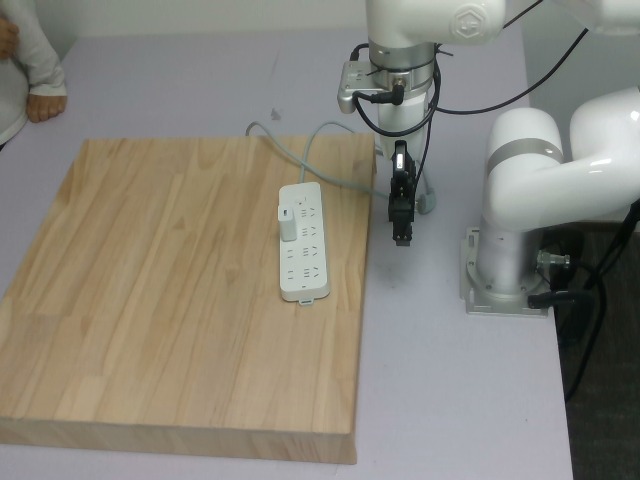}\hfill\includegraphics[width=.192\linewidth]{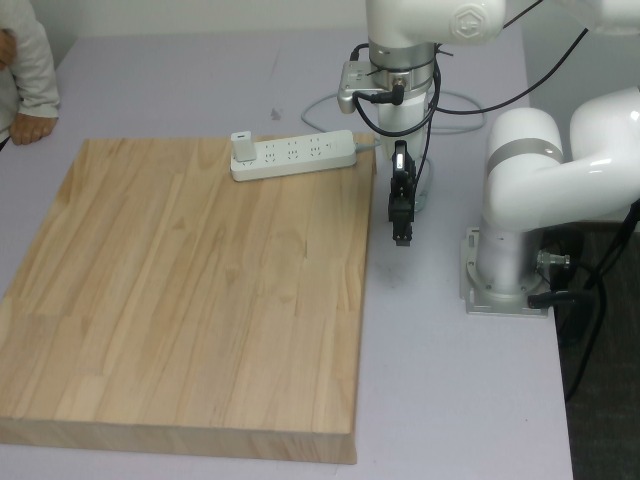}\par
\noindent \includegraphics[width=.192\linewidth]{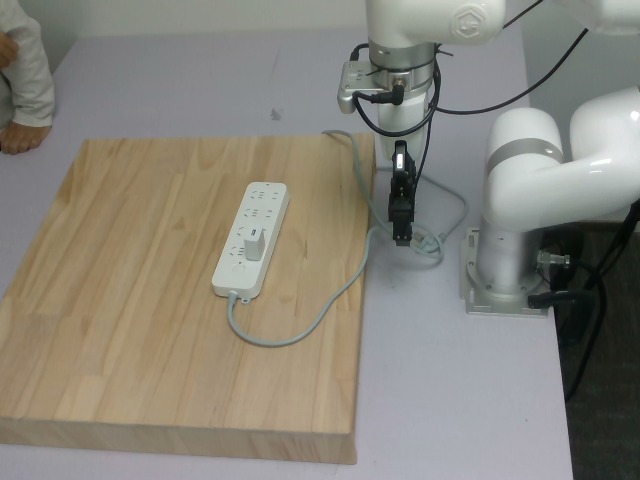}\hfill\includegraphics[width=.192\linewidth]{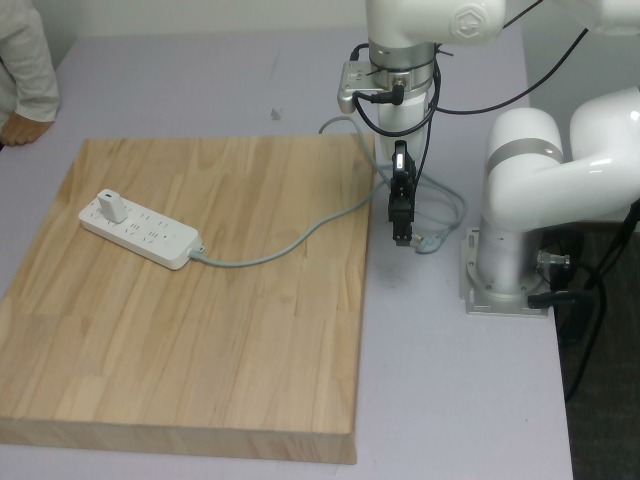}\hfill\includegraphics[width=.192\linewidth]{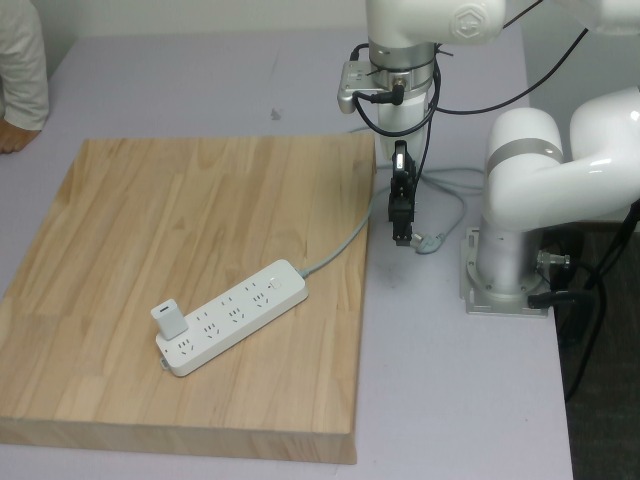}\hfill\includegraphics[width=.192\linewidth]{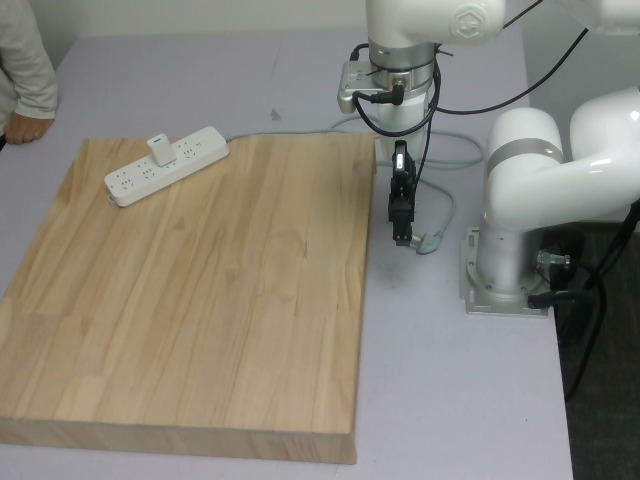}\hfill\includegraphics[width=.192\linewidth]{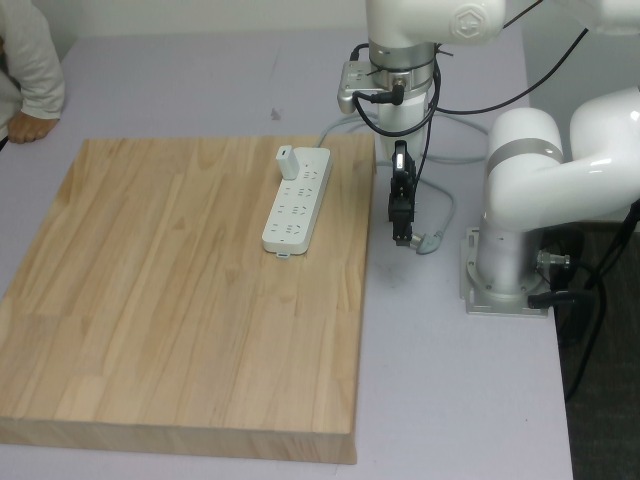}\par
\end{minipage}\par
\par\smallskip\noindent\begin{minipage}{\columnwidth}\textbf{Pullback Car}\par
\noindent \includegraphics[width=.192\linewidth]{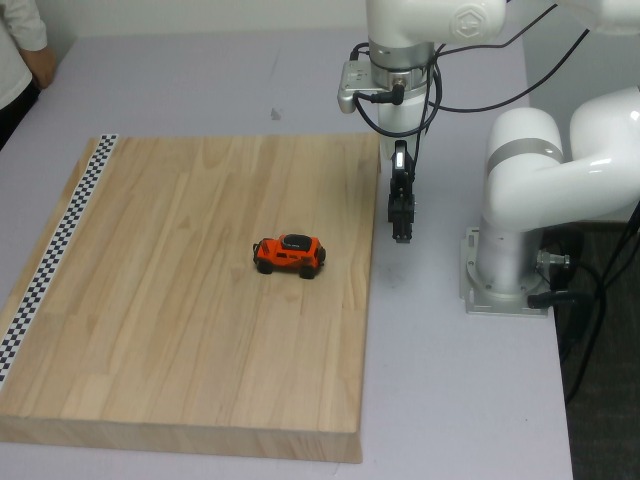}\hfill\includegraphics[width=.192\linewidth]{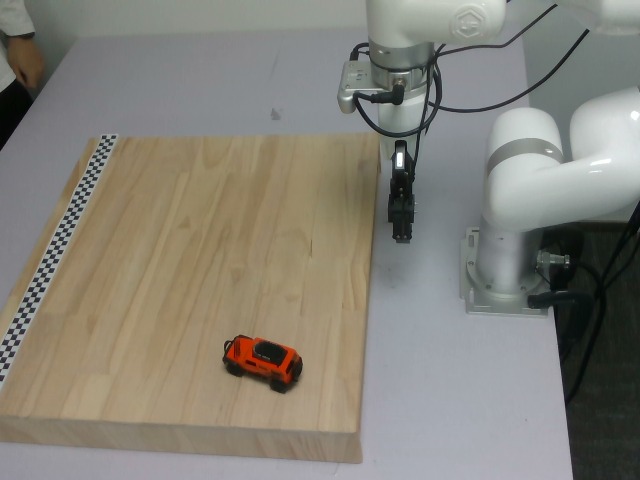}\hfill\includegraphics[width=.192\linewidth]{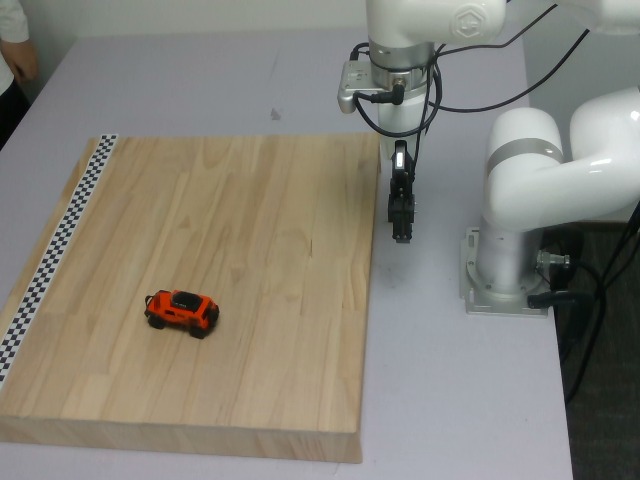}\hfill\includegraphics[width=.192\linewidth]{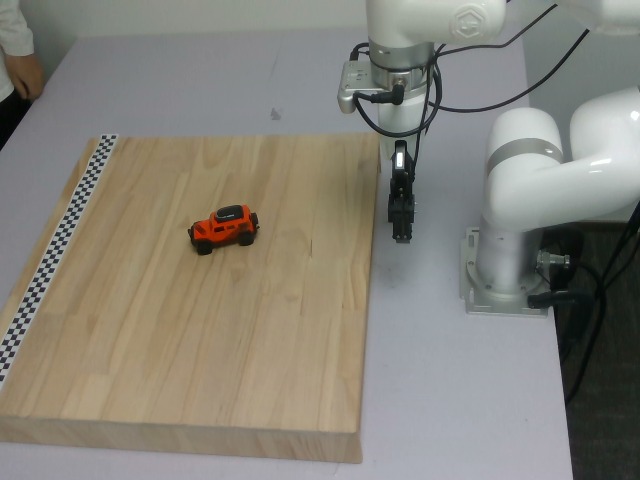}\hfill\includegraphics[width=.192\linewidth]{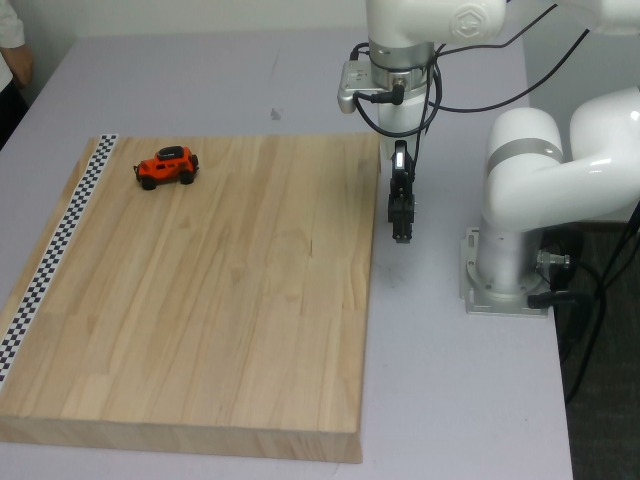}\par
\noindent \includegraphics[width=.192\linewidth]{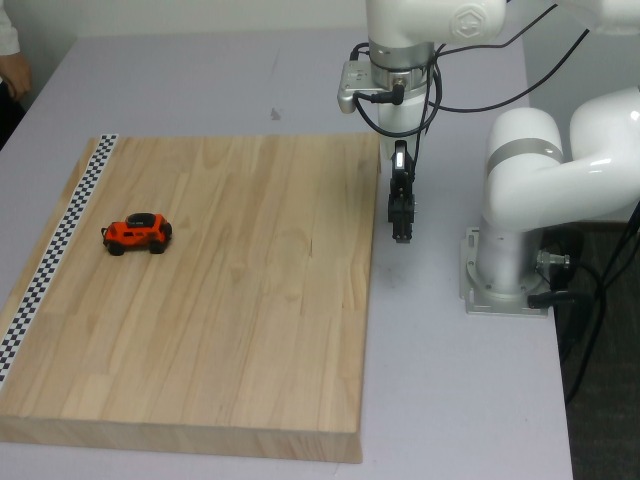}\hfill\includegraphics[width=.192\linewidth]{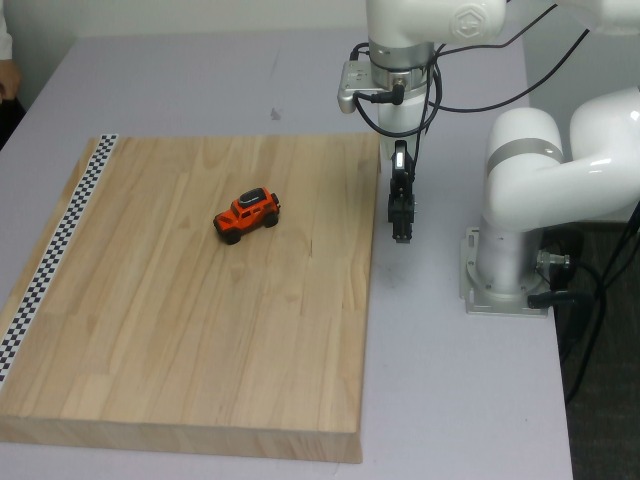}\hfill\includegraphics[width=.192\linewidth]{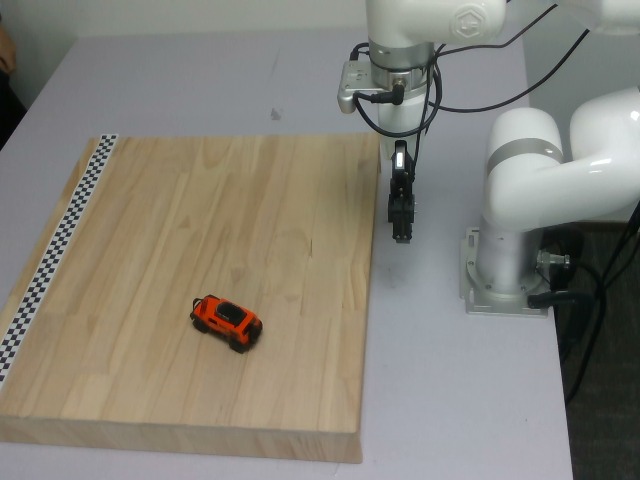}\hfill\includegraphics[width=.192\linewidth]{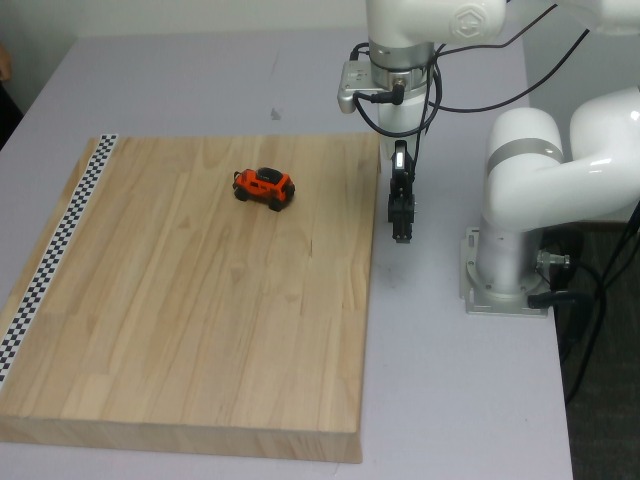}\hfill\includegraphics[width=.192\linewidth]{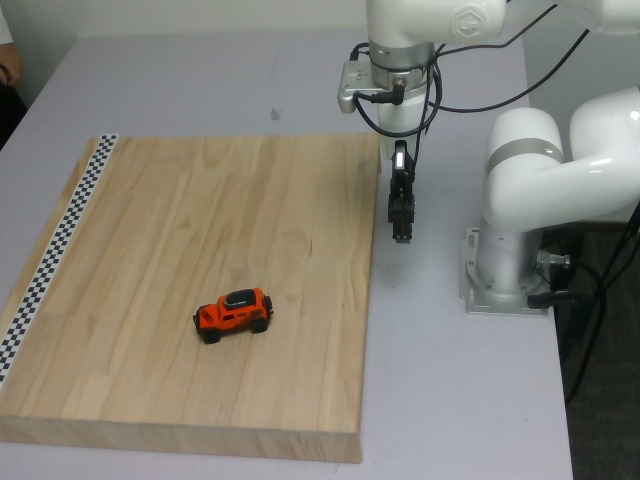}\par
\end{minipage}\par
\par\smallskip\noindent\begin{minipage}{\columnwidth}\textbf{Whiteboard Erase}\par
\noindent \includegraphics[width=.192\linewidth]{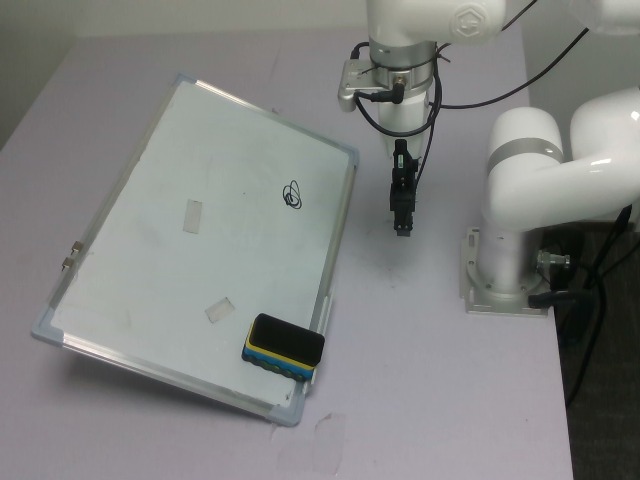}\hfill\includegraphics[width=.192\linewidth]{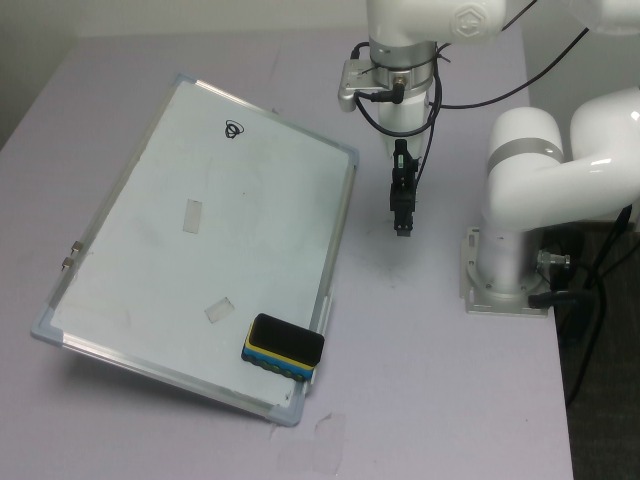}\hfill\includegraphics[width=.192\linewidth]{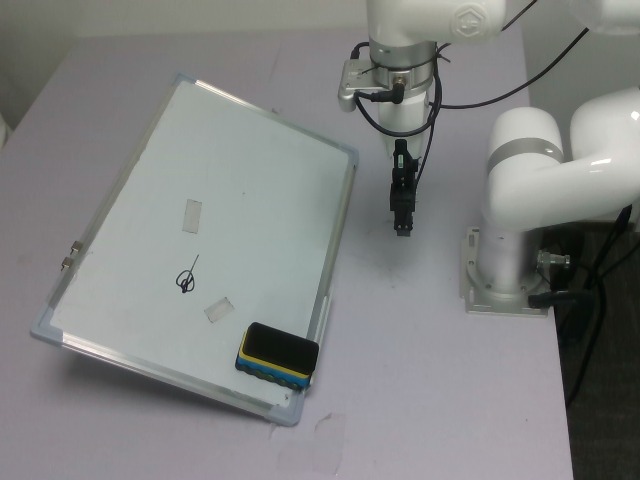}\hfill\includegraphics[width=.192\linewidth]{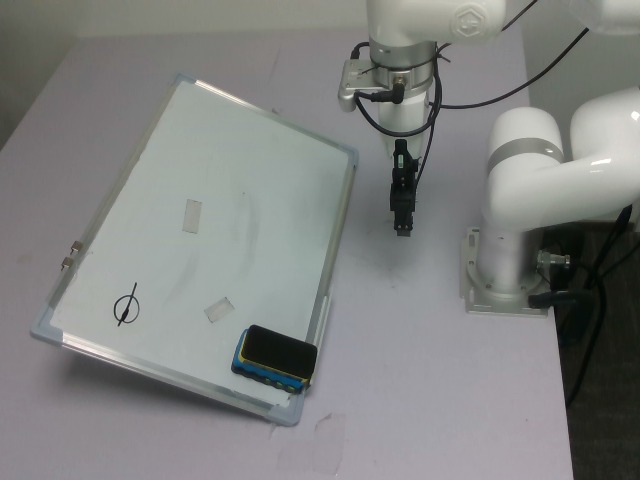}\hfill\includegraphics[width=.192\linewidth]{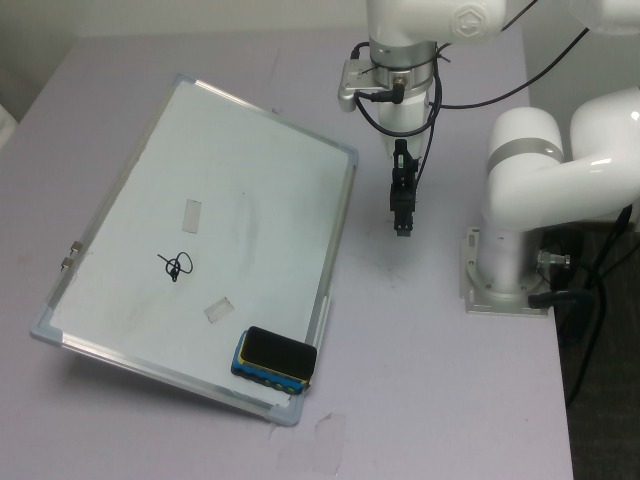}\par
\noindent \includegraphics[width=.192\linewidth]{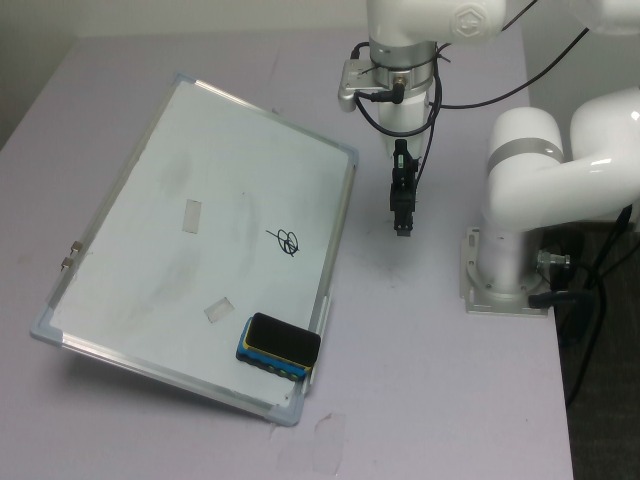}\hfill\includegraphics[width=.192\linewidth]{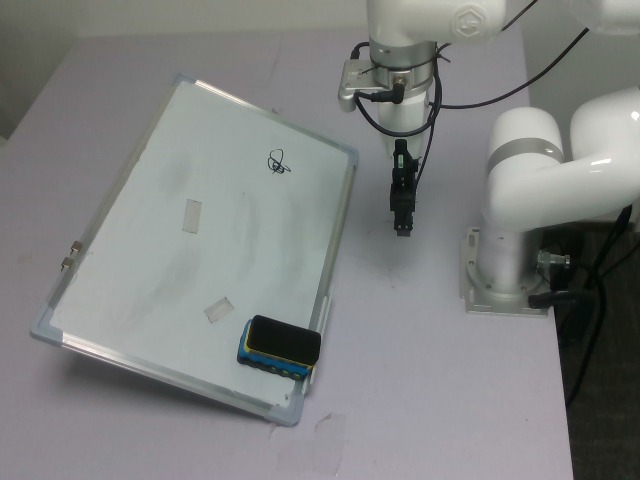}\hfill\includegraphics[width=.192\linewidth]{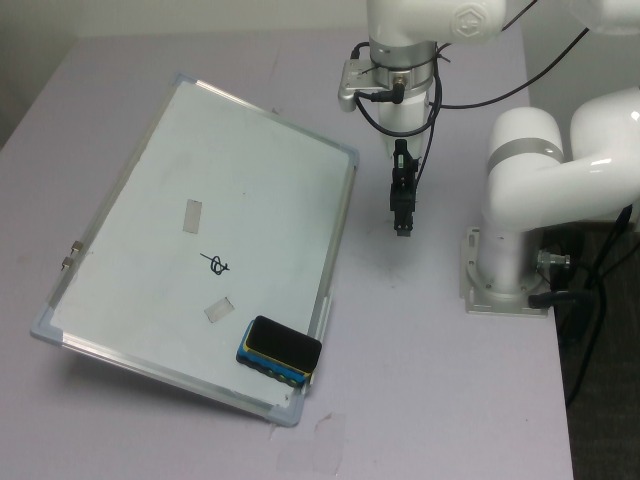}\hfill\includegraphics[width=.192\linewidth]{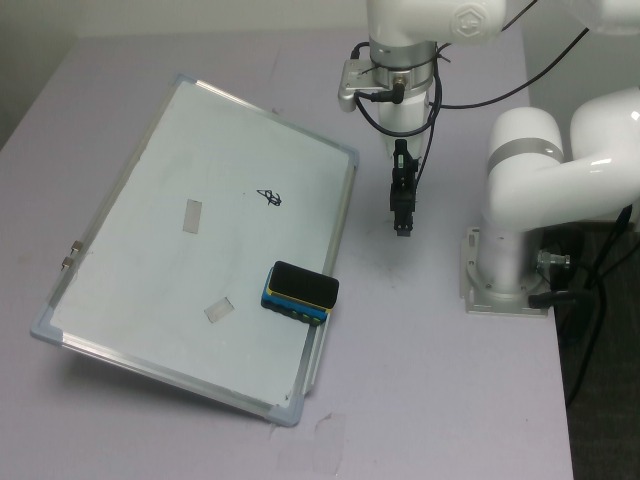}\hfill\includegraphics[width=.192\linewidth]{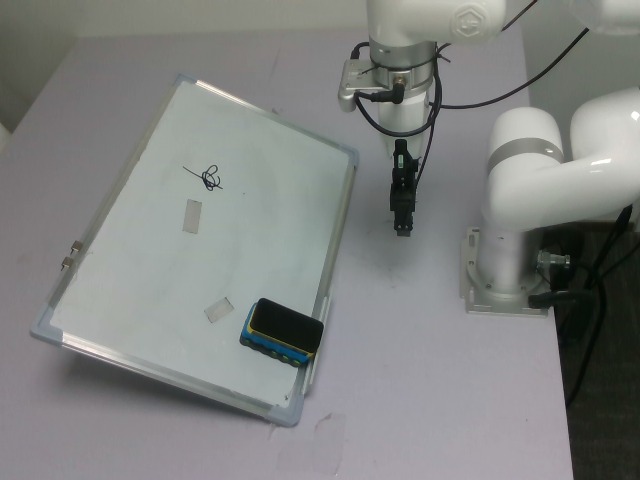}\par
\end{minipage}\par

\subsection{Real-World Qualitative Comparisons}
\label{sec:realworld_qualitative_comparisons}
\begin{figure*}[!tp]
\centering
\includegraphics[width=\textwidth,height=.87\textheight,keepaspectratio,trim=12bp 55bp 12bp 55bp,clip]{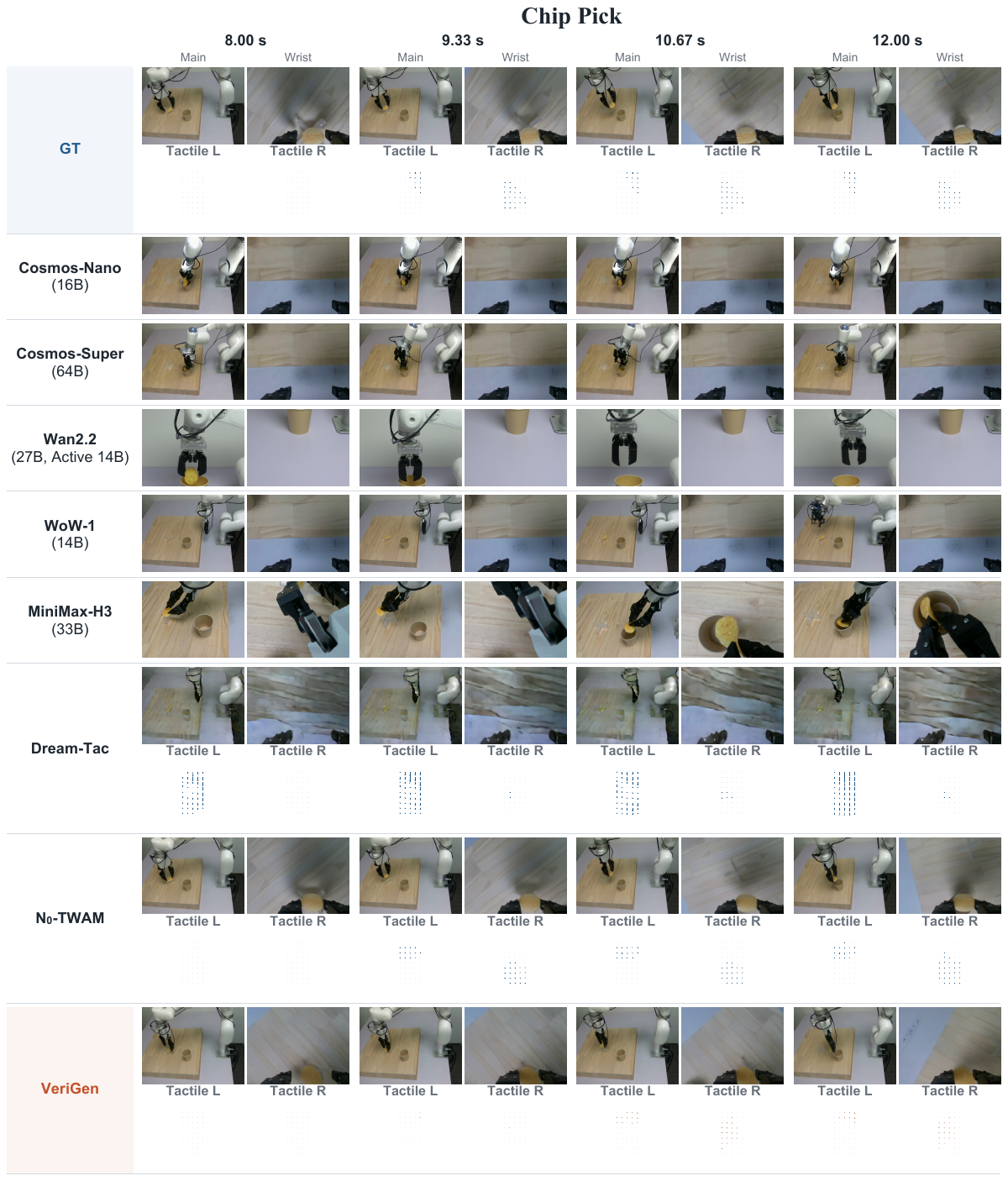}
\caption{\textbf{Cross-method qualitative comparison for Chip Pick.} GT and generated Main/Wrist views at four aligned target times, with bilateral tactile XY displacement fields where available. Baseline seeds are selected to illustrate failure modes.}
\label{fig:realworld_chip_pick_qual}
\end{figure*}

Figures~\ref{fig:realworld_chip_pick_qual}--\ref{fig:realworld_whiteboard_erase_qual} show the task-wise comparisons discussed in Appendix~\ref{sec:appendix_cross_method_qualitative}. Within each task, all methods use the same held-out episode and conditioning Main/Wrist images and caption. Four target times are selected on the native Dream-Tac endpoint grid; each model contributes its nearest existing native output, without temporal interpolation or held-frame extensions. GT appears at the top and \WMName{} at the bottom. The tactile vector fields visualize left/right sensor XY displacements in millimeters relative to the same initial reference, with common axes and a shared $3\times$ display gain. Blue arrows denote GT and baseline tactile fields, while orange arrows denote \WMName{}. Dream-Tac and $\mathcal{N}_0$-TWAM generate tactile RGB images, so we obtain their displacement fields with the unmodified Photon SDK. 

\begin{figure*}[!tp]
\centering
\includegraphics[width=\textwidth,height=.87\textheight,keepaspectratio,trim=12bp 55bp 12bp 55bp,clip]{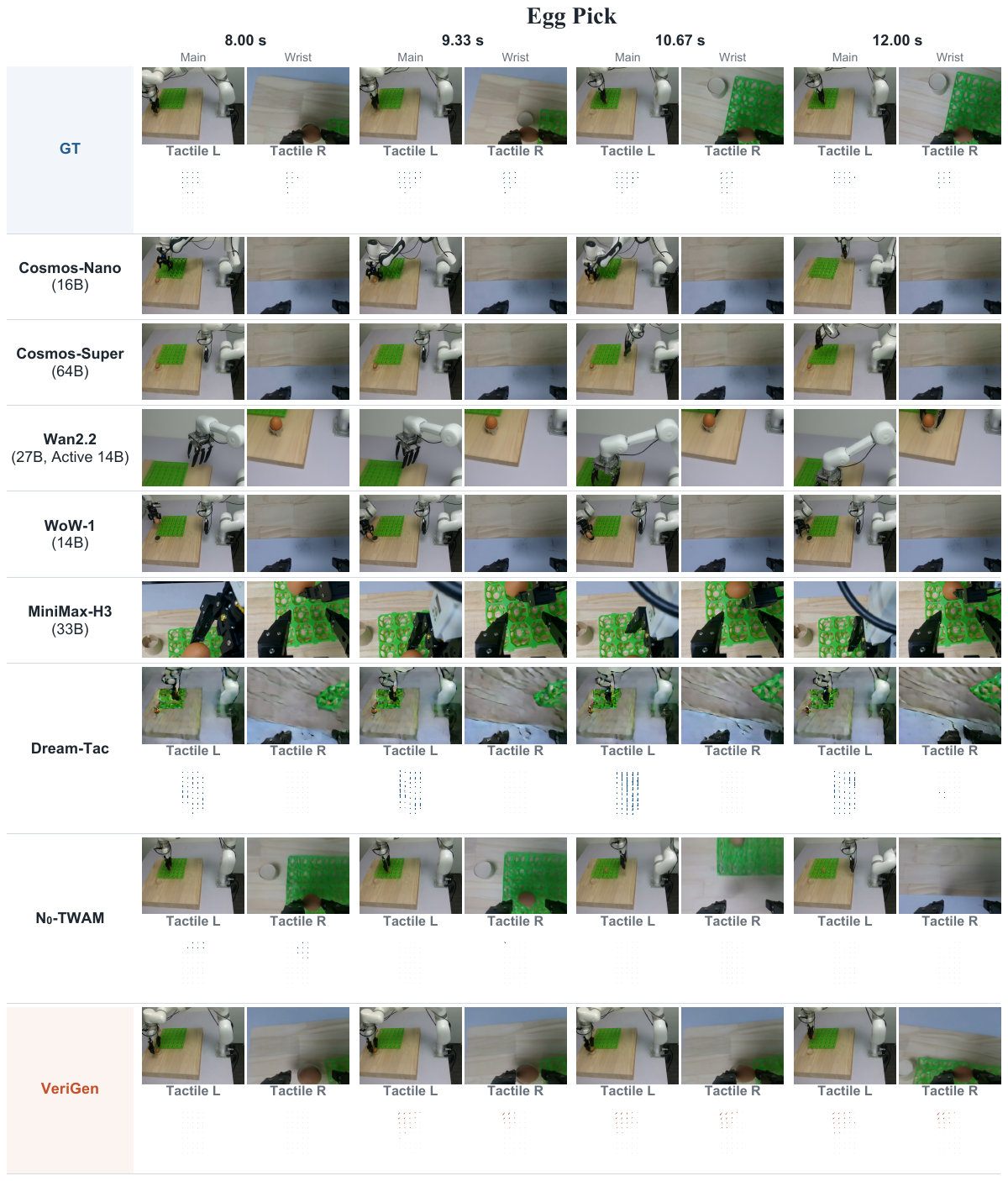}
\caption{\textbf{Cross-method qualitative comparison for Egg Pick.} GT and generated Main/Wrist views at four aligned target times, with bilateral tactile XY displacement fields where available. Baseline seeds are selected to illustrate failure modes.}
\label{fig:realworld_egg_pick_qual}
\end{figure*}

\begin{figure*}[!tp]
\centering
\includegraphics[width=\textwidth,height=.87\textheight,keepaspectratio,trim=12bp 55bp 12bp 55bp,clip]{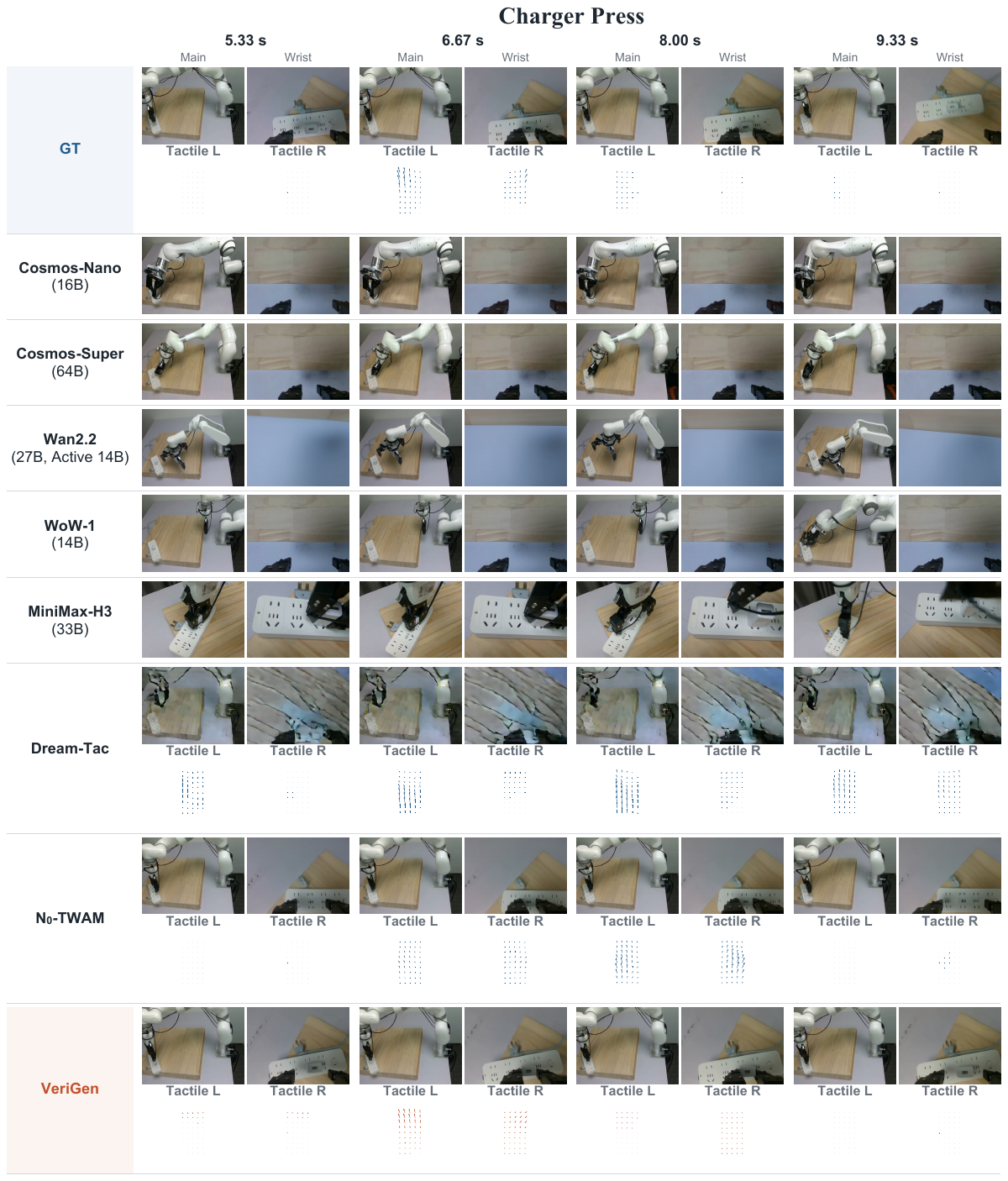}
\caption{\textbf{Cross-method qualitative comparison for Charger Press.} GT and generated Main/Wrist views at four aligned target times, with bilateral tactile XY displacement fields where available. Baseline seeds are selected to illustrate failure modes.}
\label{fig:realworld_charger_press_qual}
\end{figure*}

\begin{figure*}[!tp]
\centering
\includegraphics[width=\textwidth,height=.87\textheight,keepaspectratio,trim=12bp 55bp 12bp 55bp,clip]{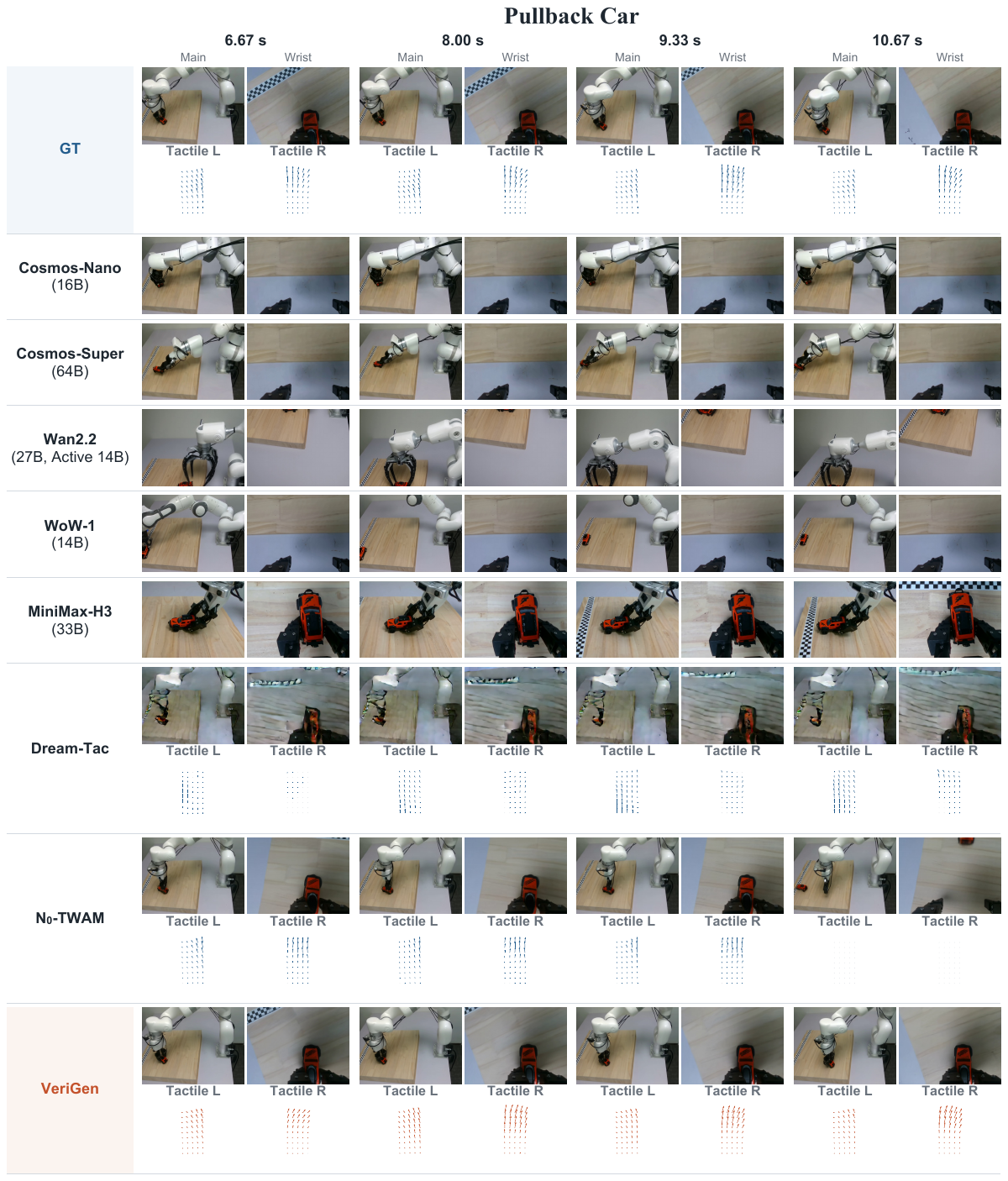}
\caption{\textbf{Cross-method qualitative comparison for Pullback Car.} GT and generated Main/Wrist views at four aligned target times, with bilateral tactile XY displacement fields where available. Baseline seeds are selected to illustrate failure modes.}
\label{fig:realworld_pullback_car_qual}
\end{figure*}

\begin{figure*}[!tp]
\centering
\includegraphics[width=\textwidth,height=.87\textheight,keepaspectratio,trim=12bp 55bp 12bp 55bp,clip]{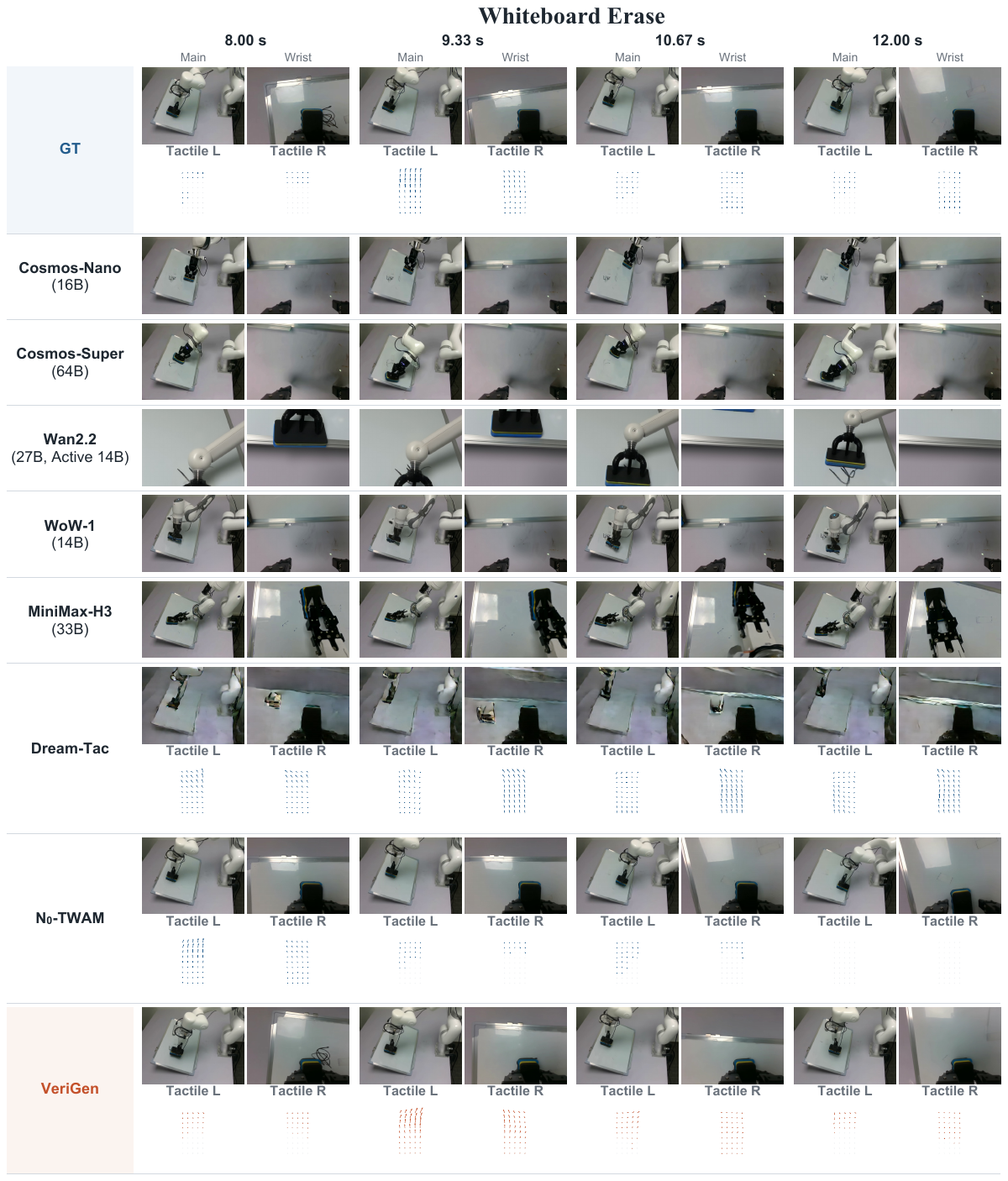}
\caption{\textbf{Cross-method qualitative comparison for Whiteboard Erase.} GT and generated Main/Wrist views at four aligned target times, with bilateral tactile XY displacement fields where available. Baseline seeds are selected to illustrate failure modes.}
\label{fig:realworld_whiteboard_erase_qual}
\end{figure*}

\begin{figure*}[!tp]
    \noindent\begin{minipage}{\columnwidth}
    \subsection{OpenViTac Qualitative Comparisons}
    \label{sec:openvitac_qualitative_comparisons}
    \end{minipage}\par\medskip
    \centering
    \includegraphics[width=\textwidth]{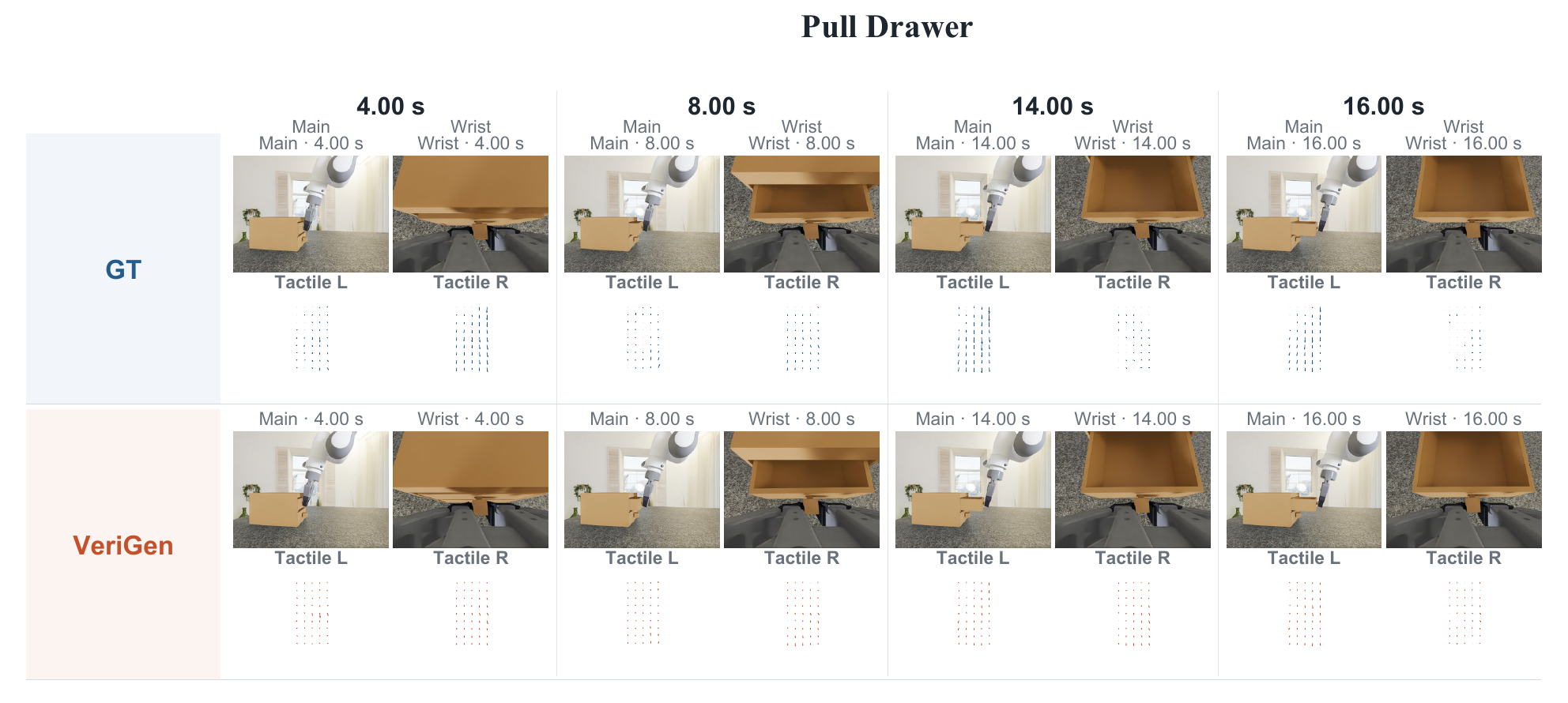}
    \caption{\textbf{Comparison of generated and ground-truth visuo-tactile trajectories for Pull Drawer.} Ground truth (top) and \WMName{} (bottom) are compared at four matched timestamps from a held-out Pull Drawer episode. Main/Wrist images and bilateral tactile XY displacement fields are shown.}
    \label{fig:openvitac_pull_drawer_qual}
\end{figure*}

\begin{figure*}[!tp]
    \centering
    \includegraphics[width=\textwidth]{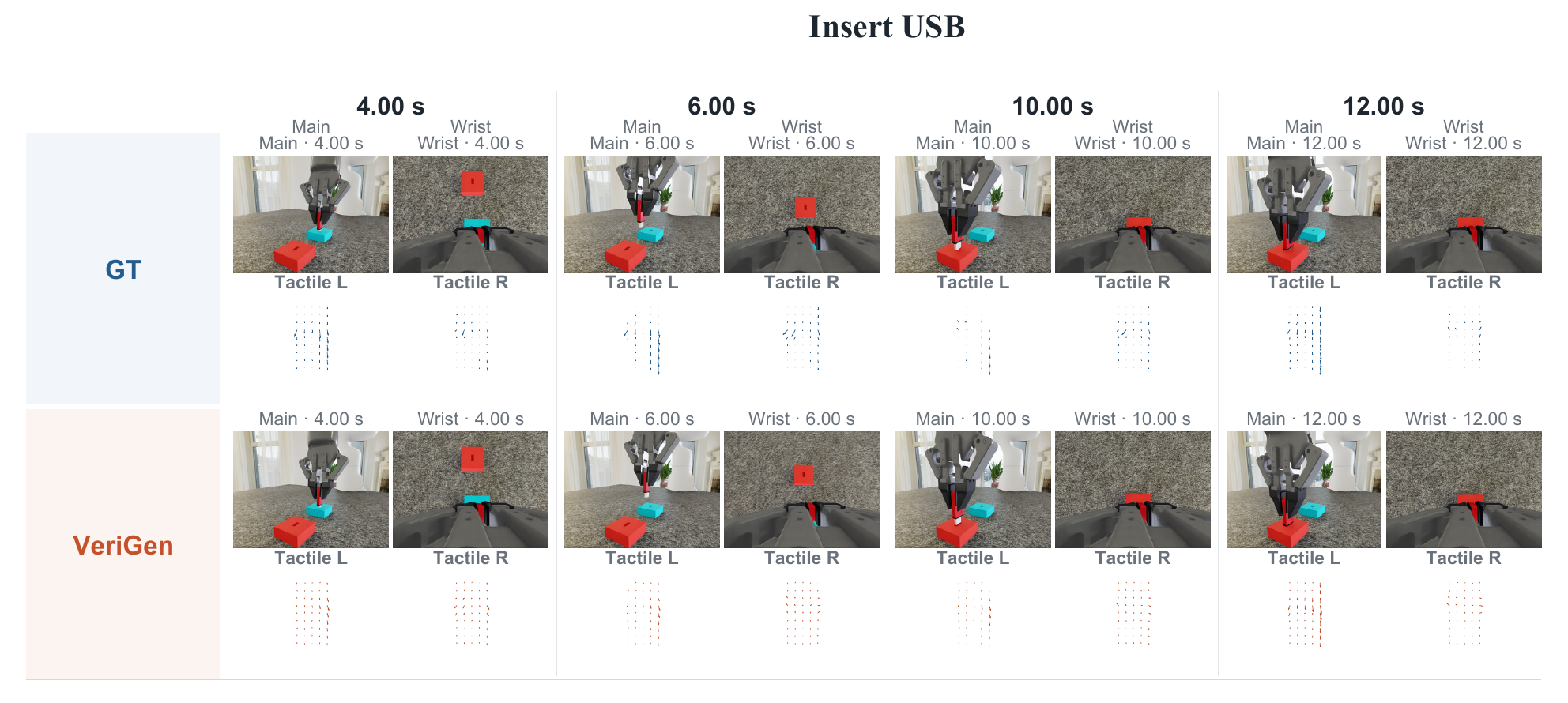}
    \caption{\textbf{Comparison of generated and ground-truth visuo-tactile trajectories for Insert USB.} Ground truth (top) and \WMName{} (bottom) are compared at four matched timestamps from a held-out Insert USB episode. Main/Wrist images and bilateral tactile XY displacement fields are shown.}
    \label{fig:openvitac_insert_usb_qual}
\end{figure*}

\onecolumn
\noindent\begin{minipage}{\textwidth}
    \captionsetup{hypcap=false}
    \centering
    \includegraphics[width=.81\linewidth]{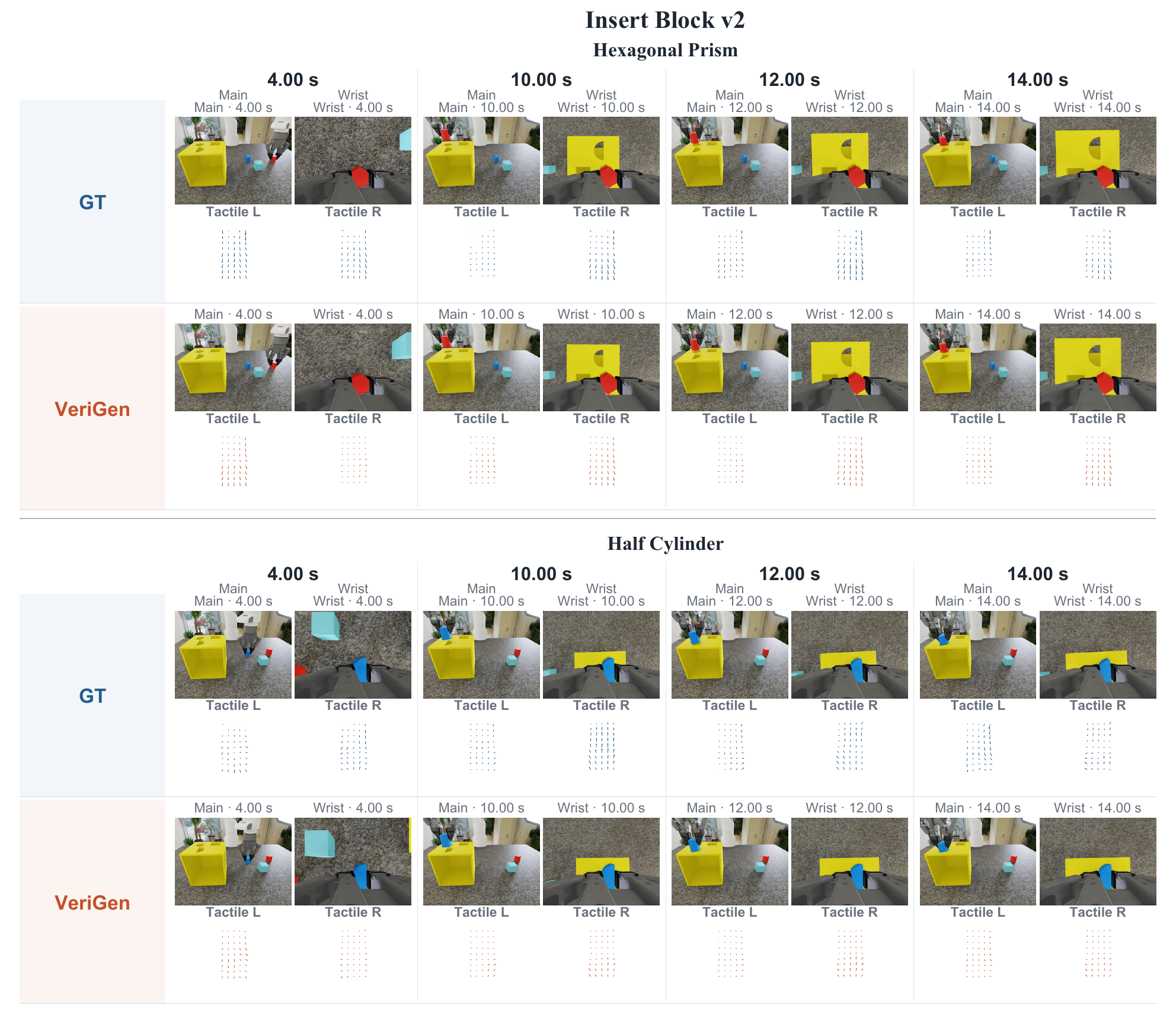}
    \captionof{figure}{\textbf{Comparison of generated and ground-truth visuo-tactile trajectories for Insert Block v2.} Hexagonal-prism insertion (upper panel) and half-cylinder insertion (lower panel) are shown using two held-out episodes. Within each panel, ground truth (top) and \WMName{} (bottom) are compared at four matched timestamps, with Main/Wrist images and bilateral tactile XY displacement fields.}
    \label{fig:openvitac_insert_block_v2_qual}
    \par\medskip
    \includegraphics[width=.81\linewidth]{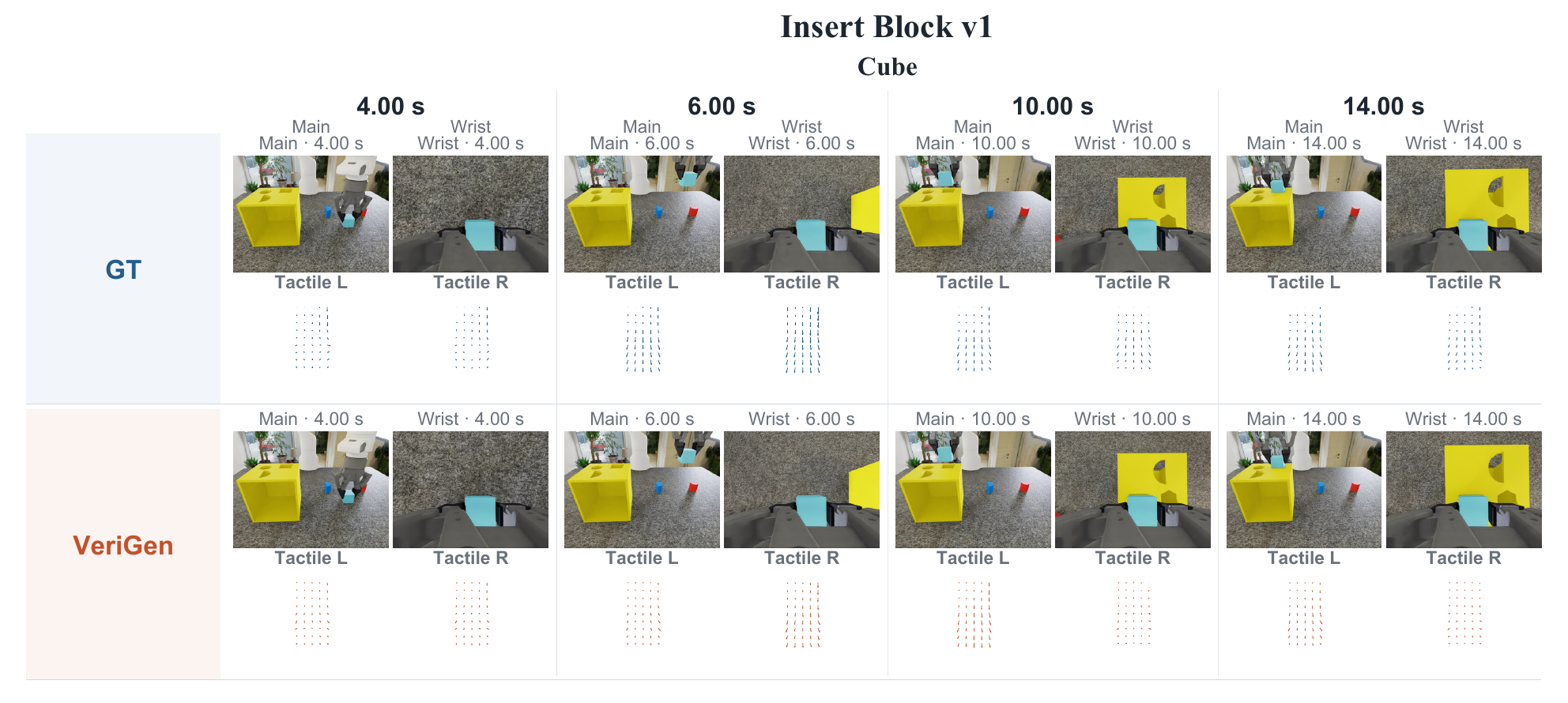}
    \captionof{figure}{\textbf{Comparison of generated and ground-truth visuo-tactile trajectories for Insert Block v1.} Ground truth (top) and \WMName{} (bottom) are compared at four matched timestamps from a held-out cube-insertion episode. Main/Wrist images and bilateral tactile XY displacement fields are shown.}
    \label{fig:openvitac_insert_block_v1_qual}
\end{minipage}

\end{document}